\documentclass{article}

\usepackage{PRIMEarxiv}

\usepackage[utf8]{inputenc} 
\usepackage[T1]{fontenc}    
\usepackage{hyperref}       
\usepackage{xcolor}         

\usepackage{url}            
\usepackage{booktabs}       
\usepackage{amsfonts}       
\usepackage{nicefrac}       
\usepackage{microtype}      
\usepackage{lipsum}
\usepackage{fancyhdr}       
\usepackage{graphicx}       
\usepackage{color}
\usepackage{algorithm}
\usepackage{algpseudocode}
\usepackage{multirow}
\usepackage{subcaption} 
\usepackage{amsmath}
\usepackage{amssymb}
\usepackage{comment}
\usepackage{bbm}
\usepackage{eso-pic}

\newtheorem{theorem}{Theorem}
\newtheorem{lemma}[theorem]{Lemma}

\newtheorem{definition}{Definition}
\newtheorem{remark}{Remark}

\newcommand{\deltaalphaRobustness}{$(\delta,\alpha)$-robustness }
\newcommand{\deltaalphaRobust}{$(\delta, \alpha)$-robust }
\newcommand{\deltaRobust}{$\delta$-robust }
\newcommand{\growingSpheres}{\textsc{GrowingSpheres }}
\newcommand{\dice}{\textsc{Dice }}
\newcommand{\betaRCE}{\textsc{BetaRCE }}
\newcommand{\face}{\textsc{FACE }}
\newcommand{\rbr}{\textsc{RBR }}
\newcommand{\roar}{\textsc{ROAR }}
\newcommand{\robx}{\textsc{RobX }}

\definecolor{algcomment}{rgb}{0, 0.1, 0.4}

\definecolor{fig2_blue}{RGB}{76,114,176}
\definecolor{fig2_yellow}{RGB}{255,179,45}
\definecolor{fig2_gray}{RGB}{69,69,69}
\definecolor{fig2_red}{RGB}{196,78,82}
\definecolor{fig2_green}{RGB}{85,168,104}

\definecolor{gray_watermark}{RGB}{69,69,69} 
\definecolor{black}{RGB}{0,0,0} 
\definecolor{mycitecolor}{RGB}{4, 212, 11} 

\hypersetup{
    breaklinks,
    colorlinks=true,
    linkcolor=algcomment,      
    citecolor=mycitecolor,      
    urlcolor=algcomment,        
    filecolor=magenta           
}

\title{Counterfactual Explanations with Probabilistic Guarantees 
       on their Robustness to Model Change}

\author{
  Ignacy St\k{e}pka\textsuperscript{1}\thanks{\footnotesize Contact: ignacy~[.]~stepka~[at]~cs.put.poznan.pl}~~
  Jerzy Stefanowski\textsuperscript{1}~
  Mateusz Lango\textsuperscript{1,2} \\
  \textsuperscript{1}Poznan University of Technology, Poznan, Poland\\
  \textsuperscript{2}Charles University, Prague, Czech Republic
}
\begin{document}
\AddToShipoutPictureBG*{
  \AtPageLowerLeft{
    \hspace{-1.37cm} 
    \raisebox{2.13cm}{ 
      \parbox{\paperwidth}{
        \centering \footnotesize \textcolor{black}{Accepted at the 31st ACM SIGKDD Conference on Knowledge Discovery and Data Mining (KDD ’25).}
      }
    }
  }
}
\maketitle


\begin{abstract}

Counterfactual explanations (CFEs) guide users on how to adjust inputs to machine learning models to achieve desired outputs. 
While existing research primarily addresses static scenarios, real-world applications often involve data or model changes, potentially invalidating previously generated CFEs and rendering user-induced input changes ineffective.
Current methods addressing this issue often support only specific models or change types, require extensive hyperparameter tuning, or fail to provide probabilistic guarantees on CFE robustness to model changes. 
This paper proposes a novel approach for generating CFEs that provides probabilistic guarantees for any model and change type, while offering interpretable and easy-to-select hyperparameters. 
We establish a theoretical framework for probabilistically defining robustness to model change and demonstrate how our \betaRCE method directly stems from it. 
\betaRCE is a post-hoc method applied alongside a chosen base CFE generation method to enhance the quality of the explanation beyond robustness. 
It facilitates a transition from the base explanation to a more robust one with user-adjusted probability bounds.  
Through experimental comparisons with baselines, we show that \betaRCE yields robust, most plausible, and closest to baseline counterfactual explanations.

\end{abstract}


\section{Introduction}
Counterfactual explanations (CFEs) are one of the most popular forms of explaining decisions made by complex, black-box machine learning (ML) algorithms.
Briefly, a counterfactual explanation of a decision $y$ made for input $x$ is an instance $x^{cf}$ that is very similar to $x$ but produces a different, more desirable prediction $y'\neq y$.
Since CFEs can be interpreted as an answer to the question: “given the decision $y$ taken for input $x$, how should $x$ be changed to produce the alternative decision $y'$?”, they offer actionable feedback to the user.
Over the years, it has been appreciated by stakeholders in various application areas such as supporting loan 
decisions \cite{wachter_counterfactual_2017}, job recruitment \cite{pearl2016_causal_book}, medicine \cite{counterfactuals_medical}, and many others \cite{guidotti_counterfactual_2022}.

Even though the basic definition of a counterfactual \cite{wachter_counterfactual_2017} specifies only two basic properties: \textit{validity} (ensuring the desired classification $y'$) and \textit{proximity} (small distance between $x$ and $x^{cf}$), many additional properties are useful from both the application \cite{keane_if_2021} and user \cite{forster_evaluating_2020,keane_if_2021} point of view.
These properties include  \textit{sparsity} (modifying values of only few features), \textit{actionability} (realistic feature changes), \textit{plausibility} (proximity to the data distribution), and many others.
Numerous methods for generating counterfactuals with different properties have been proposed 
\cite{guidotti_counterfactual_2022},
yet almost all of them deal with static problem settings, overlooking the counterfactuals' \emph{robustness to model change}.

Since counterfactuals are intended to deliver actionable feedback, they must remain valid for the period of time that is necessary for the end-user to act on the proposed changes.
Note that CFEs are generated for a fixed model, yet many applications are inherently dynamic with the model changing over time.
There are many possible causes incurring a model change, some of which include obtaining new training data, adjusting hyperparameters or model architecture, and having to remove some training data due to data expiration policies or privacy laws (someone may request their personal data to be removed \cite{ginart2019forget}). 
In such scenario, it is important to preserve the validity of the counterfactual  for the newly retrained model, so that the user will still get the desired decision while acting on the recourse offered to them before the model change.

For example, suppose a bank has generated a CFE for a customer who has been denied a loan. 
During the time in which the customer tries to improve their financial profile to meet the requirements specified in the CFE, the bank may need to update its model. 
In such scenario, it would be desired that the recommendation issued to the user would still be valid and lead to the approval of a loan.

This highlights the need for a fresh look at the properties of counterfactuals in the context of changing environments, and in particular, leads us to the concept of robustness to model change. 
This challenge 
has been considered from various perspectives, including robustness to input perturbations \cite{artelt_evaluating_2021,EhyaeiRobustFairness2023}, to model changes \cite{upadhyay_towards_2021,dutta_robust_2022,jiang_formalising_2022} or imperfect realizations of recommendations \cite{pawelczyk_probabilistically_2023,guyomard_generating_2023,maragno_finding_2023}. 
In this paper, we focus on the robustness to moderate model changes, where we expect the CFE to remain valid over time with regard to the class indicated before retraining. 
This topic has not been thoroughly explored yet, and the existing introduced methods exhibit significant limitations. Some of these methods rely on impractical assumptions, while others are restricted to specific models and often require substantial human effort to tune their non-interpretable hyperparameters.
Most importantly, they typically provide no statistical guarantees of robustness and do not specify against which model changes the counterfactual is robust.

In this work, we propose to examine the robustness of CFEs from a probabilistic perspective, empowering users and stakeholders to incorporate statistical estimates of CFE validity. 
To this end, we introduce a Bayesian-inspired framework that assesses the probability that a counterfactual explanation remains valid to model changes, clearly defined by a so-called admissible model space. 

Moreover, we also introduce a novel post-hoc method called \textsc{BetaRCE}, which generates robust counterfactual explanations in a model-agnostic manner.  
The method can be applied on top of any base counterfactual generation method, which may be selected to meet user expectations regarding different properties of CFEs. 
\betaRCE robustifies the given counterfactual by moving it in attribute space until its estimated robustness meets user requirements. 

\begin{figure}[tbp]
\centering
    \includegraphics[width=0.6\textwidth]{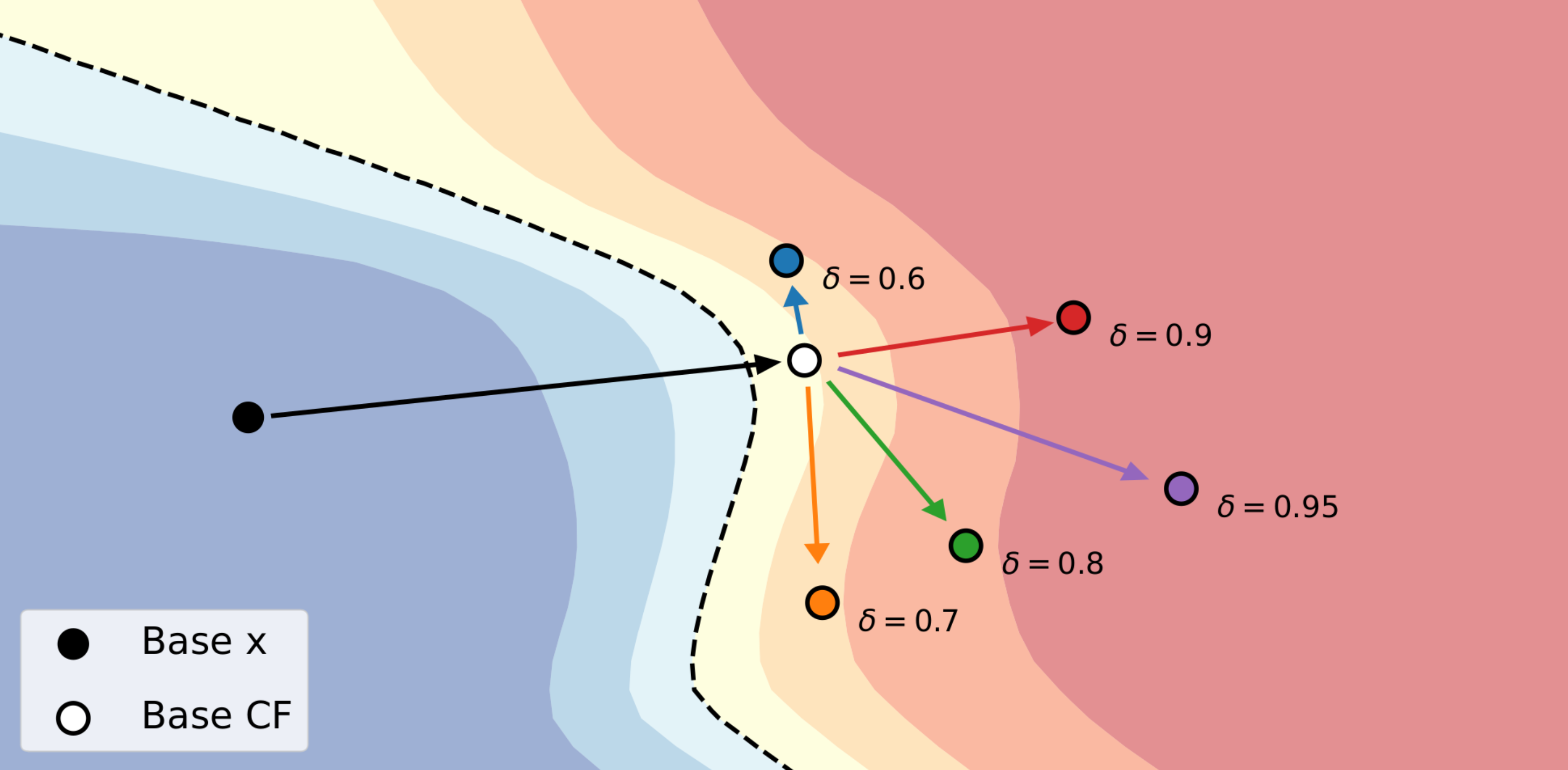}
    \caption{Our method \betaRCE post-hoc generates counterfactuals at desired levels of robustness to model change, having some probabilistic properties. First, the base CFE is generated using any base method, then \betaRCE is applied to move that CFE to a region satisfying \deltaalphaRobustness.}
    \label{fig:thumbnail}
\end{figure}

Fig.~\ref{fig:thumbnail} illustrates the essence of \textsc{BetaRCE}. Initially, we obtain the base CFE using a base explanation method for a given input example. Subsequently, we move this base CFE to a safer data region, i.e., an area with a low probability of change in classification. Here, $\delta$ represents the lower bound of the probability of a CFE preserving its original class. Higher $\delta$ values correspond to a higher degree of robustness, which often comes with a more substantial deviation from the base CFE. Crucially, in \textsc{BetaRCE} we let the user select the desired $\delta$, giving them the opportunity to decide on the tradeoff between the explanation's degree of robustness and proximity to the original example.  

To validate our method, we conducted experiments that confirmed the introduced probabilistic bounds hold in practice for various types of model changes. Furthermore, the experiments showed that counterfactuals generated by \betaRCE remain close to the base CFE and better preserve their properties compared to previous methods aimed at improving robustness.

To sum up, the main contributions of our paper are as follows:
\begin{enumerate}
    \item Designing a new theoretical framework for assessing the robustness of counterfactual explanations to model changes. The framework provides probabilistic guarantees of the robustness of counterfactual explanations according to user-defined expectations.
    \item Introducing the model-agnostic \betaRCE method, which enhances CFE's robustness in a post-hoc manner, thereby making it applicable to any base counterfactual generation method.
    \item Conducting an experimental study to investigate the properties of \betaRCE and comparing its performance to selected reference methods.
\end{enumerate}

\section{Related Work}
\label{sec:rw-robustness}

Counterfactual explanations are a widely explored topic in the contemporary ML literature. Some well-known representatives of such methods (with an indication of the counterfactual properties  they  try to take into account) include \textsc{Wachter} \cite{wachter_counterfactual_2017} -- closest to the input instance, \growingSpheres \cite{laugel_inverse_2017} -- closest and sparse, \textsc{FACE} \cite{poyiadzi_face_2020} -- located in dense regions, \dice \cite{mothilal_explaining_2020} -- multiple closest, sparse, plausible and diverse counterfactuals.  
Over the years numerous methods were introduced, all suited for various motivations and aiming to attain certain properties that are considered useful from the perspective of the application scenario.

Notably, St\k{e}pka~et~al.~\cite{Stepka_2024} demonstrated through multi-criteria analysis that no single method excels across all evaluation measures simultaneously, highlighting the need for CFE methods tailored to specific application requirements.
While plenty of these methods have significantly advanced the field, they fall outside the scope of this paper, as we focus specifically on robustness. For comprehensive surveys on counterfactual methods, we direct readers to \cite{guidotti_counterfactual_2022,vermaReview2020}. 

Most existing methods, however, do not explicitly address the issue of robustness -- a challenge that has only recently gained attention \cite{mishra_survey_2023}. Nonetheless, due to its practical importance, robustness has already been explored from various perspectives.

In the following, we briefly outline the three primary definitions of counterfactual robustness:

\textbf{Robustness to input perturbations} requires that  similar input examples should yield similar CFEs, i.e. the difference between them should be minimized. This idea is closely related to the notion of individual fairness, where two similar individuals are expected to receive similar treatment \cite{artelt_evaluating_2021,EhyaeiRobustFairness2023,artelt_explain_2022}. 

\textbf{Robustness to imperfect recourse}  refers to a situation when the user acting on the counterfactual recommendations slightly fails to meet the exact proposed changes. It is argued, that actionable recourse should allow slight imperfections in the realization of the recommendation~\cite{pawelczyk_probabilistically_2023}. Proposed methods \cite{maragno_finding_2023,pawelczyk_probabilistically_2023,guyomard_generating_2023,virgolin_robustness_2023} generally aim to ensure that the class distribution in some neighborhood of the CFE is relatively pure in order to account for the mentioned imperfect execution of the recommendation.

\textbf{Robustness to model change} ensures that even after the underlying decision model is \textit{slightly modified}, the CFE will stay valid. The \textit{slightly modified} model is usually understood as a model of which the decision boundary is different compared to the original one due to various reasons, for example, retraining the model with different hyperparameters,  training on new data from the same (or slightly different) distribution, training the current model for another epoch, etc.  

Up to this point, a few methods attempting to solve this type of robustness have been proposed, as covered by a recent survey~\cite{jiang2024surveyRobust}. 
However, most of the introduced methods either do not quantify the extent of achieved robustness, or if they do, they are inherently model-specific, constraining the versatility of such approach. 
On top of that, they also do not enable decision-makers to directly specify their expectations regarding the desired level of robustness. 
Below, we briefly describe a selection of existing methods employing varied paradigms of ensuring robustness.  

Upadhyay~et~al.~\cite{upadhyay_towards_2021} introduced \textsc{ROAR}, an end-to-end method that generates robust CFEs using a training procedure with a custom adversarial objective that optimizes for the worst perturbation of the locally approximated decision boundary. 
Nguyen~et~al.~\cite{nguyen2022RBR} proposed \textsc{RBR}, a Bayesian-inspired method that models the data distribution with Gaussian kernels and accounts for data perturbations. 
Both previously mentioned methods only account for model changes due to data shifts and require significant effort in hyperparameter tuning.
In \cite{ferrario_robustness_2022}, robustness is achieved by modifying the training procedure of the model itself through counterfactual data augmentation, which may affect model performance as well as applicability to various practical settings.

The idea of post-hoc CFE generation was introduced with Dutta~et~al.~\cite{dutta_robust_2022}'s \robx method, designed for non-differentiable tree-based models, but in principle applicable to any black-box model. 
\robx explores the local neighborhood of a counterfactual by querying the underlying decision model over synthetic data points, thereby assessing local class variability.
However, \robx has notable limitations. It relies heavily on the assumption that the decision model is well calibrated. It also requires dataset-specific fine-tuning of the hyperparameters, which significantly affects the performance of the method. These hyperparameters include $\tau$ (the threshold for the \textit{counterfactual stability} metric) and \textit{variance} (representing the size of the local neighborhood used for sampling). In scenarios where the model can change, it is uncertain whether \robx quantifies the true local variability with high fidelity. 

Building upon these efforts, there have been some initial attempts to provide formal guarantees of the robustness achieved. Jiang~et~al.~\cite{jiang_formalising_2022} introduced a theoretical framework based on interval neural networks abstraction, which assumes identical architectures of the original and modified models, and defines robustness in the context of differences in weight values. 
Since such defined model changes do not correspond to actual differences in decision functions,  Hamman~et~al.\cite{hamman_robust_2023} relaxed the assumption of identical architectures and assumed a normal distribution of the input data to derive probabilistic guarantees.
A recent work by Marzari~et~al.~\cite{marzari2024rigorousprobabilisticguaranteesrobust} introduced AP$\Delta$S, a  method that certifies probabilistic guarantees of counterfactual robustness by sampling plausible model shifts, defined as model weight perturbations.  
All these methods are limited to neural networks and are not easily transferable to other types of models.
Furthermore, the assumed model shifts may be difficult to guarantee in practical settings.

In contrast, our \betaRCE method is applicable to any machine learning algorithm and can accommodate any type of expected model change, including data shift, changes in model architecture, and retraining with modified hyperparameters.  
It is a post-hoc approach to robustifying counterfactuals, allowing the user to choose the underlying CFE method that generates explanations with the desired properties. In addition, \betaRCE generates explanations that meet the definition of \deltaalphaRobustness (see Sec.~\ref{sec:method}), providing probabilistic guarantees. 
The method has only three, interpretable and easy-to-select hyperparameters that represent the expected probability of robustness. This makes our method more versatile and user-friendly than existing approaches.

\section{Method}
\label{sec:method}

In this section, we first formally define the robustness of the counterfactual to model change and provide a theoretical framework for its estimation. Later, we describe the algorithm that stems from these theoretical foundations, called \textsc{BetaRCE}, which provides robust counterfactuals with statistical guarantees.

\subsection{Defining robustness}

Consider a binary classification problem where a machine learning model $M: \mathcal{X} \rightarrow \mathcal{Y}$ assigns a binary label $y \in \mathcal{Y}=\{0,1\}$ to each instance $x \sim \mathcal{X}$. 
Given an input instance $x^{orig}$ and a prediction $M(x^{orig})$, a counterfactual explanation $x^{cf}$ is an instance close

to $x^{orig}$ for which the model reaches the opposite decision, i.e.~$M(x^{orig}) \neq M(x^{cf})$. 

Although using the counterfactual $x^{cf}$ instead of the original instance $x^{orig}$ causes the model $M$ to reverse its decision, even a small change in the model $M'$ can potentially invalidate the counterfactual, i.e. $M'(x^{orig}) = M'(x^{cf})$. This leads us to the notion of counterfactual robustness to model change.

\begin{definition}[Robust counterfactual]

 A counterfactual $x^{cf}$ explaining the prediction of a model $M$ is robust to its change to a model $M'$ if $x^{cf}$ is identically classified by both the original and the changed model: $M(x^{cf}) = M'(x^{cf})$.
\end{definition}  

In practice, it is impossible to construct a counterfactual that is robust to arbitrarily large changes in the model. 
However, it may be desirable to offer to the user an explanation that is robust to relatively not large model changes such as retraining with different random seed, making slight modifications of training data, or changing model hyperparameters.
We formally address these changes by defining the distribution of all possible models resulting from such changes\footnote{Specific model changes that we are considering are defined in Sec.~\ref{sec:experimental-setup}.}.

\begin{definition}[Space of admissible models]
  
    The space of admissible models $\mathcal{M}_M$ is the probabilistic distribution of all models that are the result of a complete retraining of the model $M$ using arbitrary settings from the predefined set of model changes.
\end{definition}

In the related literature, model changes are defined as perturbations of model parameters~\cite{dutta_robust_2022,xu2024generallyoccurringmodelchangerobust,jiang_formalising_2022}. Here, we use a more general definition of a model change, which is defined by the user, and can include changes in random seeds, model hyperparameters, or changes in the training data set. Since the space of admissible models is potentially infinite, we relax the notion of  counterfactual's robustness to only some (sufficiently high) proportion of possible admissible models.

\begin{definition}[\deltaRobust counterfactual]
    A counterfactual $x^{cf}$ is said to be \deltaRobust if and only if it is robust to change to a model randomly drawn from the given  admissible model space $\mathcal{M}_M$ with probability at least $\delta$.
    \begin{equation}
        P(M'(x^{cf}) = M(x^{cf})) \geq \delta \qquad M' \sim \mathcal{M}_M 
    \end{equation}
\end{definition}

Therefore, the goal of generating CFE robust to model change can be described as finding $x^{cf}$ such that it has the opposite class to the original one $M(x^{orig}) \neq M(x^{cf})$, and preserves it under model changes $M(x^{cf}) = M'(x^{cf})$ sampled from the space of allowed model changes  $M' \sim \mathcal{M}_M$ with a probability of at least $\delta$. 

\subsection{Estimating robustness}
\label{sec:estimating-rob}

The application of the above definition of counterfactual robustness requires the estimation of the parameter $\delta$, i.e. the true probability that the counterfactual is classified to the given class $M(x^{cf})$ by a model from $\mathcal{M}_M$. Note that unlike the classical probability $P(y|x)$, which estimates the prediction confidence of a \emph{single} model, the probability $\delta$ measures the decision preservation over a space of models. 

We adopted a Bayesian perspective on the estimation of $\delta$  to account for the estimation error. Since robustness to a given model change is a binary variable following a \textit{Bernoulli distribution}, we used the default prior for binary data, \textit{Beta distribution}, to model the confidence of the estimate $\hat{\delta}$.
Recall that \textit{Beta} is the conjugate prior of the \textit{Bernoulli distribution}~\cite{bayesianDataAnalysis}, which allows for much simpler computations. The adoption of the concept of credible interval, specifically its lower bound, leads to the following definition, which accounts for the estimation error.

\begin{definition}[\deltaalphaRobust counterfactual]
    \label{def:delta-alpha-robustness}
    A counterfactual $x^{cf}$ is said to be \deltaalphaRobust if and only if it is robust to change to a model randomly drawn from the admissible model space $\mathcal{M}_M$ with probability at least $\delta$ given the confidence level $\alpha$.
    \begin{equation}
    \label{eq:delta-alpha-robustness}
        P(\hat{\delta} > \delta) > \alpha
    \end{equation}
    where $\hat{\delta}$ follows the a posteriori distribution representing the uncertainty regarding the estimated probability of a binary random event $[M'(x^{cf}) = M(x^{cf})]$.
\end{definition}
In simple words, \deltaalphaRobust CFE at $\alpha$ confidence level has the probability of being robust of at least $\delta$. This is reminiscent of the classic definition of PAC learning \cite{Haussler1993}, where we obtain a counterfactual that is likely to be approximately ($\delta$) robust.

A practical procedure for verifying \deltaalphaRobustness for a given counterfactual $x^{cf}$ can be implemented using various statistical techniques, but in this work, we show that using simple bootstrap estimation of the parameter $\hat{\delta}$ is enough to obtain useful  results.
More concretely, we sample $k$ estimators from the admissible model space $M' \sim \mathcal{M}_M$ and apply them to obtain a set of predictions for the counterfactual $M'(x^{cf})$. 
The outcomes compared to the desired class $M(x^{cf})$ are used to update the noninformative Jeffreys  prior (\cite{bayesianDataAnalysis}) for $\hat{\delta}$ following \textit{Beta distribution}.
Finally, the verification of the condition described in Def.~\ref{def:delta-alpha-robustness} can be validated via checking the quantile of a posteriori \textit{Beta distribution}.  

\begin{equation}
    F^{-1}_{Beta}(1 - \alpha) \geq \delta 
\end{equation}  

Note that although the above procedure requires sampling and thus retraining of multiple models, the same small sample of models can be used to check all counterfactuals, greatly reducing the computational requirements as all the models can be trained and stored beforehand.
Similarly, the use of the conjugate \textit{Beta distribution} allows for very effective Bayesian updating of the posterior distribution given the observed data.
The pseudocode of the verification procedure is presented in Alg.~\ref{alg:pseudocode2}.

\begin{algorithm}[htbp]
\caption{Bootstrap verification of \deltaalphaRobustness}
\label{alg:pseudocode2}
\begin{algorithmic}[1]
    \Statex \textbf{Input}
    \Statex $x^{cf}, y^{cf}$ - counterfactual explanation and its desired class
    \Statex $\mathcal{M}$ - space of admissible models
    \Statex  $k$ - number of estimators
    \Statex \textbf{Procedure}
    \State $a,b \gets (0.5, 0.5)$  \Comment{\textcolor{algcomment}{Initialize the parameters of a priori beta distribution (noninformative Jeffreys)}}
    \For{$i \in 1..k $}
    \State $M' \gets \text{a sample from } \mathcal{M}_M$    
    \Statex \Comment{\textcolor{algcomment}{Update posterior distribution}}
    \If{$M'(x^{cf}) == y^{cf}$} 
    \State $a \gets a +1 $ 
     \Else 
         \State $b \gets b +1 $ 
      \EndIf
    \EndFor
    \State     \Return $F^{-1}_{Beta(a,b)}(1 - \alpha) \geq \delta $ \Comment{\textcolor{algcomment}{Check the condition for counterfactual robustness (Def.~\ref{def:delta-alpha-robustness})}}
\end{algorithmic}
\end{algorithm}

\begin{theorem}
\label{th:alg1}
    A counterfactual $x^{cf}$ positively verified by Alg.~\ref{alg:pseudocode2} meets the condition defined by Eq.~\ref{eq:delta-alpha-robustness} and therefore is \deltaalphaRobust.
\end{theorem}
The proof can be found in App.~\ref{app:proof2}\footnote{The full appendix is available online at: \url{https://ignacystepka.com/projects/betarce.html}}.

\subsection{\betaRCE - a post-hoc method for making CFEs \deltaalphaRobust}
Multiple counterfactual construction methods have been proposed in the literature, and according to various studies, the selection of the most appropriate counterfactual strongly depends on user preferences~\cite{Stepka_2024}.
Therefore, we present \betaRCE, a post-hoc approach that generates a \deltaalphaRobust counterfactual by making a small perturbation to the counterfactual $x^{cf}$ constructed by a method selected by the user (to meet his expectations regarding the selected evaluation measures).

\paragraph{Objective function}  
To define the objective function, we first introduce two auxiliary functions. A counterfactual $x^{cf}$ is said to be valid whenever the underlying model classifies it to a different class than the original example $x^{orig}$:
\begin{equation}
\label{eq:valid-cf}
    {valid}(x^{cf}, x^{orig}) = \mathbbm{1}_{M(x^{cf}) \neq M(x^{orig})}
\end{equation}

A counterfactual is \deltaalphaRobust whenever it positively passes the Bootstrap verification algorithm (Alg.~\ref{alg:pseudocode2}). Formally, we define the outcome of that verification procedure as an indicator function:
\begin{equation}
\label{eq:robust_term}
    {robust}(x) = \mathbbm{1}_{Alg.~\ref{alg:pseudocode2}(x) = true}
\end{equation}
Therefore, validity checks whether the counterfactual $x'$ has the desired class, and robustness checks \deltaalphaRobustness using the verification procedure described in Sec.~\ref{sec:estimating-rob}.

Using the introduced notation, we derive the objective optimized by \betaRCE:
\begin{equation}
    \label{eq:optimisation-goal}
    x^* = \operatorname*{argmin}_{x' \in \mathcal{X}} {d(x^{cf}, x')~~\text{s.t.}~~valid(x', x^{orig}) \land robust(x') }
\end{equation}
where $d(x^{cf}, x')$ is a distance from the original counterfactual $x^{cf}$ to its robust version. 

\paragraph{Optimization algorithm}  
The above objective formulation (Eq.~\ref{eq:optimisation-goal}) is non-convex and non-differentiable, so any  zero-order optimization algorithm could be used. Here we  chose \growingSpheres \cite{laugel_inverse_2017} as it is a simple and fast optimization method originally designed to find adversarial examples that closely resemble the original input while inducing sparse feature changes. It is therefore well suited to finding a robust counterfactual that closely resembles the original one, especially since it can directly optimize our objective Eq.~\ref{eq:optimisation-goal}. However, we acknowledge that many other zero-order optimization methods could be used to optimize that objective.

\growingSpheres performs the optimization in two main steps, each involving the generation of examples uniformly distributed in a sphere constructed around the given instance $x^{cf}$.
In the first step, the method searches for the largest radius of a sphere containing only non-robust or invalid instances.
The optimization is iterative, halving the original large radius of a sphere until a lower bound on the distance from $x^{cf}$ to both valid and robust counterfactuals is established.
In the second step, the method repeats the optimization process in the opposite direction, generating random instances in the iteratively growing sphere, but not closer than the estimated lower bound. 
The optimization process ends when the first valid and robust counterfactual is found. \growingSpheres has two hyperparameters: $\eta$, which denotes the initial sphere radius, and $n$, representing the number of sampled instances from a sphere.  We set these parameters to the default values from the original paper.  
The pseudocode of \betaRCE is depicted in Alg.~\ref{alg:rce-growsph}. The function $Sphere$ is a sampling procedure internally employed in \textsc{GrowingSpheres}, with arguments respectively denoting the center to sample around, the shorter radius, and the longer radius (sampling occurs between these two radii). We provide an intuitive visualization of all steps of the algorithm in App.~\ref{app:betarce-visualization}.

\begin{remark}
    Any CFE returned from Alg.~\ref{alg:rce-growsph} is \deltaalphaRobust and respects the Def.~\ref{def:delta-alpha-robustness} due to the fact of being verified via Alg.~\ref{alg:pseudocode2}. 
\end{remark}

\begin{algorithm}[htb]
\caption{\betaRCE}
\label{alg:rce-growsph}
\begin{algorithmic}[1]
    \Statex \textbf{Input}
        \Statex $x^{orig}, y^{orig}$ - input instance and its class
        \Statex $\delta$, $\alpha$, $\mathcal{M}_M$, $k$ - \betaRCE hyperparameters
        \Statex $\eta$, $n$ - \growingSpheres hyperparameters
        \Statex \textbf{Procedure}
        \Statex \Comment{\textcolor{algcomment}{Construction of the base CFE}}
        \State $x^{cf} \gets BaseCFE(M, x^{orig}, y^{orig})$ 
        \Statex \Comment{\textcolor{algcomment}{\growingSpheres warmup: estimate the lower bound on the distance to robust and valid CFEs}}
        \State $Z \sim_n Sphere(x^{cf}, 0, \eta)$ \Comment{\textcolor{algcomment}{Uniformly sample $n$ candidates}}
        \While{$ \{x' \in Z: valid(x') \land robust(x')\} \neq \emptyset$}
            \State $\eta \gets 0.5\eta$
            \State $Z \sim_n Sphere(x^{cf}, 0, \eta)$
        \EndWhile
        \Statex \Comment{\textcolor{algcomment}{\growingSpheres search: find closest, robust, and vaild CFEs}}
        \State $a_0 \gets \eta$, $a_1 \gets 2\eta$ 
        \State $Z \sim_n Sphere(x^{cf}, a_0, a_1)$ 
        \While{$ \{x' \in Z: valid(x') \land robust(x')\} == \emptyset$}
            \State $a_0 \gets a_1$, 
            \State $a_1 \gets a_1 + \eta$
            \State $Z \sim_n Sphere(x^{cf}, a_0, a_1)$ 
        \EndWhile
        \Statex \Comment{\textcolor{algcomment}{Return robust and valid CFE closest to the base CFE}}
        \State \Return $\min_{x' \in Z: valid(x') \land robust(x')}  d(x', x^{cf})$
\end{algorithmic}

\end{algorithm}

\paragraph{Hyperparameters}  
The optimization objective of \betaRCE requires three parameters: a lower bound on the robustness probability $\delta$; a credible interval confidence level $\alpha$; and the number of classification models $k$ used to estimate the robustness of the counterfactuals.
The parameters $\delta$ and $\alpha$ are directly related to the probabilistic guarantees and have a clear statistical interpretation. 
They can be chosen by the user according to the requirements of an application problem or, alternatively, with the use of conventional statistical heuristics. 
Since increasing the parameter $k$ simply leads to more confident robustness estimates, it should be chosen to meet the time efficiency requirements of a given application and with the desired robustness level in mind.

\begin{theorem}
    \label{th:max-delta}
    The maximum verifiable $\delta$ by Alg.~\ref{alg:pseudocode2} with $k$ estimators, given $\alpha > 0.5$ and initial a priori Beta distribution with $a = b ~\vert~ a,b \in \mathbb{Z}_+$ parameters,  can be computed via applying all $k$ Bayesian updates to the first parameter of the Beta distribution, and calculating the inverse CDF at $1-\alpha$:
    \begin{equation}
        \label{eq:parameters}
        \delta_{max} = F^{-1}_{Beta_{(a + k, b)}}(1-\alpha)
    \end{equation}   
\end{theorem}

\noindent
\emph{Proof sketch:}
In Theorem~\ref{th:max-delta} $a + k$ represents the most optimistic situation, where all $k$ estimators agree that the counterfactual is robust, resulting in the most extreme right-skewed beta distribution and the highest attainable $\delta$ with $k$ estimators.
A proof and detailed analysis of this formula can be found in App.~\ref{appendix:proof}.


\section{Experiments}

In this section, we first outline the experimental setup (Sec.~\ref{sec:experimental-setup}). Next, we empirically verify the applicability of our introduced theoretical framework (Sec.~\ref{sec:exp-betarce}) and perform a sensitivity analysis of \betaRCE hyperparameters (Sec.~\ref{sec:hyper}). Finally, we compare the performance of \betaRCE to other baseline methods (Sec.~\ref{sec:exp-comp}).

\subsection{Experimental setup}
\label{sec:experimental-setup}

We conducted experiments on six datasets: HELOC, Wine, Diabetes, Breast Cancer, Car evaluation, and Rice -- all commonly used in related studies. Each dataset varies in size and number of attributes, but all are numerical to meet the requirements of the methods used later. Brief characteristics of these datasets are presented in App.~\ref{app:sec:datasets}.  For reproducibility purposes, we open-source our code\footnote{\url{https://github.com/istepka/betarce}}.

We conducted experiments with three types of models: neural networks~\cite{Goodfellow-et-al-2016}, gradient boosted trees (LightGBM~\cite{ke2017lightgbm}) and logistic regression~\cite{ISL}. The results for the latter two are provided in App.~\ref{app:sec:comparative-analysis}. The neural network architecture comprised three layers, each consisting of 128 neurons and ReLU~\cite{nair2010rectified} activation functions. The training procedure utilized a binary cross-entropy loss function, Adam optimizer~\cite{adam} with 1e-3 learning rate, and early stopping with patience of 5 epochs. The models were trained for a maximum of 100 epochs, with data batches of size 128 The comprehensive list of hyperparameters can be found in App.~\ref{app:sec:models}.
As base CFE generation methods, we used two popular algorithms: \dice ~\cite{mothilal_explaining_2020} and \growingSpheres \cite{laugel_inverse_2017} which were briefly introduced in Sec.~\ref{sec:rw-robustness}.

To simulate model change scenarios, we designed the following three types of experiments with varying admissible model spaces:

\begin{itemize}
    \item \emph{Architecture} -- the training data remains unchanged, but we vary the parameters of the neural classifier in terms of (1) layers: 3-5, and (2) number of neurons per layer: 64-256.
    \item \emph{Bootstrap} -- the architecture of the retrained model remains the same,  but its training data is modified by performing bootstrap, i.e., the same number of data points is sampled with replacement from the original dataset.
    \item  \emph{Seed} -- both the data and model parameters remain unchanged. The difference in models arises from using different initial random states, which  impacts the decision boundary \cite{randomSeeds}.
\end{itemize}
For each dataset, experiment type, and method setting, we  calculated metrics in a cross-validation setting (3CV) where for each fold we sampled 30 random data points and 30 random retrained models $M'$, resulting in $3 \times 30 \times 30 = 2700$ examples per dataset-experiment-method configuration.

In all experiments, we set the confidence level to $\alpha = 0.9$. In the first experiment (Sec.~\ref{sec:exp-betarce}), we opted to utilize credible intervals to demonstrate that both lower and upper bounds on the robustness hold in practice. 
Later, in the comparative analysis (Sec.~\ref{sec:exp-comp}) we only employed the lower bound, following the exact definition from Def.~\ref{def:delta-alpha-robustness}. 
Since $\alpha = 0.9$ was used for credible intervals, the lower bound was set to $0.95$ for consistency (as it only uses the left tail).
With this confidence level, we used the introduced formula (Eq.~\ref{eq:parameters}) to select the number of estimators ($k = 32$) required to achieve the highest $\delta$ (i.e., $\delta = 0.9$) investigated in the experiments.

To verify robustness,  we compute the \textit{Empirical Robustness} metric representing the average validity of CFEs, originally generated for the model $M$, when evaluated with a changed model $M'$.
\begin{equation}
\label{eq:emp-rob}
    \text{Empirical Robustness} = \frac{1}{N}\sum_{i=1}^{N} \mathbbm{1}_{M(x^{cf}_i) = M^{'}(x^{cf}_i)}
\end{equation}  
where $N$ is the number of $x^{orig}$ examples used in a given experiment. The changed model $M'$ is drawn by uniformly sampling the space of admissible model changes $\mathcal{M}_M$.  For instance, in the \emph{Architecture} experiments, each model $M'$ uses a randomly sampled set of architectural parameters from the space of admissible model changes.

\begin{figure*}[htbp]
        \includegraphics[width=\textwidth]{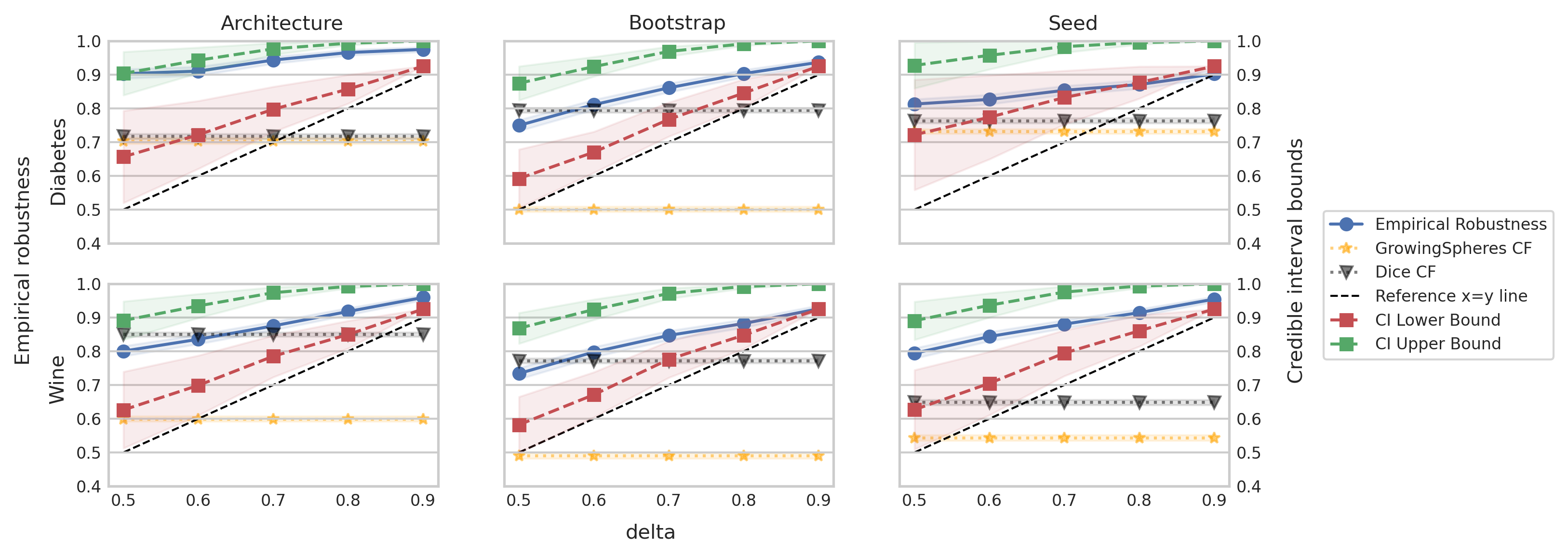}
        \caption{The average \textcolor{fig2_blue}{Empirical Robustness} of counterfactuals generated by \betaRCE at $\alpha = 90\%$, with \growingSpheres generating base CFEs. \textcolor{fig2_red}{Red} and \textcolor{fig2_green}{green} dashed lines are average \textcolor{fig2_red}{lower} and \textcolor{fig2_green}{upper} bounds of estimated $\alpha$ credible intervals (CI). The shaded areas in the back represent standard deviations. Horizontal \textcolor{fig2_yellow}{yellow} and \textcolor{fig2_gray}{gray} lines show the base robustness of CFEs obtained with \textcolor{fig2_gray}{\dice} and \textcolor{fig2_yellow}{\growingSpheres} without \betaRCE applied. The plots with all four datasets and two \betaRCE base CFE generation methods (\dice and \growingSpheres) are available in App.~\ref{app:sec:cred-interv-the}.}
        
        \label{fig:empirical-validation}
\end{figure*}

\subsection{Validation of \betaRCE theoretical framework}
\label{sec:exp-betarce}

In the first experiment, we validated whether the theoretical framework, from which we derived \textsc{BetaRCE}, holds in considered experimental settings. To achieve this, we opted not only to use the lower bound from our \deltaalphaRobustness definition (Def.~\ref{def:delta-alpha-robustness}), but also calculate the upper bound to extract more information from the method. In Fig.~\ref{fig:empirical-validation} we present the results for two datasets, Diabetes and Wine. 
The base CFEs were generated using \textsc{GrowingSpheres}.

In Fig.~\ref{fig:empirical-validation}, the blue line with spherical points illustrates the \textit{Empirical robustness} obtained at various levels of $\delta$. The robustness line lies between the green and red dashed lines with square-shaped points, representing the average lower and upper bounds of the credible interval. This shows that the \deltaalphaRobustness estimations are indeed valid and hold in practice. In addition, the plot also includes the results of two baselines, \growingSpheres and \dice (yellow and black horizontal dotted lines), indicating a consistent improvement in robustness over both of those baselines.

Only for the Diabetes dataset and Seed scenario, we notice that the empirical robustness at $\delta = 0.9$ is slightly lower than the estimated average lower bound (however still above the $\delta$). We attribute this discrepancy to the $90\%$ confidence level of the method, which acknowledges that the estimate may occasionally be incorrect.

\subsection{Hyperparameter sensitivity analysis}
\label{sec:hyper}

\begin{figure}[tbp]
    \centering
    \includegraphics[width=0.58\textwidth]{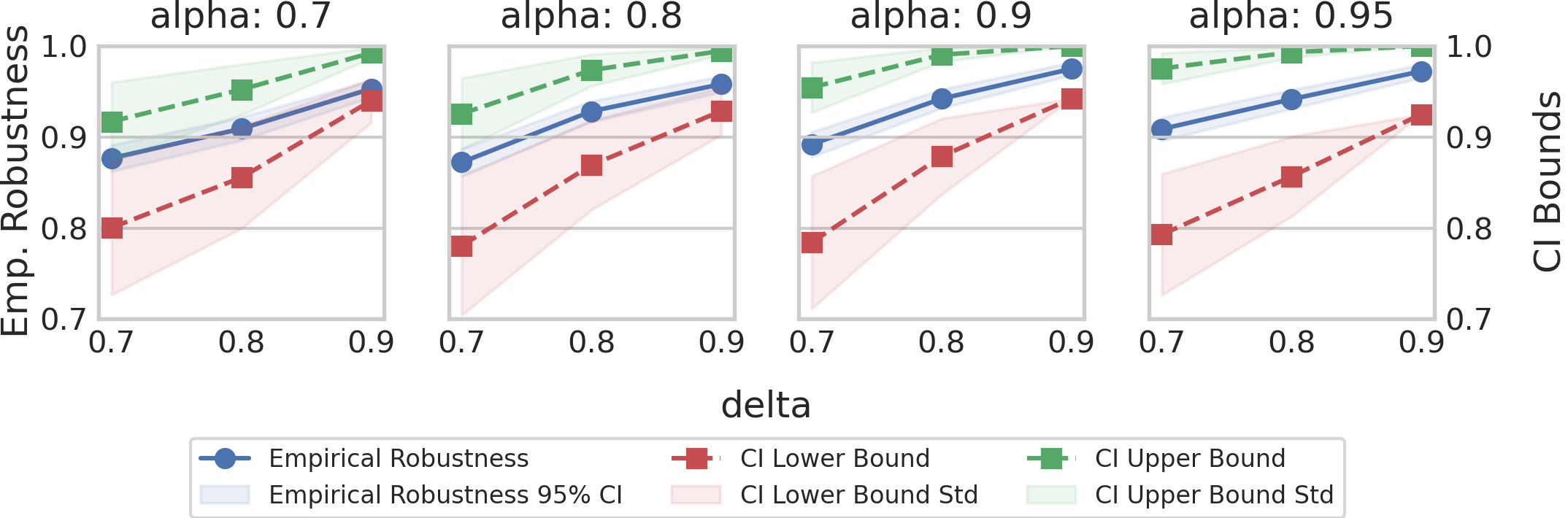}
    \hfill
    \includegraphics[width=0.4\textwidth]{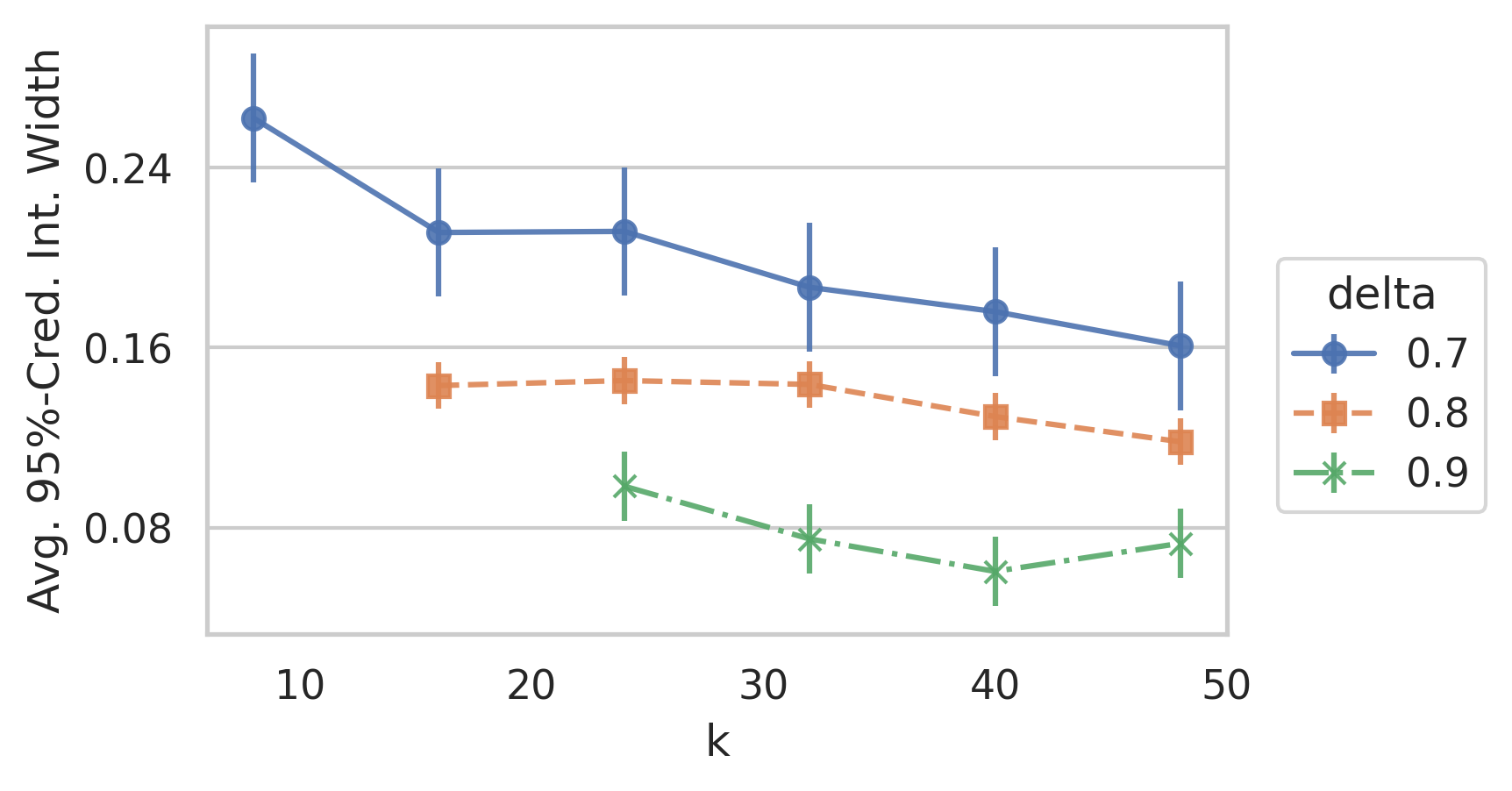}
    \caption{Left: The average empirical robustness computed at various confidence ($\alpha$) and robustness ($\delta$) levels on the HELOC dataset, with $k$ set to $32$. Right: The impact of the number of $k$ estimators on the width of $90\%$ credible interval, computed on the HELOC dataset.}
    \label{fig:hparams-sens}
\end{figure}

In this section, we present the experiments assessing the impact of using different confidence values $\alpha$  and different numbers of estimators $k$. 
Selected results for varying  $\alpha$ and  $k$ are shown in Fig~\ref{fig:hparams-sens}. The hyperparameter sensitivity analysis is extended in App.~\ref{app:sec:hparams-confidence} and~\ref{app:sec:hparams-k}. 

The results for different $\alpha$ values indicate that as it increases, the empirical robustness tends to move further away from the lower bound. 
This suggests that higher $\alpha$ values yield more confident estimates with a lower likelihood of violating the lower bound.

Regarding the number of estimators $k$, we observed that increasing this parameter leads to narrower credible interval widths. 
It is anticipated because higher values of $k$ result in a greater diversity of estimated Beta distribution shapes. 
Specifically, the number of possible Beta distributions is constrained by the combinations of all attainable $a$ and $b$ values, which increases with a larger $k$.

\subsection{Comparative study with other methods}
\label{sec:exp-comp}

In the final experiment, we empirically compare the performance of \betaRCE with several baselines, each representing a different type of method and aimed at achieving different CFE properties (see Sec.~\ref{sec:rw-robustness}). 
The first type of baselines includes standard CFE generation methods: \dice~\cite{mothilal_explaining_2020}, \face~\cite{poyiadzi_face_2020}, and \growingSpheres~\cite{laugel_inverse_2017}, which do not claim to guarantee any robustness to model change. 
The next type are end-to-end CFE generation methods, explicitly generating robust CFEs: \rbr~\cite{nguyen2022RBR} and \roar~\cite{upadhyay_towards_2021}. 
Finally, we also include RobX~\cite{dutta_robust_2022}; a method most similar to ours, as it also operates on top of a base CFE generation method in a post-hoc fashion. 
The above-mentioned robust baselines aim to increase overall robustness of CFEs by moving the CFE to a more robust region in the feature space. Notably, they require manual selection of hyperparameters in a dataset-specific manner. 
Therefore, to ensure a fair comparison, for \rbr and \roar, we performed a hyperparameter search to find the most promising settings. 
In the case of \textsc{RobX}, we chose four combinations of the most relevant hyperparameters, $\tau$ and variance, where $\tau$ values were selected using the histogram technique described by the authors, and variances were chosen to vary significantly (specifically $0.1$ and $0.01$). 
The comprehensive description of the hyperparameter selection process is described in App.~\ref{app:sec:hparams-baselines}. 
Note, that this comparison is not straightforward since each method is built on different premises and has multiple parameters to tune, making comparison to our probabilistic goals challenging.

We compare the above-listed baselines to two variants of our method ($\delta = 0.8$, $\delta = 0.9$) on the following quality measures.
\textit{Empirical robustness}, already defined in Eq.~\ref{eq:emp-rob}, quantifying the degree of robustness to model change. 
\textit{Proximity} -- the distance between the counterfactual and the original example: 
\begin{equation}
    Proximity = d(x^{cf}, x^{orig})
\end{equation}
In the experiments we calculate \textit{Proximity} in two variants, $L_1$ and $L_2$, corresponding to Manhattan and Euclidean distance, respectively.

\textit{Plausibility} -- the average distance to $n$ closest neighbors of the CFE in the training set:
\begin{equation}
    Plausibility = \frac{1}{n}\sum_{i=1}^n d(x^{cf}, x_i)
\end{equation}

\textit{Distance to Base} -- a metric calculated only for the post-hoc methods, measuring the $L_1$ distance between the robust CFE and the base one:
\begin{equation}
    Distance~to~Base = d(x^{rcf}, x^{bcf})
\end{equation}


\begin{table*}[ht]
    \centering
    \scriptsize
    \setlength{\tabcolsep}{2pt}
    \caption{Comparative study results. RobX and \betaRCE were using \growingSpheres as the base counterfactual explainer. Parameters used in a given method are listed next to this method's name; for \robx these are $\tau$ and variance, while for \betaRCE -- $\delta$ (we omit $\alpha$ as it is fixed at $0.95$). The values in each cell represent the mean $\pm$ standard error. The column \textbf{Type} sorts the methods by categories. The abbreviations Btsr and Arch used next to \betaRCE in the \textbf{Type} column stand for Bootstrap and Architecture, respectively. }
    \label{tab:comparative-analysis}
    \begin{tabular}{ccl|cccc|ccc}
    \toprule
        \multirow{2}{*}{\textbf{Dataset}} & \multirow{2}{*}{\textbf{Type}} & \multirow{2}{*}{\textbf{Method}} & \multicolumn{4}{c}{\textbf{Metrics}} & \multicolumn{3}{c}{\textbf{Empirical Robustness}} \\  \cmidrule(r){4-7} \cmidrule(r){8-10} 
         &  &  & Dist. to Base $\downarrow$ & Proximity $L_1$ $\downarrow$ & Proximity $L_2$ $\downarrow$ & Plausibility $\downarrow$ & Architecture $\uparrow$ & Bootstrap $\uparrow$ & Seed $\uparrow$ \\ \midrule
        \multirow{13}{*}{\rotatebox[origin=c]{90}{Diabetes}} &  \multirow{3}{*}{Standard CFEs} & \dice & - & 1.002 $\pm$ 0.001 & $0.645 \pm 0.001$ & 0.499 $\pm$ 0.001 & 0.916 $\pm$ 0.002 & 0.889 $\pm$ 0.003 & 0.916 $\pm$ 0.003 \\ 
         & & \growingSpheres & - & 0.800 $\pm$ 0.001 & $0.345 \pm 0.001$ & 0.358 $\pm$ 0.001 & 0.939 $\pm$ 0.003 & 0.852 $\pm$ 0.003 & 0.853 $\pm$ 0.003 \\ 
         & & \face & - & $0.880 \pm 0.001$ & $0.401 \pm 0.001$ & $0.248 \pm 0.001$ &  $0.869 \pm 0.005$  & $0.694 \pm 0.006$ & $0.721 \pm 0.006$ \\ \cline{2-10}
         & \multirow{2}{*}{Robust e2e}  & \rbr & - & $0.714 \pm 0.001$ & $0.339 \pm 0.001$ & $0.319 \pm 0.001$ & $0.618 \pm 0.006$ & $0.617 \pm 0.005$ & $0.576 \pm 0.005$ \\ 
         & & \roar & - & $10.887 \pm 0.001$ & $4.703 \pm 0.001$ & $4.424 \pm 0.001$ & $0.415 \pm 0.005$  & $0.417 \pm 0.005$  & $0.408 \pm 0.005$ \\ \cline{2-10}
         & \multirow{2}{*}{Robust post-hoc} & RobX(0.5,0.1) & 1.224 $\pm$ 0.001 & 1.432 $\pm$ 0.001 & $0.651 \pm 0.001$ & 0.324 $\pm$ 0.001 & 0.998 $\pm$ 0.001 & 0.947 $\pm$ 0.004 & 0.969 $\pm$ 0.004 \\ 
         & & RobX(0.5,0.01) & 0.429 $\pm$ 0.001 & 0.748 $\pm$ 0.001 & $0.339 \pm 0.001$& 0.289 $\pm$ 0.001 & 0.970 $\pm$ 0.003 & 0.872 $\pm$ 0.006 & 0.922 $\pm$ 0.005 \\ 
         & \multirow{2}{*}{\betaRCE Arch}  & \betaRCE(0.8) & 0.488 $\pm$ 0.013 & 0.870 $\pm$ 0.013 & $0.382 \pm 0.005$ & 0.372 $\pm$ 0.004 & 0.966 $\pm$ 0.004 & - & - \\ 
         &  & \betaRCE(0.9) & 0.607 $\pm$ 0.014 & 0.953 $\pm$ 0.013 & $0.420 \pm 0.006$ & 0.378 $\pm$ 0.004 & 0.975 $\pm$ 0.004 & - & - \\ 
         & \multirow{2}{*}{\betaRCE Btsr}  & \betaRCE(0.8) & 0.445 $\pm$ 0.006 & 0.840 $\pm$ 0.008 & $0.359 \pm 0.003$ & 0.359 $\pm$ 0.002 & - & 0.903 $\pm$ 0.006 & - \\ 
         &  & \betaRCE(0.9) & 0.583 $\pm$ 0.007 & 0.949 $\pm$ 0.008 & $0.407 \pm 0.003$ & 0.369 $\pm$ 0.002 & - & 0.937 $\pm$ 0.005 & - \\ 
         & \multirow{2}{*}{\betaRCE Seed}  & \betaRCE(0.8) & 0.247 $\pm$ 0.006 & 0.813 $\pm$ 0.008 & $0.346 \pm 0.003$ & 0.350 $\pm$ 0.002 & - & - & 0.871 $\pm$ 0.006 \\ 
         &  & \betaRCE(0.9) & 0.315 $\pm$ 0.006 & 0.862 $\pm$ 0.008 & $0.367 \pm 0.003$ & 0.353 $\pm$ 0.002 & - & - & 0.902 $\pm$ 0.006 \\ \midrule
         \multirow{13}{*}{\rotatebox[origin=c]{90}{HELOC}} & \multirow{3}{*}{Standard CFEs} & \dice & - & 3.190 $\pm$ 0.004 & $1.163 \pm 0.001$ & 1.003 $\pm$ 0.001 & 0.912 $\pm$ 0.002 & 0.781 $\pm$ 0.004 & 0.815 $\pm$ 0.003 \\ 
         &  & \growingSpheres & - & 2.782 $\pm$ 0.003 & $0.717 \pm 0.001$ & 0.773 $\pm$ 0.001 & 0.862 $\pm$ 0.003 & 0.794 $\pm$ 0.003 & 0.752 $\pm$ 0.004 \\ \ 
         & & \face & - & $2.254 \pm 0.001$ & $0.659 \pm 0.001$ & $0.441 \pm 0.001$ & $0.829 \pm 0.005$ & $0.717 \pm 0.006$ &  $ 0.717 \pm 0.006$ \\  \cline{2-10}
         & \multirow{2}{*}{Robust e2e} & \rbr & - & $1.682 \pm 0.001$ & $0.505 \pm 0.001$ & $0.468 \pm 0.001$ & $0.754 \pm 0.005$ & $0.690 \pm 0.005$ & $0.706 \pm 0.005$ \\  
         & & \roar & - & $19.803 \pm 0.001$ & $5.427 \pm 0.001$ & $4.786 \pm 0.001$ & $0.591 \pm 0.005$ & $0.51 \pm 0.005$ &  $0.588 \pm 0.005$  \\ \cline{2-10}
         & \multirow{2}{*}{Robust post-hoc} & RobX(0.5,0.01) & 1.145 $\pm$ 0.002 & 2.341 $\pm$ 0.001 & $0.636 \pm 0.001$ & 0.598 $\pm$ 0.001 & 0.939 $\pm$ 0.005 & 0.814 $\pm$ 0.007 & 0.890 $\pm$ 0.006 \\ 
         &  & RobX(0.5,0.1) & 3.548 $\pm$ 0.005 & 3.938 $\pm$ 0.004 & $1.144 \pm 0.001$ & 0.575 $\pm$ 0.001 & 0.991 $\pm$ 0.002 & 0.957 $\pm$ 0.004 & 0.955 $\pm$ 0.005 \\ \ 
         & \multirow{2}{*}{\betaRCE Arch}  & \betaRCE(0.8) & 1.538 $\pm$ 0.049 & 2.912 $\pm$ 0.053 & $0.749 \pm 0.014$& 0.802 $\pm$ 0.011 & 0.904 $\pm$ 0.006 & - & - \\ 
         &  & \betaRCE(0.9) & 1.697 $\pm$ 0.031 & 2.927 $\pm$ 0.036 & $0.753 \pm 0.009$ & 0.783 $\pm$ 0.007 & 0.935 $\pm$ 0.005 & - & - \\ 
         & \multirow{2}{*}{\betaRCE Btsr} & \betaRCE(0.8) & 2.288 $\pm$ 0.041 & 3.451 $\pm$ 0.044 & $0.889 \pm 0.011$ & 0.859 $\pm$ 0.008 & - & 0.833 $\pm$ 0.007 & - \\ 
         &  & \betaRCE(0.9) & 3.547 $\pm$ 0.071 & 4.501 $\pm$ 0.073 & $1.156 \pm 0.019$ & 1.044 $\pm$ 0.015 & - & 0.880 $\pm$ 0.006 & - \\ 
         & \multirow{2}{*}{\betaRCE Seed}  & \betaRCE(0.8) & 1.420 $\pm$ 0.021 & 2.526 $\pm$ 0.028 & $0.653 \pm 0.007$ & 0.726 $\pm$ 0.004 & - & - & 0.826 $\pm$ 0.007 \\ 
         &  & \betaRCE(0.9) & 1.927 $\pm$ 0.030 & 2.906 $\pm$ 0.035 & $0.750 \pm 0.009$ & 0.776 $\pm$ 0.006 & - & - & 0.902 $\pm$ 0.006 \\ 
         \bottomrule
    \end{tabular}
\end{table*}

The detailed results for Diabetes and HELOC datasets are presented in Tab.~\ref{tab:comparative-analysis} and the results for the remaining datasets are in App.~\ref{app:sec:comparative-analysis}.
The results indicate that \betaRCE outperforms \robx in terms of the \textit{Distance to Base} metric. 
Moreover, both the \textit{Proximity} and \textit{Plausibility} metrics are either preserved or slightly improved, validating our goal of not deteriorating the properties of the base counterfactual.
This suggests that \betaRCE is effective at finding CFEs
with a user-defined robustness level
that are not significantly different from the base counterfactual, preserving its original properties and aligning with one of our primary motivations.

Regarding \textit{Empirical Robustness}, we observe that \robx often achieves a higher score than \textsc{BetaRCE}. 
However, this comes at a cost, as indicated by the fact that the \textit{Distance to Base} metric for \robx is frequently worse than for \textsc{BetaRCE}. 
In some cases, \robx with well-selected hyperparameters achieves nearly perfect robustness but significantly deteriorates all other metrics (e.g., more than doubles the \textit{Distance to Base} in comparison to \betaRCE). 
Recall that the goal of \betaRCE is to have robustness greater than a specified lower bound while preserving as much of the base counterfactual as possible, and these goals have been achieved. 
The only setting where the lower bound is violated by the computed \textit{Empirical Robustness} is the Bootstrap-HELOC combination for \betaRCE at $\delta=0.9$, where that value is slightly lower than the expected lower bound. 
However, the difference is negligible, and we attribute it to \textsc{BetaRCE}'s confidence level parameter.

When it comes to the comparison to other robust (end-to-end) baselines, the analysis suggests that even though we performed a hyperparameter search on them, they do not perform well in the investigated scenarios.  
The \textit{Empirical Robustness} achieved by both \rbr and \roar is substantially outperformed by \betaRCE in all tested scenarios, with differences sometimes exceeding 50 p.p.

Finally, let us notice that \robx parameters do not generalize well across different types of model changes, indicating that dataset-specific tuning (as described by authors) is insufficient to achieve high robustness across the board. 
In contrast, \betaRCE requires no hyperparameter tuning, and its three parameters are interpretable and easy to select based on decision-maker expectations.

We also conducted additional experiments to verify the robustness of \betaRCE to a misdefined space of admissible model changes. 
The details are presented in App.~\ref{app:sec:gen}.
In summary, although the theoretical probabilistic bounds do not hold for out-of-distribution model changes, they are often practically satisfied by \betaRCE.
The results reveal good transferability of CFEs generated for Architecture and Bootstrap scenarios to Seed, as well as Seed CFEs to Architecture. The transfer to the Bootstrap scenario seems to be more difficult, sometimes resulting in a drop of \textit{Empirical Robustness} as high as 4 p.p.


\section{Final Remarks}

In this paper, we introduced \textsc{BetaRCE}, a novel post-hoc, model-agnostic method for generating robust counterfactual explanations. 
This method is the first proposal that provides probabilistic guarantees on the robustness of CFEs to model change in a model-agnostic fashion. 
Moreover, its parametrization is tied to probabilistic expectations, enabling users to select the expected robustness in a more natural way compared to other approaches requiring extensive tuning of not-interpretable hyperparameters.

The experiments have confirmed that the introduced probabilistic bounds, estimated in practice with a set of models, hold for the \textit{Empirical Robustness}, as demonstrated across three types of model changes. 

Notably, counterfactuals generated by \betaRCE not only improved robustness, but also were closer to the base CFEs, retaining their properties better than CFEs generated by existing methods, while achieving a similar level of robustness.

Several avenues for future research emerge from our study. First, our approach relies on a fairly simple bootstrap estimation method to model the second-order probability distribution of robustness. While our experiments demonstrated the utility of this approach, exploring more sophisticated estimation methods could yield interesting insights.
Additionally, the question of precisely quantifying model changes remains open. Specifically, distinguishing between slight, moderate, and substantial model changes -- and determining appropriate responses to each -- requires further investigation. Defining the space of admissible models more rigorously could enhance the precision of robustness estimates.
Another potential direction is the use of auxiliary models to predict beta distribution parameters, which might replace querying a set of estimators, consequently accelerating optimization. Finally, examining alternative optimization algorithms could provide further opportunities to enhance the efficiency of \textsc{BetaRCE}.

\section*{Acknowledgments}
This research was funded in whole by the National Science Centre, Poland (Grant No. 2023/51/B/ST6/00545). For the purpose of Open Access, the author has applied a CC-BY public copyright license to any Author Accepted Manuscript (AAM) version arising from this submission. 

\bibliographystyle{unsrt}  
\bibliography{mybibfile}  

\bigskip
\appendix

\section{Reproducibility}
\label{sec:app-reproducibility}

To access the full version of the appendix, navigate to:   
\url{https://ignacystepka.com/projects/betarce.html}

\subsection{Code Availability}
\label{sec:app-code}
To ensure reproducibility and enable further experimentation with \textsc{BetaRCE}, we make the source code publicly available on GitHub: \url{https://github.com/istepka/betarce}.

\subsection{Datasets}
\label{sec:app-datasets}
Wine, Breast Cancer, Car eval, and Rice datasets were obtained from \url{https://archive.ics.uci.edu/}. Diabetes dataset was sourced from \url{https://www.kaggle.com/datasets/mathchi/diabetes-data-set}. HELOC (fico) dataset from \url{https://community.fico.com/s/explainable-machine-learning-challenge?tabset-158d9=d157e}.

\section{Proof of Theorem \ref{th:alg1}}
\label{app:proof2}
Assume that $\hat{\delta}$ follows a priori Beta distribution with parameters $a, b$, therefore the a priori probability of $P(\hat{\delta})$ has the following probability density function:
\begin{equation}
\begin{aligned}
f(\hat{\delta};a ,b )&={\frac {1}{\mathrm {B} (a ,b )}}\hat{\delta}^{a -1}(1-\hat{\delta})^{b -1}\\
&\propto \hat{\delta}^{a -1}(1-\hat{\delta})^{b -1}
\end{aligned}
\end{equation}
where the beta function $\mathrm {B}$ is a normalization constant.

Algorithm~\ref{alg:pseudocode2} samples $k$ random variables $X_i=\mathbbm{1}_{M'(x^{cf}) == y^{cf}}$ from the space of admissible model changes. Applying the Bayes theorem, we obtain the following a posteriori distribution:

\begin{equation}
\begin{aligned}
f(\hat{\delta}|\mathbf {X} ,a ,b )&\propto  f(\mathbf {X}|\hat{\delta};a ,b ) f(\hat{\delta};a ,b ) \\
&\propto \left(\prod _{i=1}^{k}\hat{\delta}^{x_{i}}(1-\hat{\delta})^{1-x_{i}}\right)\hat{\delta}^{a -1}(1-\hat{\delta})^{b -1}\\&\propto\hat{\delta}^{\sum x_{i}+a -1}(1-\hat{\delta})^{k-\sum x_{i}+b -1}
\end{aligned}
\end{equation}
Using $z = \sum_{i=1}^k x_{i}$ to denote the number of times the counterfactual was robust for the sampled model, we obtain
\begin{equation}
\label{eq:app:1}
\begin{aligned}
f(\hat{\delta}|\mathbf {X}; a ,b )&\propto\hat{\delta}^{z+a -1}(1-\hat{\delta})^{k-z+b -1}\\
&=f(\hat{\delta};a +z,b + (k-z) )\\
\end{aligned}
\end{equation}
which is exactly the Beta distribution.
Note that Algorithm~\ref{alg:pseudocode2} adds 1 to $a$ every time the counterfactual is robust to the sampled model, so effectively adds $z$ to $a$ during the entire execution. Similarly, $k-z$ is added to $b$. Therefore, Algorithm~\ref{alg:pseudocode2} estimates the posterior distribution of $P(\hat{\delta})$.

According to Definition~\ref{def:delta-alpha-robustness}, \deltaalphaRobust counterfactual satisfies the following condition:
$$P(\hat{\delta}>\delta) > \alpha$$
Applying Eq.~\ref{eq:app:1}, we obtain:
\begin{equation}
\begin{aligned}
P(\hat{\delta}>\delta) &= \int_{\delta}^1 f(\hat{\delta};a +z,b + (k-z) ) \, d\hat{\delta}    \\
&=   1- F_{Beta}(\delta) 
\end{aligned}
\end{equation}
where $F_{Beta}$ is the cumulative distribution function of Beta distribution.
\begin{equation}
\begin{aligned}
    P(\hat{\delta}>\delta) > \alpha &\Rightarrow 1- F_{Beta}(\delta) > \alpha  \\
    &\Rightarrow F_{Beta}(\delta) < 1- \alpha  \\
    &\Rightarrow F^{-1}_{Beta}(1- \alpha) >\delta   \\
\end{aligned}
\end{equation}
which is consistent with line 10 of Algorithm 1.

\section{Proof of Theorem~\ref{th:max-delta}}
\label{appendix:proof}
In this section, we provide a proof of the Theorem~\ref{th:max-delta}. 
  
\subsection{Background}
The cumulative distribution function (CDF) of a probability distribution is a function describing the following relationship:
\begin{equation}
F(x) = P(X \leq x) = u
\end{equation}
where $X$ is a random variable, $x$ is a real number, and $u$ is a probability between 0 and 1.
The inverse cumulative distribution function (inverse CDF), also known as the quantile function, is used to find the value $x$ for a given probability $u$:
\begin{equation}
F^{-1}(u) = x
\end{equation}
This function returns the value $x$ such that the probability of the random variable $X$ being less than or equal to $x$ is $u$.

\subsection{Proof}

Let:
\begin{itemize}
    \item $n + m = k$ and $n > m$, where $n,m,k \in \mathbb{Z}_+$
    \item $a  = b$ where $a,b \in \mathbb{R}_+$ be a priori parameters of the Beta distribution: $Beta(a, b)$. 
\end{itemize}

We begin by stating Lemma~\ref{le:Fdiff} which asserts that for any $\alpha$ greater than $0.5$, the CDF of a Beta distribution will always be greater if its first parameter is greater than the second one. 

\begin{lemma}
\label{le:Fdiff} 
    \begin{equation}
    \underset{x \in (0.5, 1]}{\forall} F_{Beta{(a + n, b + m)}}(x) > F_{Beta{(a + m, b + n)}}(x)
    \end{equation}
\end{lemma}  

In order to prove Lemma~\ref{le:Fdiff}, we first simplify the underlying equations in a following way:

\begin{equation}
\scriptsize
    \begin{split}
        F_{Beta{(a + n, b + m)}}(x) > F_{Beta{(a + m, b + n)}}(x) \\
        = \frac{B(x; a+n, b+m)}{B(a+n, b+m)} > \frac{B(x; a+m, b+n)}{B(a+m, b+n)}~~\vert~\text{\footnotesize Beta func. symmetry}\\
        = B(x; a+n, b+m) > B(x; a+m, b+n) \\ 
        = \int_0^x t^{a+n-1}(1-t)^{b+m-1} dt > \int_0^x t^{a+m-1}(1-t)^{b+n-1} dt 
    \end{split}
\end{equation}

WLOG, for simplicity of notation, we can assume $a = b = 1$. Therefore, the equation above simplifies to:
\begin{equation}
    \begin{split}
        \int_0^x t^{n}(1-t)^{m} dt > \int_0^x t^{m}(1-t)^{n} dt 
    \end{split}
\end{equation}

From that, it is sufficient to show that the left-hand function is strictly greater than the right-hand function in the integrated domain.

\begin{lemma}
\label{le:FalwaysG}
    \begin{equation}
        \underset{t \in (0.5, 1]}{\forall} \underset{n,m \in \mathbb{Z}_+, n>m}{\forall}  t^{n}(1-t)^{m} >  t^{m}(1-t)^{n}
    \end{equation}
\end{lemma}  

Proof: Lemma~\ref{le:FalwaysG} can be proven via simple arithmetic manipulations:

\begin{equation}
    \begin{split}
        t^{n}(1-t)^{m} >  t^{m}(1-t)^{n} \\ 
        = t^{n-m}(1-t)^m > (1-t)^n \\ 
        = t^{n-m} > (1-t)^{n-m} \\ 
        \implies t > (1-t) \\
        = t > 0.5 \\
        \text{QED}
    \end{split}
\end{equation}
This completes the proof of Lemma~\ref{le:FalwaysG}

The result of Lemma~\ref{le:FalwaysG}, $t > 0.5$, finishes the proof for Lemma~\ref{le:Fdiff}, which in turn  proves Theorem~\ref{th:max-delta}, because the highest attainable value of CDF is at $1-\alpha$, where $\alpha > 0.5$

\section{Examining \betaRCE parameters in detail}

In the main paper, we briefly outlined the relationship between parameters in \betaRCE. Remember, \betaRCE relies on three internal parameters that impact its performance: $\delta$, representing the lower bound for the probability of robustness; $\alpha$, indicating the method's confidence level; and $k$, denoting the number of estimators. Their interplay is defined by the following equation, also featured in the paper:

\begin{equation}
\delta_{max} = F^{-1}_{Beta{(a + k, b)}}(1 - \alpha)
\end{equation}

This equation offers an intuitive approach to determining the parameters based on practical application requirements. The maximum achievable $\delta$ (and consequently $(\delta,\alpha)$-robustness) is constrained by the number of estimators $k$ and the selected confidence level $\alpha$.

The interpretation of this equation is straightforward: $F^{-1}(1 - \alpha)$ identifies the lower bound of robustness at $1 - \alpha$. The inverse Cumulative Distribution Function ($F^{-1}$) is derived from the estimated Beta distribution $Beta_{(a + k, b)}$, with $a$ and $b$ representing default priors of the distribution. Here, $k$ is added to the $a$ parameter of the distribution, as it contributes to the right-skewness of the distribution.

To provide a clearer understanding, below we present a visual representation of how parameters in the Beta distribution influence its shape:

\begin{figure}[h]
    \centering
    \includegraphics[width=0.7\textwidth]{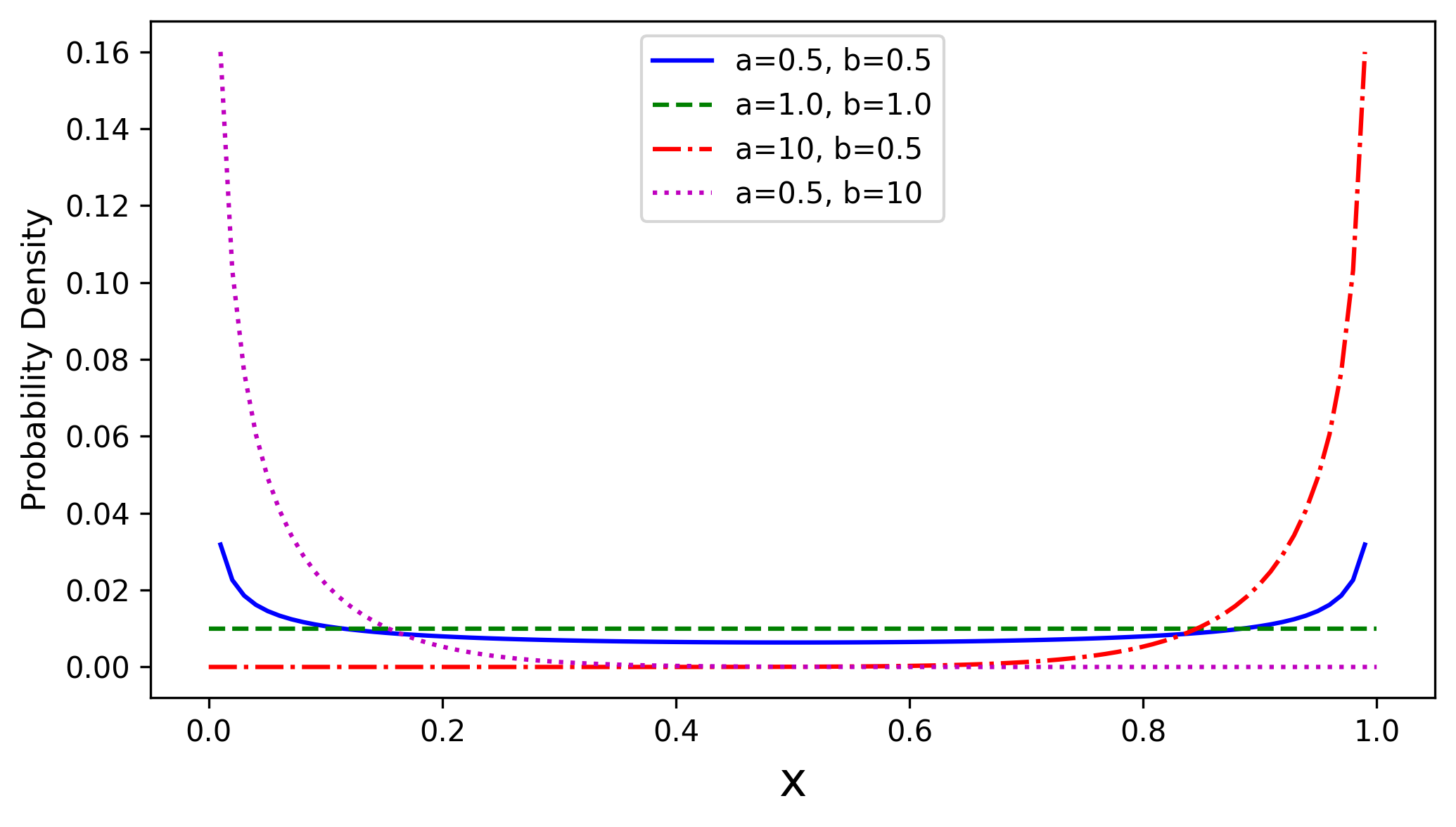}
    \caption{Some priors of Beta distribution}
       \label{fig:beta-visual}
\end{figure}

Increasing the $a$ parameter skews the distribution to the right, while altering the $b$ parameter skews it to the left. Therefore, adding $k$ to $a$ identifies the most optimistic (positively skewed) distribution obtainable with the given parameters. Consequently, this facilitates the calculation of the most optimistic lower bound that can be attained: $\delta_{max}$. The proof for this statement is in Sec.~\ref{appendix:proof}.

Fig.~\ref{fig:beta-visual} visualizes the shape of noninformative Jeffreys prior used in the paper: (0.5, 0.5). This prior is a U-shaped distribution, with slightly denser tails. Another plausible option was to utilize a prior of (1.0, 1.0), resulting in a uniform distribution.

Below, we provide a plot illustrating the relationship between all these parameters:

\begin{figure}[h]
    \centering
    \includegraphics[width=0.7\textwidth]{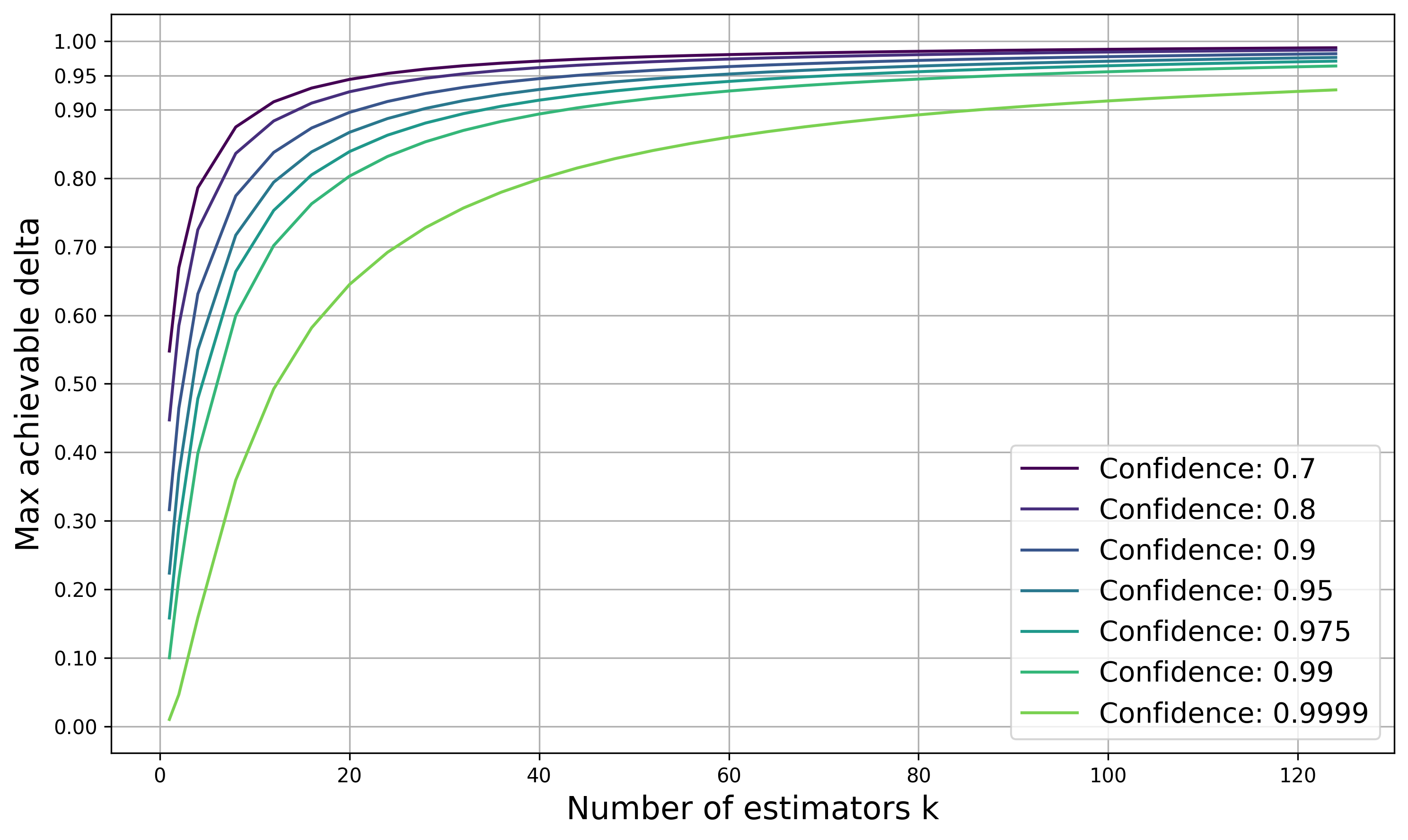}
    \caption{Max achievable $\delta$ values as a function of $k$ and $\alpha$.}

    \label{fig:max-delta-plots}
\end{figure}

Furthermore, we include an auxiliary table (Tab.~\ref{appendix:tab-params} containing precomputed $\delta_{max}$ values (assuming priors equal to $0.5$) to facilitate parameter selection in \betaRCE for the reader:

\begin{table}[h]
\caption{A table of ready-to-use parameter settings. The columns stand for $\alpha$ values, rows for $k$, and cells for $\delta$.}
    \footnotesize
\centering
\begin{tabular}{lrrrrrrr}
    \toprule
    $k$ $\alpha$  &  0.7 &  0.8 &  0.9 &  0.95 &  0.975 &  0.99 & 0.999 \\
    \midrule
    1   &  0.387 &  0.316 &  0.224 &  0.158 &  0.112 &  0.071 &  0.022 \\
    2   &  0.531 &  0.464 &  0.368 &  0.292 &  0.232 &  0.171 &  0.079 \\
    4   &  0.684 &  0.631 &  0.549 &  0.478 &  0.416 &  0.347 &  0.219 \\
    12  &  0.864 &  0.838 &  0.794 &  0.753 &  0.714 &  0.665 &  0.557 \\
    20  &  0.914 &  0.896 &  0.867 &  0.839 &  0.812 &  0.777 &  0.696 \\
    28  &  0.937 &  0.924 &  0.902 &  0.881 &  0.860 &  0.833 &  0.769 \\
    36  &  0.950 &  0.940 &  0.922 &  0.905 &  0.888 &  0.867 &  0.814 \\
    44  &  0.959 &  0.950 &  0.936 &  0.921 &  0.907 &  0.889 &  0.845 \\
    52  &  0.965 &  0.957 &  0.945 &  0.933 &  0.921 &  0.905 &  0.866 \\
    60  &  0.969 &  0.963 &  0.952 &  0.941 &  0.931 &  0.917 &  0.883 \\
    68  &  0.973 &  0.967 &  0.958 &  0.948 &  0.938 &  0.926 &  0.896 \\
    76  &  0.976 &  0.971 &  0.962 &  0.953 &  0.945 &  0.934 &  0.906 \\
    84  &  0.978 &  0.973 &  0.965 &  0.958 &  0.950 &  0.940 &  0.914 \\
    92  &  0.980 &  0.976 &  0.968 &  0.961 &  0.954 &  0.945 &  0.922 \\
    100 &  0.981 &  0.977 &  0.971 &  0.964 &  0.958 &  0.949 &  0.928 \\
    108 &  0.983 &  0.979 &  0.973 &  0.967 &  0.961 &  0.953 &  0.933 \\
    116 &  0.984 &  0.981 &  0.975 &  0.969 &  0.963 &  0.956 &  0.937 \\
    124 &  0.985 &  0.982 &  0.976 &  0.971 &  0.966 &  0.958 &  0.941 \\
    \bottomrule
    \end{tabular}
\label{appendix:tab-params}
\end{table}

\section{Experimental setup }
In this section, we provide more details on the implementation of experiments.

\subsection{General}
For all experiments, we utilized a 3-fold cross-validation approach, with 2 folds allocated for training and a single fold for evaluation. 
During evaluation on each fold, we randomly sampled 30 instances for the generation of robust counterfactuals and then assessed the \textit{Empirical Robustness} on 30 new models (from the space of admissible model changes).

For each model, we randomly split the training data into 80-20 \textit{train}-\textit{validation} sets to facilitate model training and parameter tuning.

\subsection{Datasets}
\label{app:sec:datasets}

Below, we provide information about the datasets used in our study. 
\begin{center}
    \footnotesize
    \begin{tabular}{lccc}
    \textbf{Dataset} & \textbf{Rows} & \textbf{Columns} & \textbf{Imbalance Ratio} \\ \hline
    HELOC & 2502 & 24 & 1.66 \\
    Wine & 6497 & 12 & 1.73 \\
    Diabetes & 768 & 9 & 1.87 \\
    Breast Cancer & 569 & 31 & 1.68 \\
    Car eval & 1728 & 6 & 2.34 \\
    Rice & 3810 & 7 & 1.34\\
    
    \end{tabular}
\end{center}
Preprocessing of these datasets involved dropping rows containing missing values and performing min-max normalization.

\subsection{Hyperparameters of the Baselines}
\label{app:sec:hparams-baselines}

Below, we present the hyperparameters that were searched for every end-to-end CFE generation method, both the standard and robust ones:

\begin{itemize}
    \footnotesize
    \item \textbf{\dice}
    \begin{itemize}
        \item Diversity Weight: \{0.05, 0.1, 0.2\}
        \item Proximity Weight: \{0.05, 0.1, 0.2\}
        \item Sparsity Weight: \{0.05, 0.1, 0.2\}
    \end{itemize}

    \item \textbf{\face}
    \begin{itemize}
        \item Fraction: \{0.1, 0.3, 0.5\}
        \item Mode: \{knn, epsilon\}
    \end{itemize}

    \item \textbf{\rbr}
    \begin{itemize}
        \item Max Distance: 1.0
        \item Num Samples: 100
        \item Delta Plus: \{0.0, 0.1, 0.2\}
        \item Epsilon OP: 0.0
        \item Epsilon PE: 0.0
        \item Sigma: \{0.5, 1.0, 1.5\}
        \item Perturb Radius (synthesis): \{0.1, 0.2, 0.3\}
    \end{itemize}

    \item \textbf{\roar}
    \begin{itemize}
        \item Delta Max: \{0.01, 0.05, 0.1\}
        \item Learning Rate (LR): \{0.01, 0.05, 0.1\}
        \item Norm: \{1, 2\}
    \end{itemize}

    \item \textbf{\robx}
    \begin{itemize}
        \item N: 1000
        \item $\tau$: \{0.4, 0.5, 0.6, 0.7, 0.8\}
        \item Variances: \{0.1, 0.01\}
    \end{itemize}
\end{itemize}

For all visualizations, we chose the hyperparameter configuration that provided the highest empirical robustness to ensure a fair comparison. The only exception is the post-hoc method \robx, as it is also crucial to evaluate the distance to the base counterfactual in such methods. Therefore, to highlight different aspects of \robx, we included two distinct settings in all comparisons: one optimized for empirical robustness and another that strikes a balance with a good distance to the base CFE.

\subsection{Models}
\label{app:sec:models}
In our experiments, we utilize three models as the core black-boxes: a neural network (NN), LightGBM, and logistic regression (LR). These models are implemented using torch, lightgbm, and sklearn, respectively. The validation sets were employed for early stopping in the NN and as the evaluation set for LightGBM. Detailed specifications are provided below:

\subsubsection{Neural Network}  

\begin{center}
    \footnotesize
    \begin{tabular}{lcc}
    \textbf{Parameter} & \textbf{Fixed hparams} & \textbf{Hparams to Vary} \\ \hline
    Layers & 3 & 3-5 \\
    Neurons per layer & 128 & 64-256 \\
    Activations & ReLU & \\
    Terminal activation & Sigmoid & \\
    Optimizer & Adam & \\
    Learning rate & 1e-3 & \\
    Loss & BCE & \\
    Early stopping & 5 & \\
    Dropout & 0.4 & \\
    Batch size & 128 & \\
    Seed & 42 & \\
    \end{tabular}
\end{center}

\subsubsection{LightGBM}

\begin{center}
    \footnotesize
    \begin{tabular}{lcc}
    \textbf{Parameter} & \textbf{Fixed hparams} & \textbf{Hparams to Vary} \\ \hline
    No. of leaves & 15 & 10-20 \\
    No. of estimators & 30 & 15-40 \\
    Min. child samples & 20 & 10-20 \\
    Subsample & 0.8 & 0.5-1.0 (freq: 0.1) \\
    Objective & binary & \\
    Seed & 42 &
    \end{tabular}
\end{center}

\subsubsection{Logistic Regression}
\begin{center}
    \footnotesize
    \begin{tabular}{lcc}
    \textbf{Parameter} & \textbf{Fixed hparams} & \textbf{Hparams to Vary} \\ \hline
    solver & lbfgs & lbfgs, newton-cg, sag \\
    penalty & l2 & l2, none \\
    max\_iter & 100 & 50 - 200 \\ 
    C & 1 & 0.1 - 1.0 \\ 
    seed & 42 & \\ 
    \end{tabular}
\end{center}


\section{\betaRCE intrinsic analysis}
In this section, we expand on the analysis presented in the main body of the paper regarding the impact of \betaRCE parameters on various aspects of the method's performance.

\subsection{Credible intervals for robustness}
\label{app:sec:cred-interv-the}

\begin{figure*}[htbp]
    \caption{With \growingSpheres as a base counterfactual explanation.}
    \includegraphics[width=\textwidth]{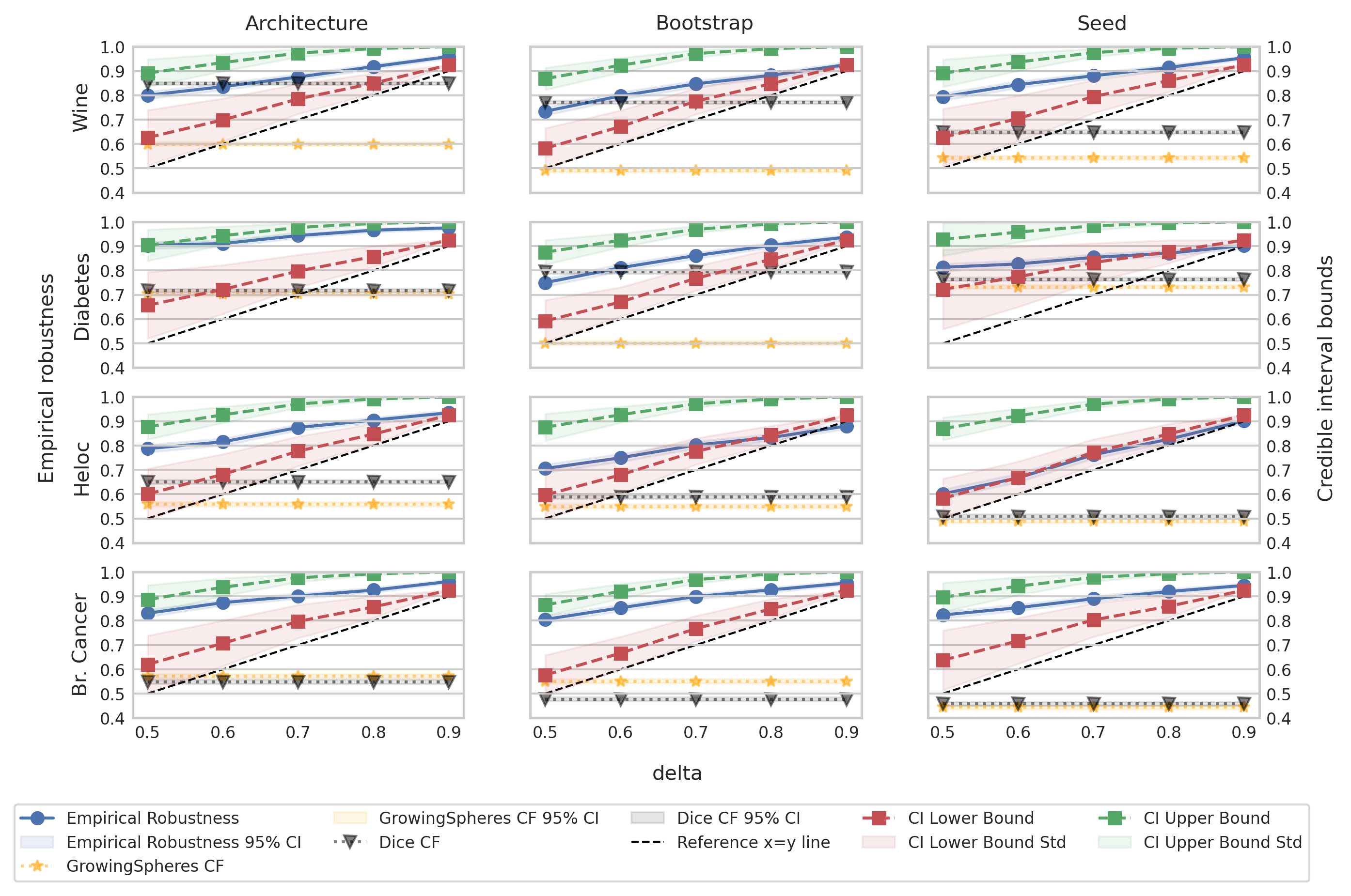}
    \label{app:fig2-gs}
\end{figure*}

\begin{figure*}[htbp]
    \caption{With \dice as a base counterfactual explanation.}
    \includegraphics[width=\textwidth]{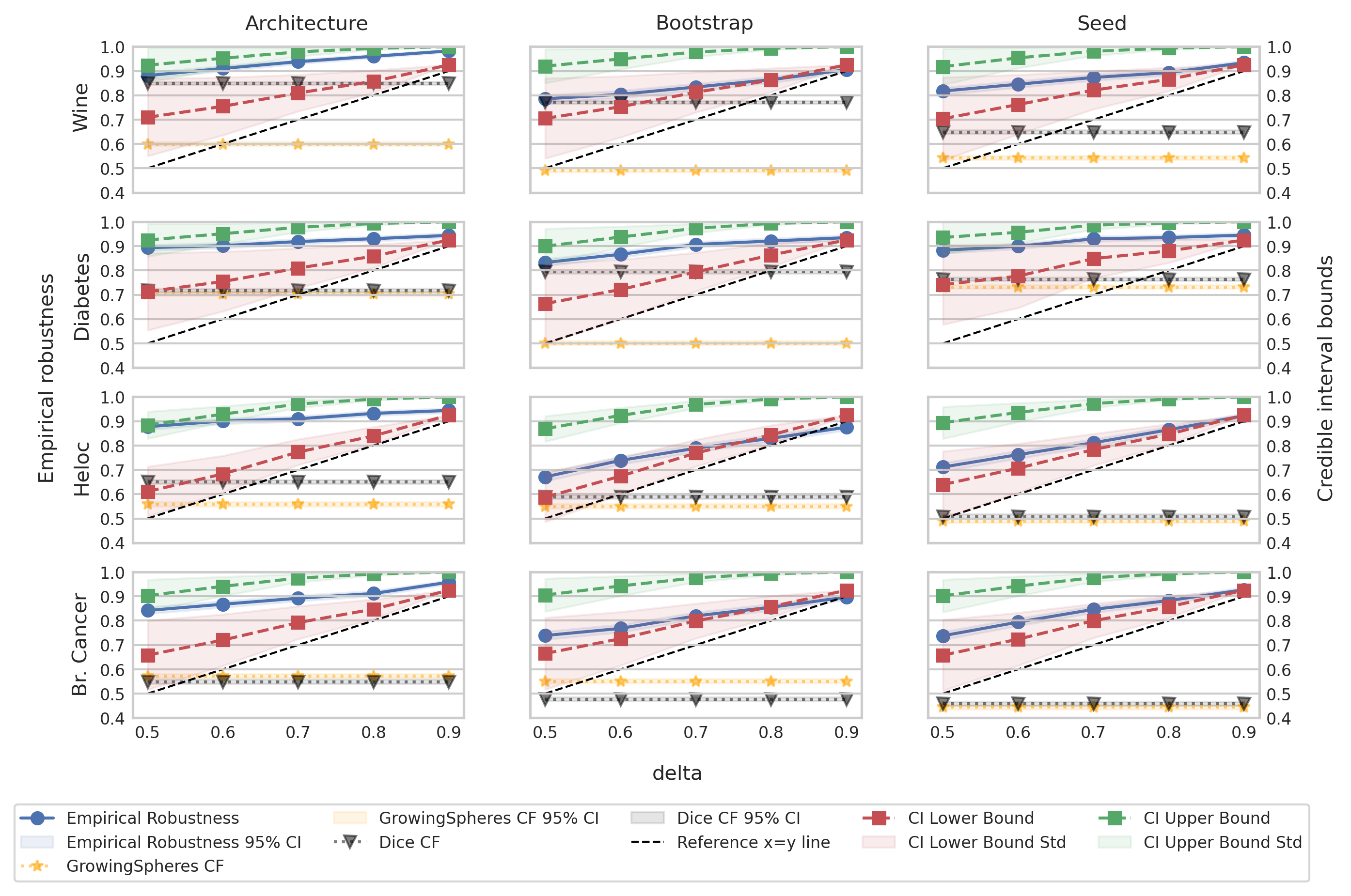}
    \label{app:fig2-dice}
\end{figure*}

In this section, we present the full version of Fig.~\ref{fig:empirical-validation} from the main paper with \growingSpheres as a base explainer (Fig.~\ref{app:fig2-gs}), and we also include an additional plot with \dice (Fig.~\ref{app:fig2-dice}) serving as a base explainer. 

\clearpage
\clearpage
\subsection{Exploring the impact of the confidence parameter}
\label{app:sec:hparams-confidence}

The parameter $\alpha$ reflects the overall confidence in the estimates provided by our method. Here, we briefly look into how different $\alpha$ values influence the model's performance.

Our first analysis juxtaposes $\alpha$ with \textit{Empirical Robustness} across three $\delta$ (Fig.~\ref{fig:app-emprob-alpha-delta}).

As observed, the average \textit{Empirical Robustness} shows a slight increase with higher confidence values. This aligns with the notion that greater prediction confidence leads to a more secure robustness estimate, consequently yielding a higher average \textit{Empirical Robustness}.

The subsequent visualization (Fig~\ref{fig:app-emprob-ci-bounds-confidence}) illustrates this enhanced security with higher $\alpha$ values, as indicated by the greater distance between the blue line and the red line, representing the lower bound of the credible interval:

\begin{figure}[htbp]
    \centering
    \begin{subfigure}{0.45\textwidth}
        \includegraphics[width=\textwidth]{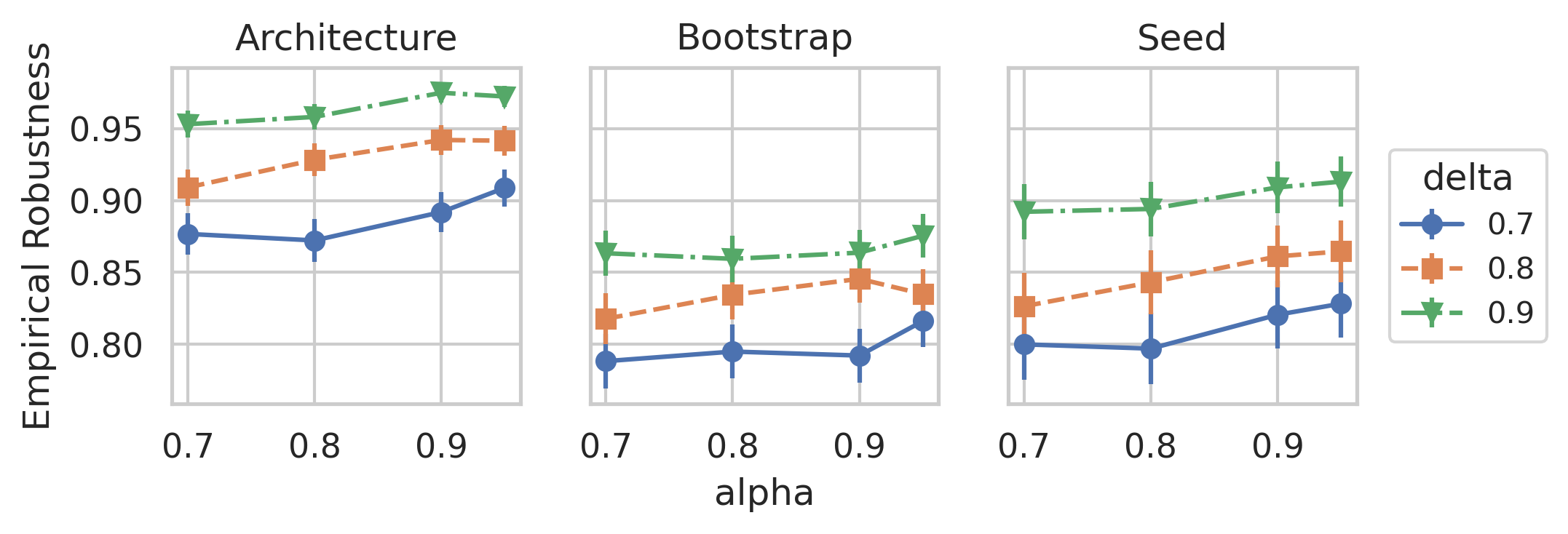}
        \vspace{0.02cm}
        \caption{Empirical robustness as a function of $\alpha$ and $\delta$}
        \vspace{0.35cm}
        \label{fig:app-emprob-alpha-delta}
    \end{subfigure}
    \hfill
    \begin{subfigure}{0.5\textwidth}
        \includegraphics[width=\textwidth]{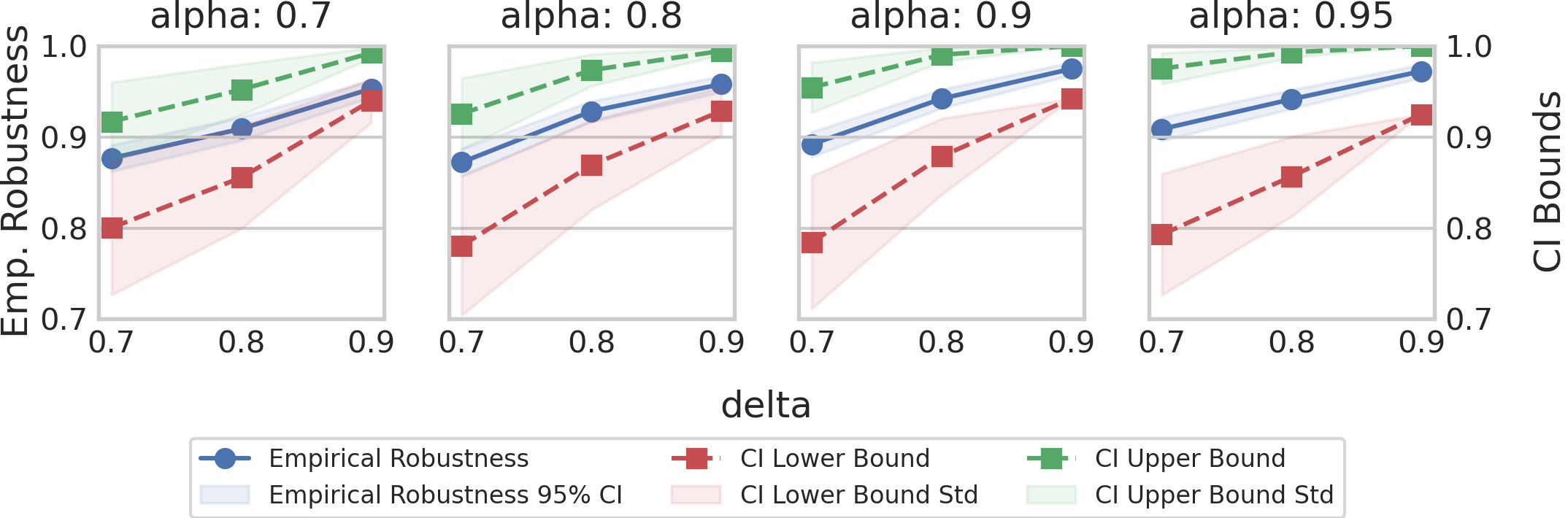}
        \caption{The relationship of empirical robustness and $\alpha$ confidence and its placement credible intervals}
        \label{fig:app-emprob-ci-bounds-confidence}
    \end{subfigure}
\end{figure}

\subsection{Exploring the impact of the number of estimators}
\label{app:sec:hparams-k}

\begin{figure}[h]
    \centering
    \begin{subfigure}{0.48\textwidth}
        \includegraphics[width=\textwidth]{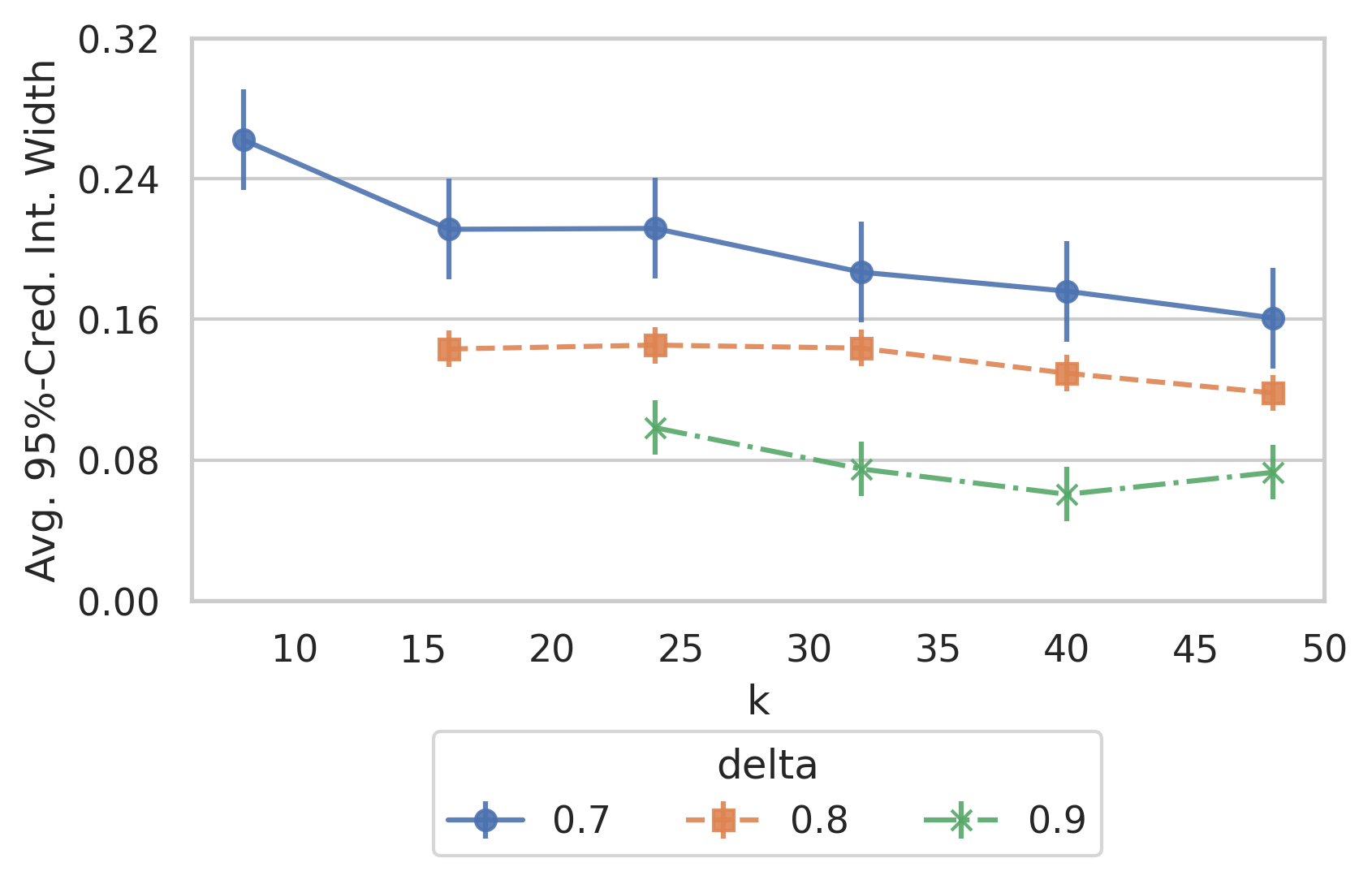}
        \label{fig:app-kmlps-width}
    \end{subfigure}
    \hfill
    \begin{subfigure}{0.48\textwidth}
        \includegraphics[width=\textwidth]{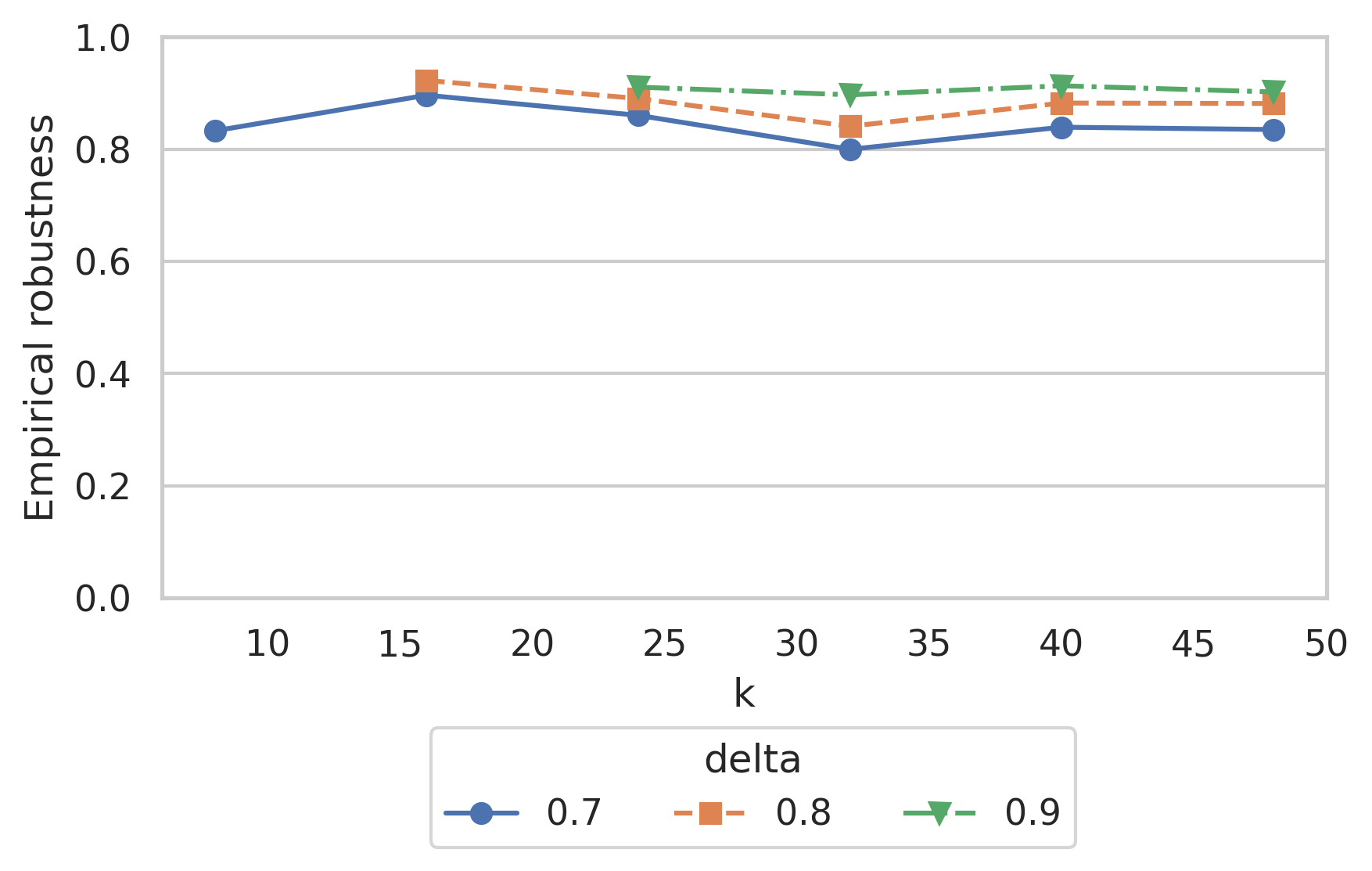}
        \label{fig:app-kmlps-emprob}
    \end{subfigure}
    \caption{The impact of parameter $k$ on the performance of \betaRCE. The subfigure on the left side shows how $k$ affects the credible interval width, while the one on the right side illustrates the relationship between $k$ and \textit{Empirical Robustness}.}
    \label{fig:app-kmlps-combined}
\end{figure}

In this section, we examine how varying the number of estimators, denoted as $k$, affects the performance of \betaRCE. As depicted in Fig.~\ref{fig:app-kmlps-combined}, increasing $k$ results in narrower credible intervals, indicating a higher level of confidence in the robustness range. Additionally, the relationship between $k$ and the \textit{Empirical Robustness} is illustrated. 

Increasing $k$ allows for more combinations of parameters $a$ and $b$ to form the Beta distribution, making the distribution more flexible and better suited to the empirical data. Consequently, this leads to narrower credible intervals.

However, as observed in the right pane of Fig.~\ref{fig:app-kmlps-combined}, the average \textit{Empirical Robustness} does not appear to strongly depend on $k$. From these empirical experiments, the conclusion that increasing $k$ enhances \textit{Empirical Robustness} cannot be drawn. Therefore, our recommendation is to use the lowest possible $k$ that achieves the desired $\delta$, as determined by Eq.~\ref{eq:parameters}.

\subsection{Investigating the generalization capabilities across different experiment types}
\label{app:sec:gen}
In this section, we conduct an experimental analysis to investigate how \betaRCE performs when its admissible model space contains different model change types than those encountered during deployment. 
Specifically, we sample from an admissible model space that does not overlap with the one used for evaluation. 

The results of these experiments are presented in two tables (Tables \ref{app:tab:generalization-gs} and \ref{app:tab:generalization-dice}). 
Each table shows the results for a different base CFE method, averaged across four datasets, with $\delta = 0.9$ and $\alpha = 0.95$.

The diagonal in the table is the normal, in-distribution setting, while all the other cells contain generalizations. 
As observed, even though the changes are out-of-distribution, \betaRCE still robustifies counterfactuals to a satisfiable extent. 
It is worth to note, that the probabilistic bounds do not hold for out-of-distribution changes, but from the practical perspective it is useful to generalize well for such changes, which \betaRCE seems to do well. 

The diagonal in the table represents the normal, in-distribution setting, while all other cells contain generalizations. 
As observed, even though the changes are out-of-distribution, \betaRCE still robustifies counterfactuals to a satisfactory extent. 
It is worth noting that the probabilistic bounds do not hold for out-of-distribution changes, but from a practical perspective, it is useful to generalize well for such changes, which \betaRCE seems to be able to accomplish.

\begin{table}
\centering
    \footnotesize
\caption{Empirical Robustness of \betaRCE with \growingSpheres as the base CFE generation method. The results are averaged over all datasets.}
\begin{tabular}{l|ccc}
 & \multicolumn{3}{c}{\textbf{Generalization}} \\
 \textbf{Original} & $\text{Architecture}$ & $\text{Bootstrap}$ & $\text{Seed}$ \\
\midrule
$\text{Architecture}$ & $0.913 \pm 0.007$ & $0.865 \pm 0.009$ & $0.923 \pm 0.007$ \\
$\text{Bootstrap}$ & $0.939 \pm 0.006$ & $0.877 \pm 0.008$ & $0.909 \pm 0.007$ \\
$\text{Seed}$ & $0.927 \pm 0.007$ & $0.866 \pm 0.009$ & $0.890 \pm 0.008$ \\
\end{tabular}
\label{app:tab:generalization-gs}
\end{table}

\begin{table}
\centering
    \footnotesize
\caption{Empirical Robustness of \betaRCE with \dice as the base CFE generation method. The results are averaged over all datasets.}
\label{app:tab:generalization-dice}
\begin{tabular}{l|ccc}
 & \multicolumn{3}{c}{\textbf{Generalization}} \\
 \textbf{Original} & $\text{Architecture}$ & $\text{Bootstrap}$ & $\text{Seed}$ \\
\midrule
$\text{Architecture}$ & $0.937 \pm 0.005$ & $0.875 \pm 0.007$ & $0.930 \pm 0.005$ \\
$\text{Bootstrap}$ & $0.927 \pm 0.005$ & $0.847 \pm 0.007$ & $0.913 \pm 0.006$ \\
$\text{Seed}$ & $0.929 \pm 0.005$ & $0.805 \pm 0.008$ & $0.918 \pm 0.005$ \\
\end{tabular}
\end{table}

\section{Computational Complexity}

The computational complexity of \textsc{BetaRCE} is determined primarily by two factors: (1) the complexity of the chosen optimization algorithm and (2) the number of estimators ($k$) used for bootstrap robustness verification. The latter directly affects the complexity of evaluating the objective function's constraints, as each evaluation requires querying $k$ estimators. Consequently, $k$ inference calls are introduced as a constant multiplier to the overall complexity.

The optimization algorithm employed to solve Eq.~\ref{eq:optimisation-goal} plays the most significant role in determining \textsc{BetaRCE}'s complexity. In this paper, we utilize \growingSpheres for optimization. While this algorithm does not offer theoretical guarantees regarding the complexity of finding an optimal or $\epsilon$-optimal solution, its complexity is heavily influenced by $\eta$ and $n$ hyperparameters. 

First, $\eta$ controls the granularity of the iterative expansion of the sphere's perimeter. Second, $n$ specifies the number of samples evaluated on the perimeter during each iteration. When combined with \textsc{BetaRCE}, each step of \growingSpheres requires $n \cdot k + 1$ model evaluations. Here, $k$ estimators perform inference on each of the $n$ samples to compute the robustness term (Eq.~\ref{eq:robust_term}), while the additional $+1$ accounts for the evaluation of the validity term (Eq.~\ref{eq:valid-cf}).

\begin{figure}[h]
    \centering
    \includegraphics[width=0.7\columnwidth]{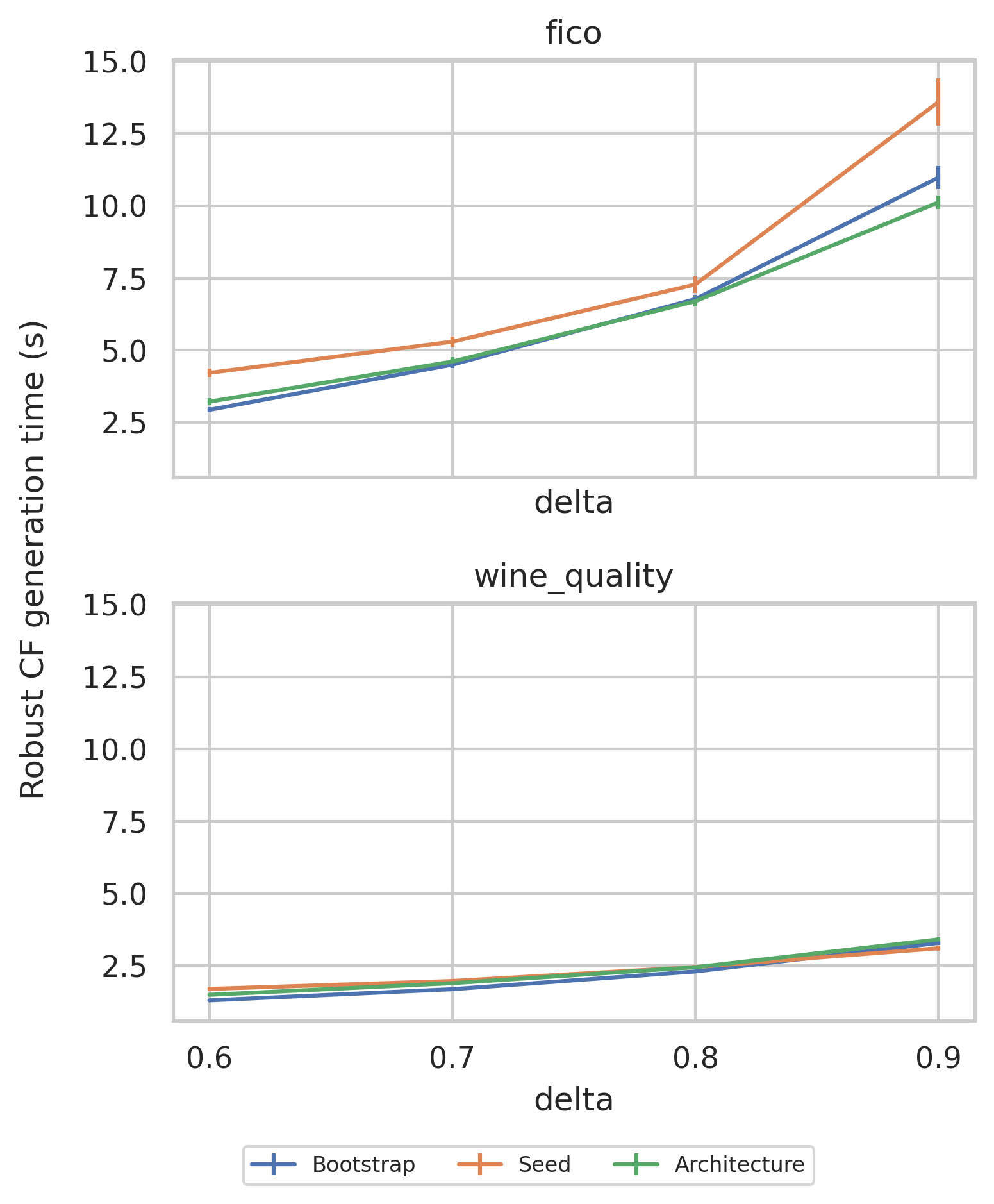}
    \caption{Robust counterfactual generation time as a function of $\delta$.}
    \label{fig:app:dist-base-correlation}
\end{figure}

\vspace{-0.5cm}
\section{Step-by-step visualization of \betaRCE algorithm}
\label{app:betarce-visualization}

\begin{figure*}[htbp]
    \centering
    \begin{subfigure}[b]{0.39\textwidth}
        \includegraphics[width=\textwidth]{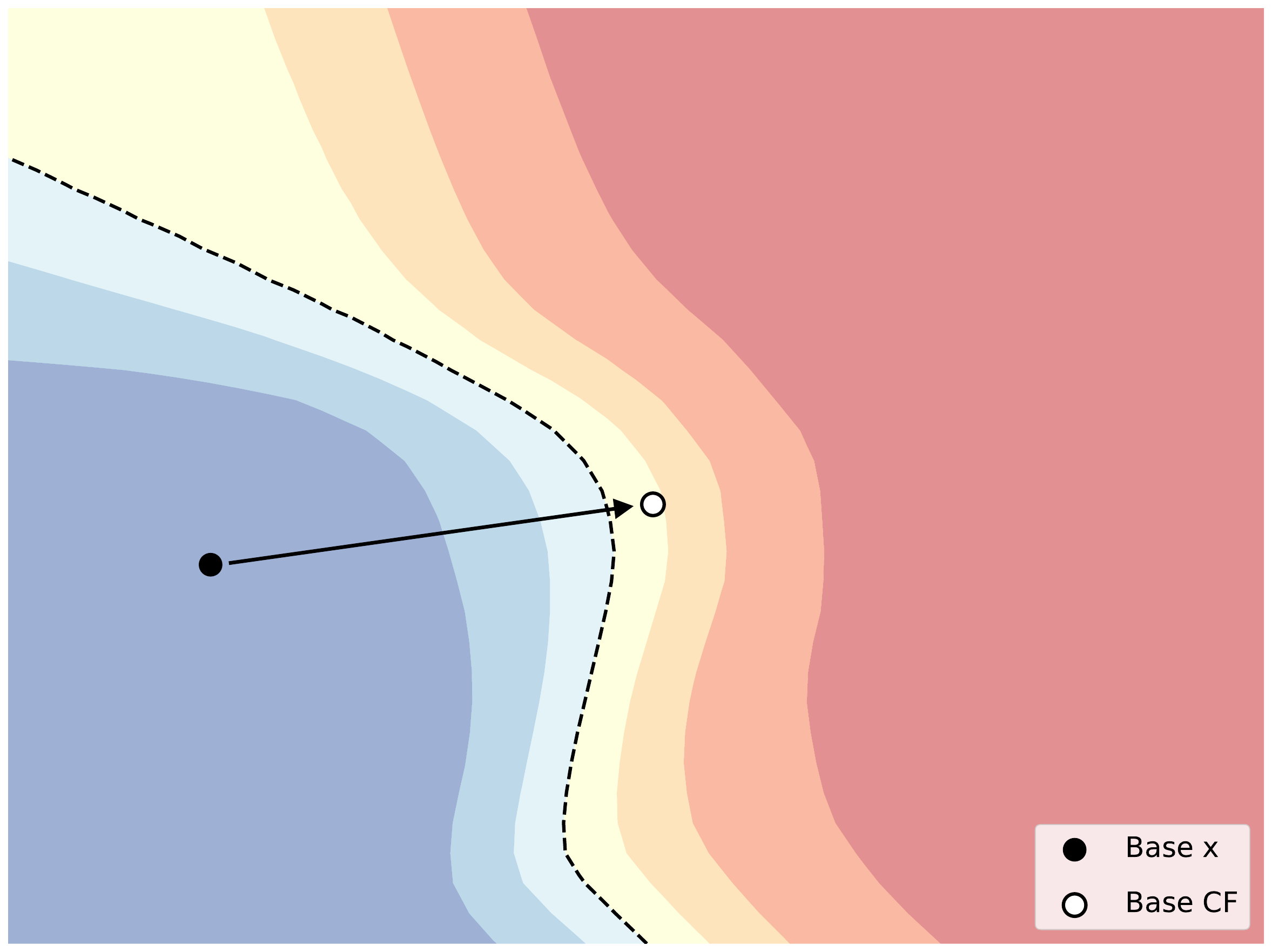}
        \caption{}
        \label{app:10:fig:panel1}
    \end{subfigure}
    \begin{subfigure}[b]{0.39\textwidth}
        \includegraphics[width=\textwidth]{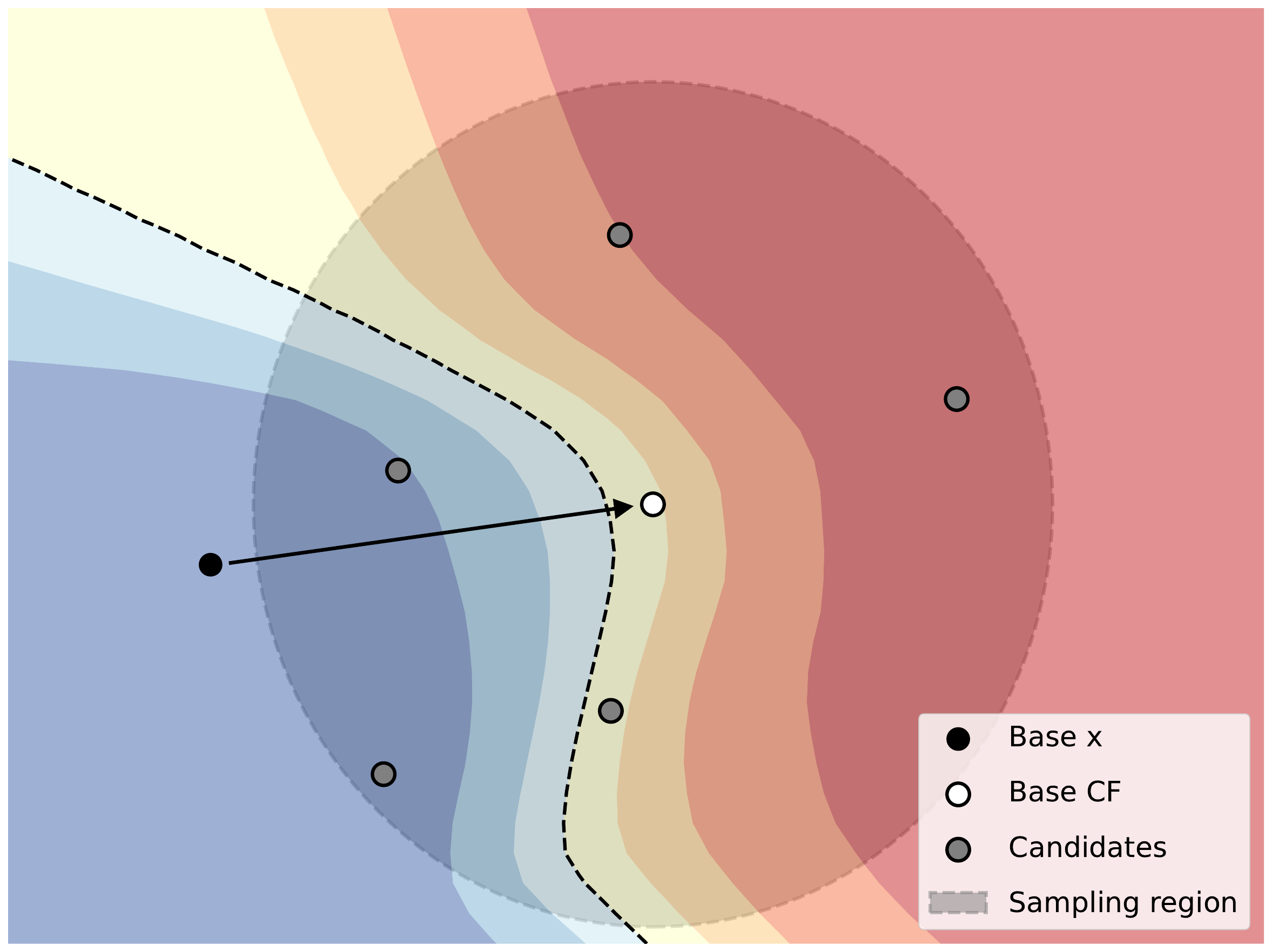}
        \caption{}
        \label{app:10:fig:panel2}
    \end{subfigure}
    
    \begin{subfigure}[b]{0.39\textwidth}
        \includegraphics[width=\textwidth]{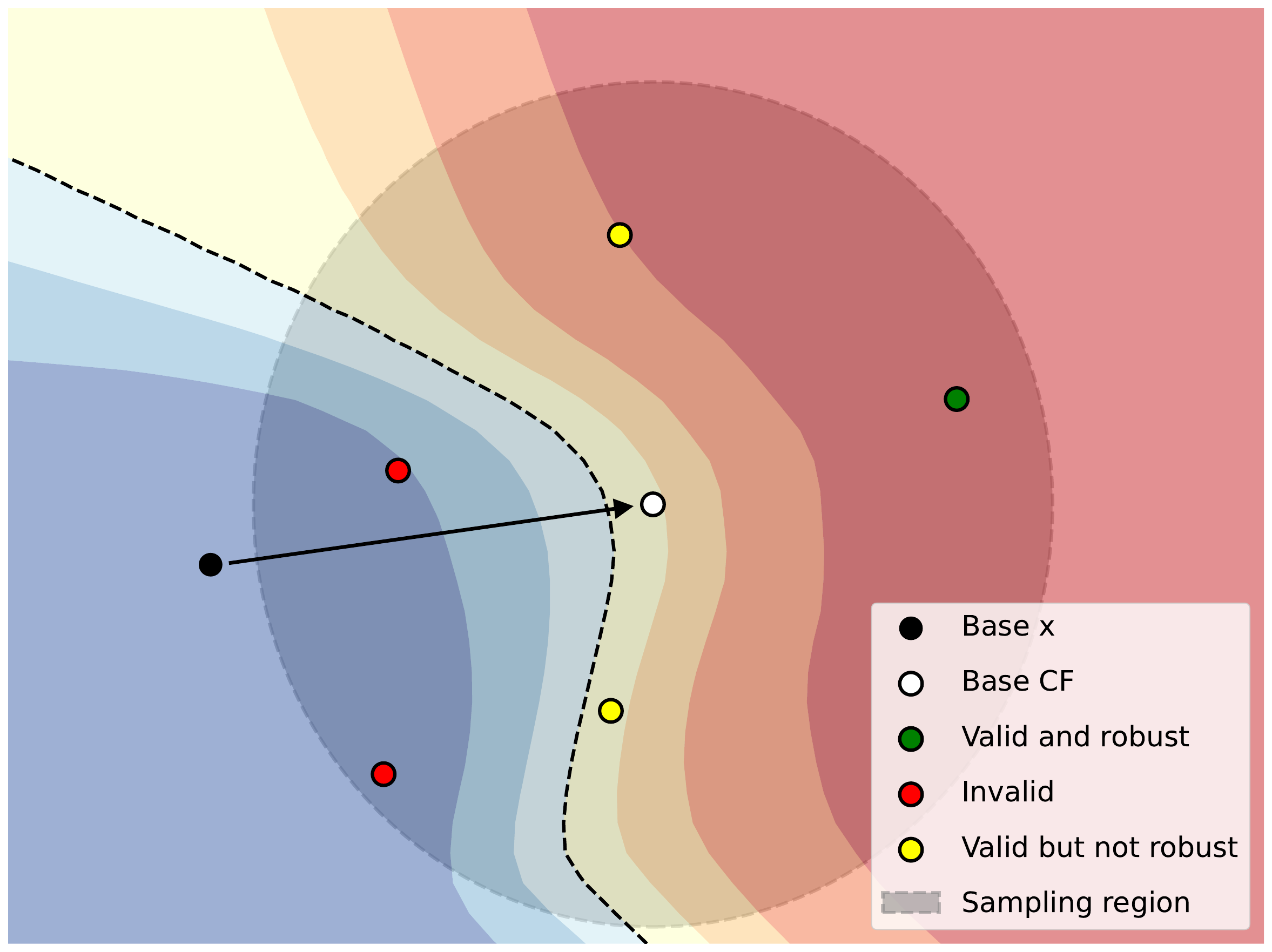}
        \caption{}
        \label{app:10:fig:panel3}
    \end{subfigure}
    \begin{subfigure}[b]{0.39\textwidth}
        \includegraphics[width=\textwidth]{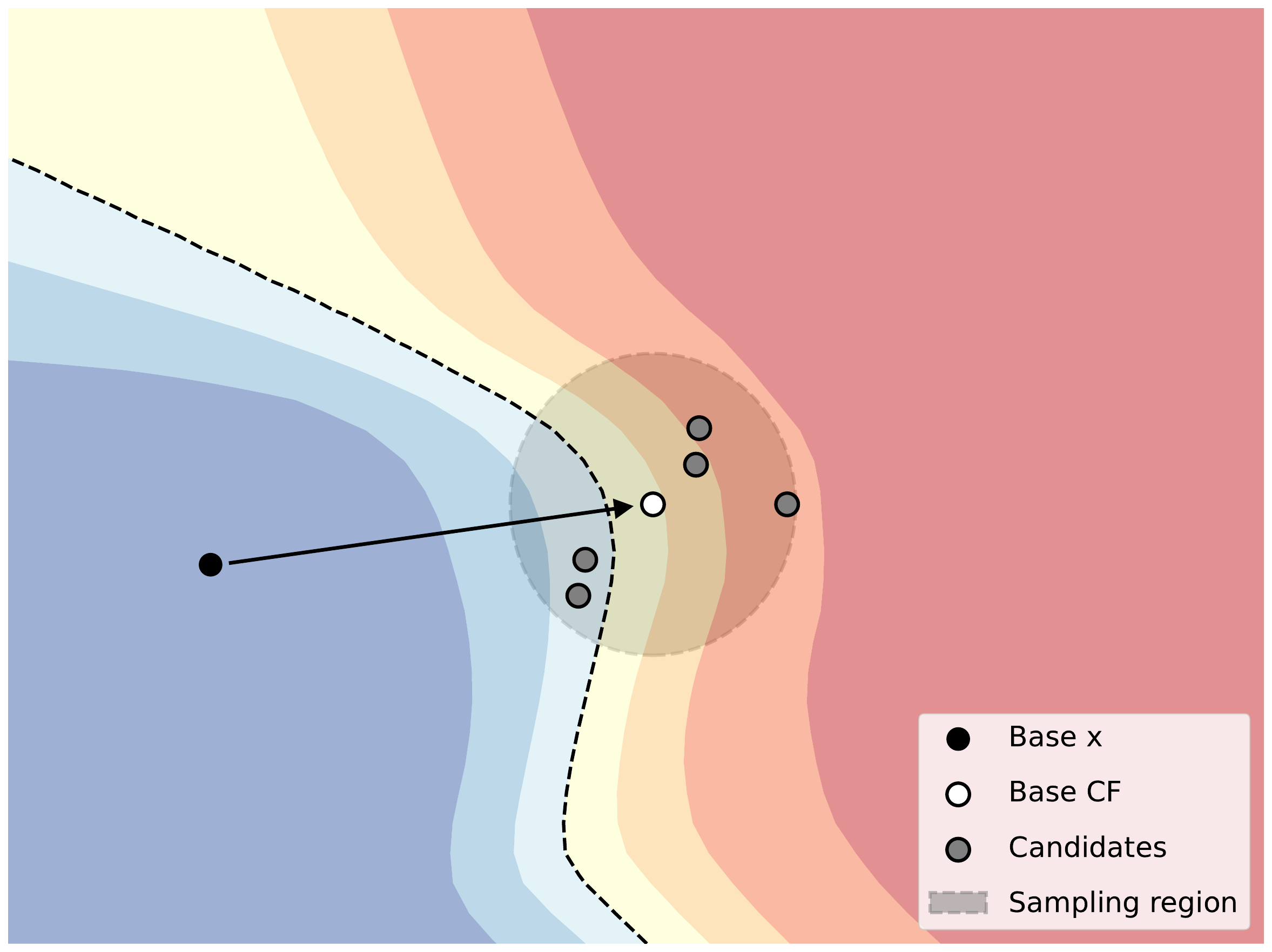}
        \caption{}
        \label{app:10:fig:panel4}
    \end{subfigure}
    
    \begin{subfigure}[b]{0.39\textwidth}
        \includegraphics[width=\textwidth]{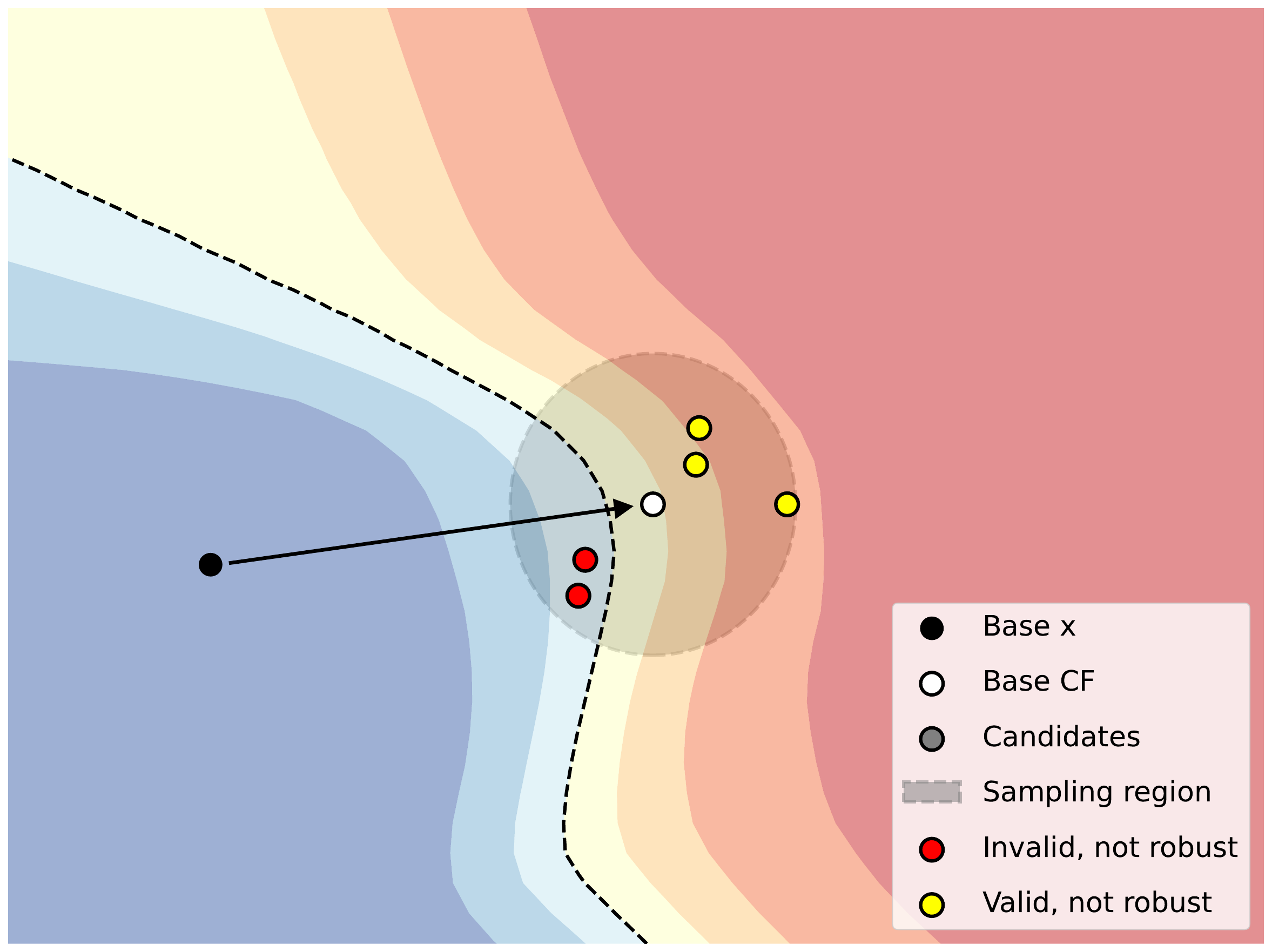}
        \caption{}
        \label{app:10:fig:panel5}
    \end{subfigure}
    \begin{subfigure}[b]{0.39\textwidth}
        \includegraphics[width=\textwidth]{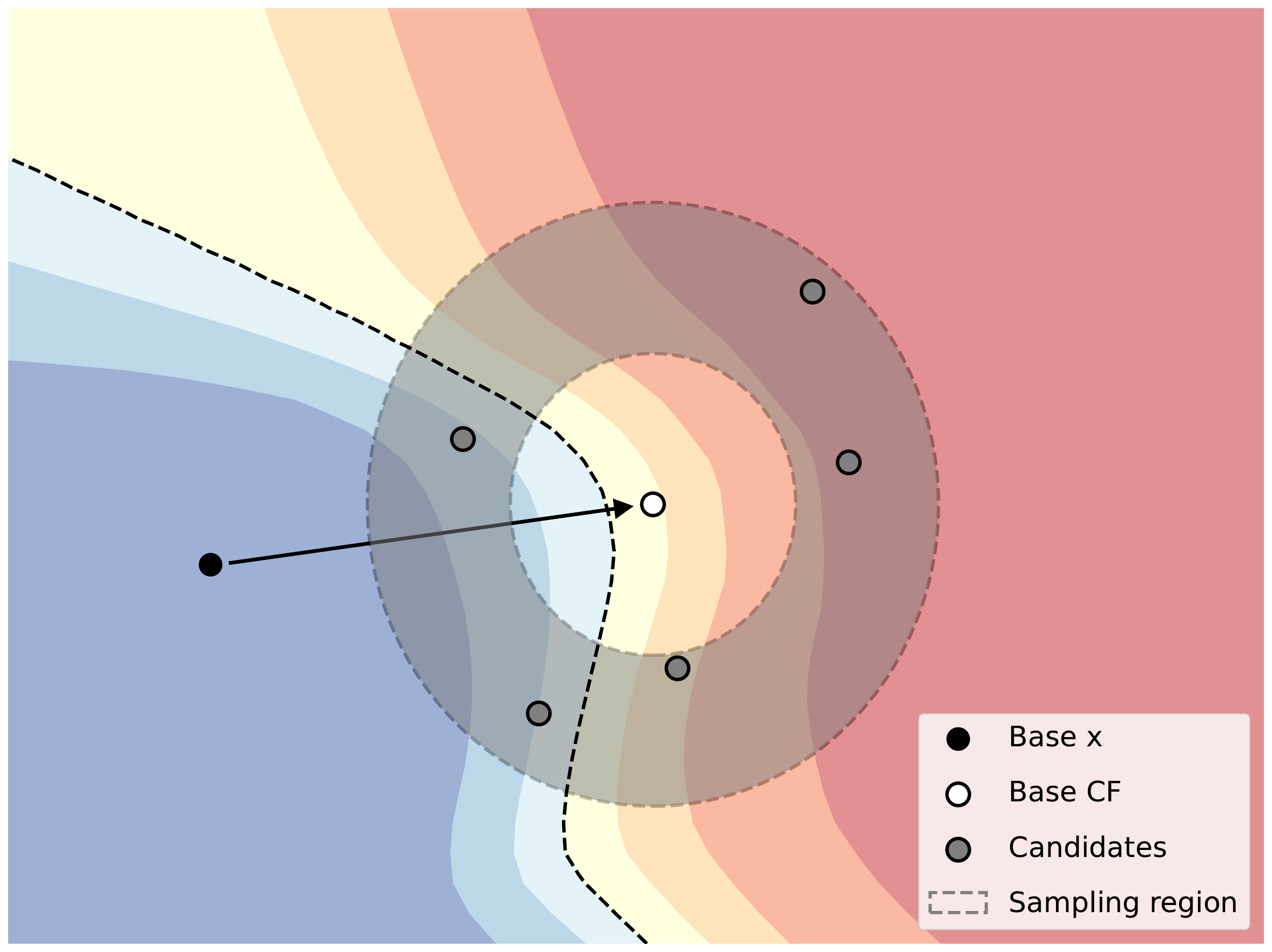}
        \caption{}
        \label{app:10:fig:panel6}
    \end{subfigure}
    
    \begin{subfigure}[b]{0.39\textwidth}
        \includegraphics[width=\textwidth]{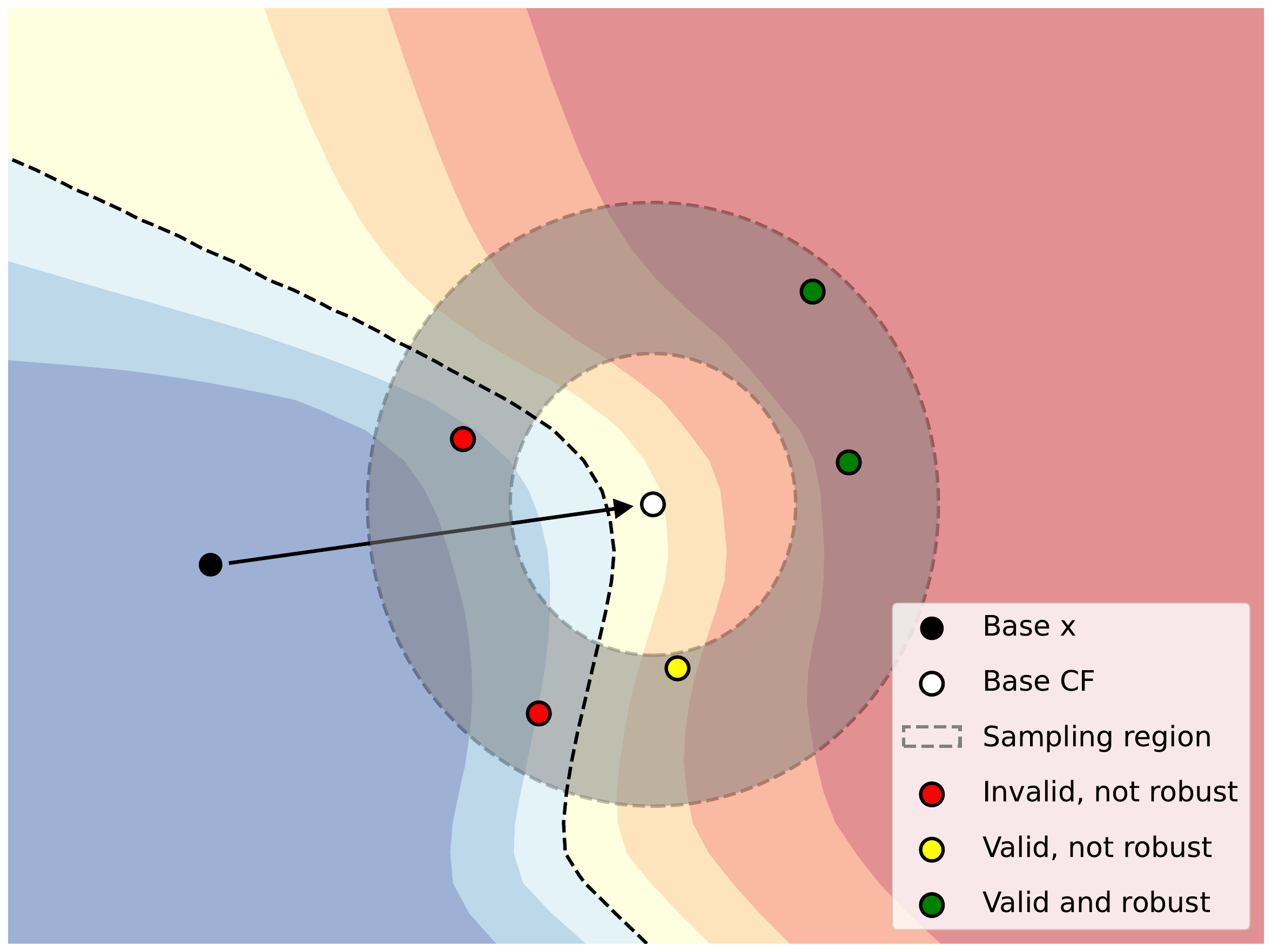}
        \caption{}
        \label{app:10:fig:panel7}
    \end{subfigure}
    \begin{subfigure}[b]{0.39\textwidth}
        \includegraphics[width=\textwidth]{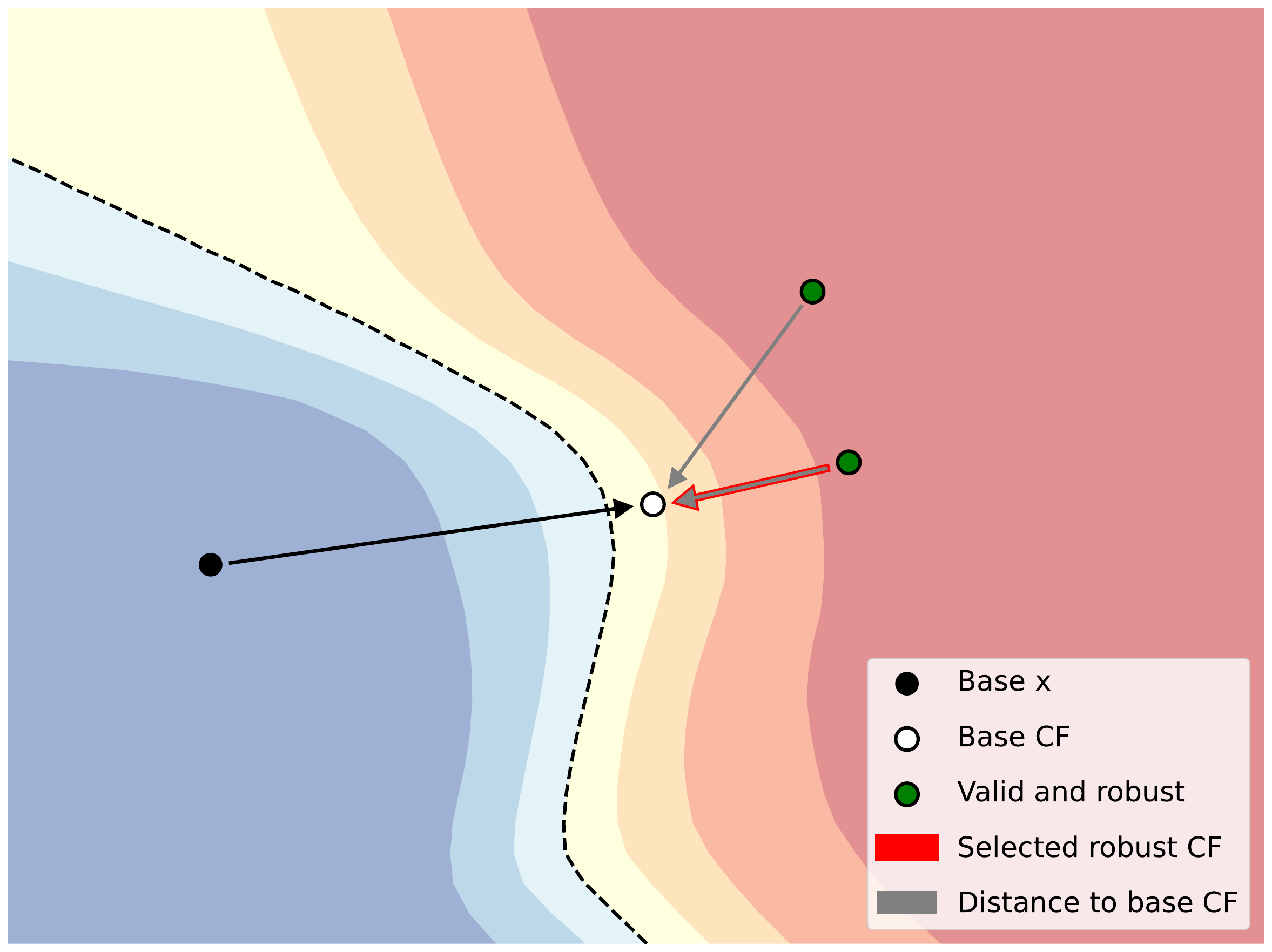}
        \caption{}
        \label{app:10:fig:panel8}
    \end{subfigure}
    
    \caption{A visualization of \betaRCE with \growingSpheres search algorithm.}
    \label{app:fig:8panelplot}
\end{figure*}

In this section, we provide an intuitive visualization of how \betaRCE works when integrated with \growingSpheres. 
Fig.~\ref{app:fig:8panelplot} illustrates eight consecutive steps of the \betaRCE algorithm for a simple example. 

First, a base counterfactual explanation is generated for a given input (see Fig.~\ref{app:10:fig:panel1}). Next, we move to the warmup stage of the \growingSpheres search algorithm, as described in Alg.~\ref{alg:rce-growsph} (lines 2-6). In particular, first, in Fig.~\ref{app:10:fig:panel2} five candidates are sampled from a large sphere (line 2), second, in Fig.~\ref{app:10:fig:panel3} robustness and validity are evaluated (line 3). Since there were both valid and robust examples in the sphere, its radius is halved (line 4). Next, Fig.~\ref{app:10:fig:panel4} and Fig.~\ref{app:10:fig:panel5} show similar steps for a sphere with smaller radius (line 5), but this time, no valid and robust examples were found. Thus, the warmup stage of \growingSpheres is over. 

In the next figures, we show an iteration of the search procedure (lines 7-13), where candidates are sampled from a region between increasing lower and upper radius.  Fig.~\ref{app:10:fig:panel6} illustrates the sampling of candidates (line 8) and Fig.~\ref{app:10:fig:panel7} shows the candidates being evaluated (line 9). Since two robust and valid examples were found, the algorithm is terminated (lines 9-13) and the closest counterfactual (Fig.~\ref{app:10:fig:panel8}) is returned as the robust counterfactual explanation (line 14).  

\clearpage

\begin{figure}[htbp]
\begin{center}
    
    \begin{subfigure}[b]{0.33\textwidth}
        \includegraphics[width=\textwidth]{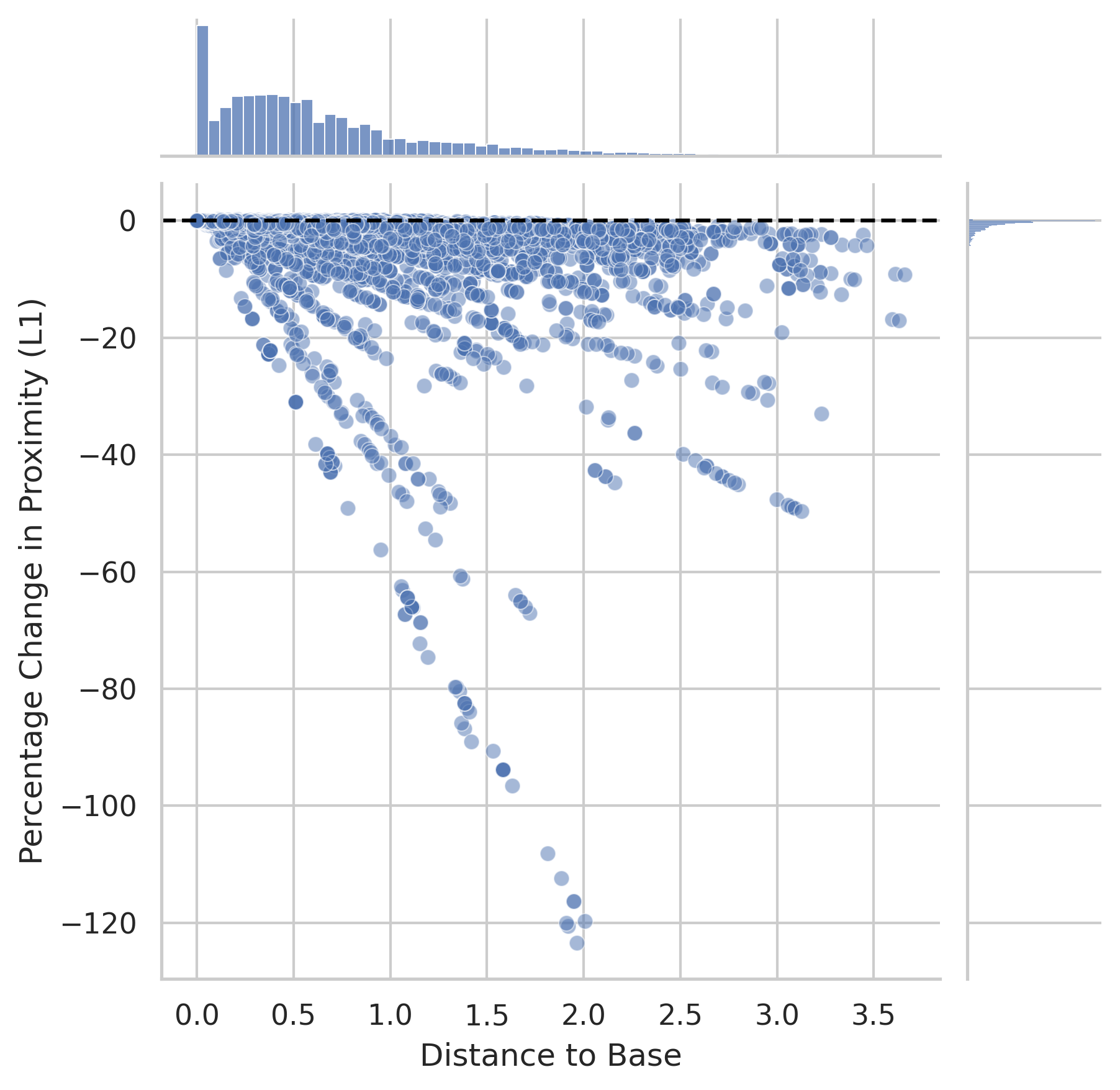}
    \end{subfigure}
    \begin{subfigure}[b]{0.33\textwidth}
        \includegraphics[width=\textwidth]{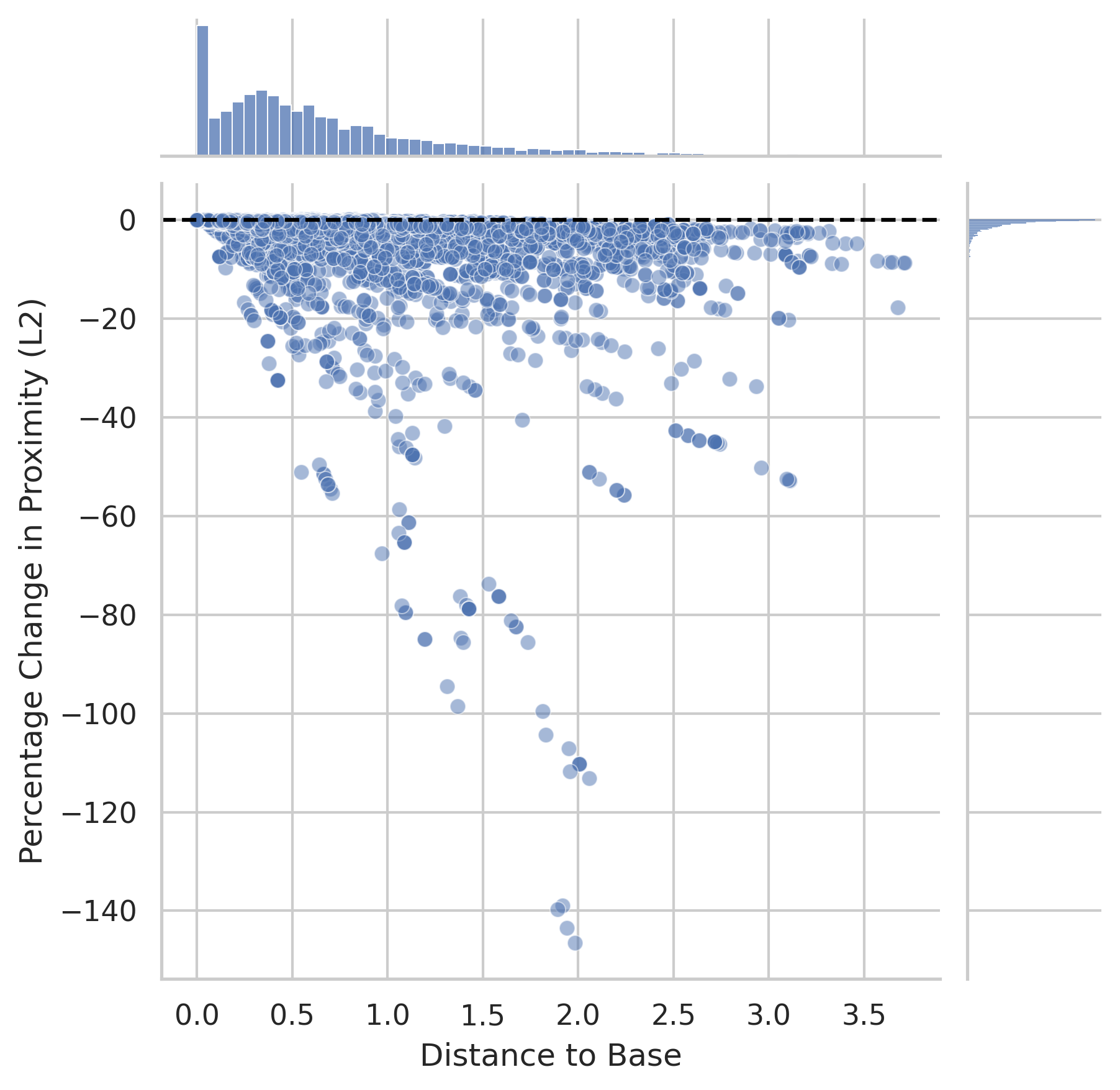}
    \end{subfigure}
    \begin{subfigure}[b]{0.33\textwidth}
        \includegraphics[width=\textwidth]{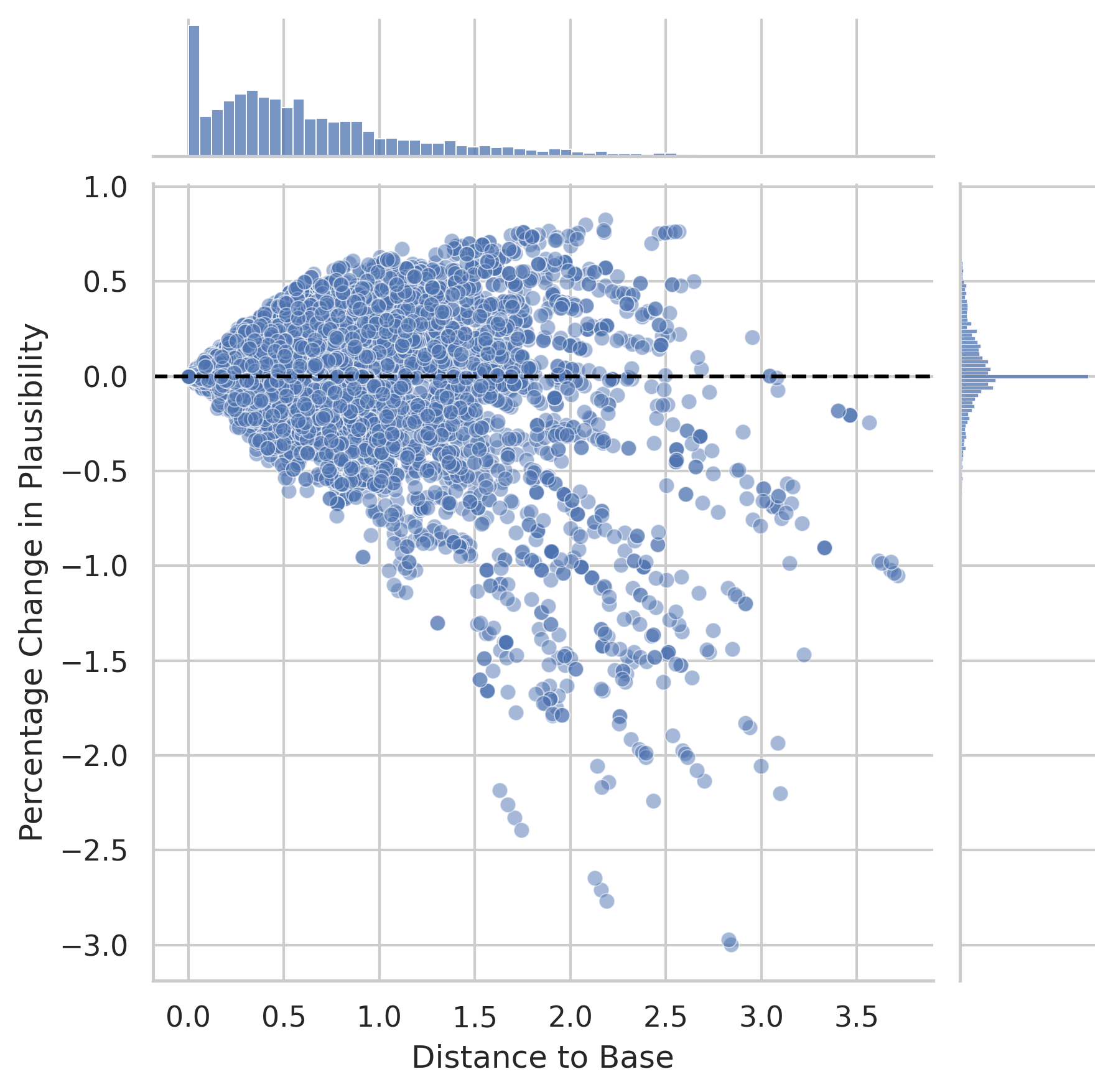}
    \end{subfigure}
\end{center}
    \caption{A plot of the correlation between \textit{distance to base} and other metrics.}
    \label{fig:app:dist-base-correlation}
\end{figure}

\section{The relationship between \textit{Distance to Base} and other metrics}

In this section, we briefly share the empirical observation that the \textit{Distance to Base} metric is negatively correlated with the degradation of CFE performance metrics, focusing on different properties of CFE. 
Hence, we believe that post-hoc methods which introduce smaller changes to the original CFE (in terms of \textit{Distance to Base}) are generally better at preserving its properties.
Fig.~\ref{fig:app:dist-base-correlation} illustrates how proximity metrics (both L1 and L2) deteriorate on average as \textit{Distance to Base} increases. We also note that this deterioration is significantly less prominent with the \textit{Plausibility} metric. 

To support these findings, we calculated correlation coefficients in Tab.~\ref{tab:app:correlation}.

\begin{table}[htbp]
\centering
    \footnotesize
\caption{Correlation between \textit{distance to base} and other metrics. All p-values are lesser than 0.001.}
\label{tab:app:correlation}
    \begin{tabular}{ll|c}

Metric & Correlation type & rho \\ \hline
\textit{Proximity L1} & Pearson & -0.31  \\ 
 \textit{Proximity L1} & Spearman & -0.70  \\ 
 \textit{Proximity L2} & Pearson & -0.31  \\ 
 \textit{Proximity L2} & Spearman & -0.73  \\ 
 \textit{Plausibility} & Pearson & -0.31 \\ 
\textit{Plausibility} & Spearman & -0.02\\ 
Average Pearson &  & -0.31 \\ 
Average Spearman &  & -0.48  \\ 

    \end{tabular}
\end{table}

\section{Background on Bernoulli and Beta Distributions}
\label{app:bernouli-beta-tutorial}

Before getting into the specifics of \betaRCE parameters, it's crucial to understand the foundational distributions underlying our method: the Bernoulli distribution and the Beta distribution.

\subsection{Bernoulli Distribution}
The Bernoulli distribution is a discrete probability distribution for a random variable that takes only two values, typically 0 and 1. It's often used to model binary outcomes, such as success/failure or yes/no scenarios. The probability density function (PDF) of a Bernoulli distribution is given by:

\begin{equation}
P(X = x) = p^x(1-p)^{1-x}, \quad x \in \{0,1\}
\end{equation}

where $p$ is the probability of success (i.e., $X = 1$).

\subsection{Beta Distribution}
The Beta distribution is a continuous probability distribution defined on the interval [0, 1]. It's characterized by two shape parameters, $a$ and $b$, which control its shape. The PDF of a Beta distribution is:

\begin{equation}
f(x; a, b) = \frac{x^{a-1}(1-x)^{b-1}}{B(a, b)}, \quad 0 \leq x \leq 1
\end{equation}

where $B(\alpha, \beta)$ is the Beta function.

\subsection{Conjugate Relationship and Conjugate Priors}

In Bayesian statistics, a conjugate prior is a prior distribution that, when combined with the likelihood function, yields a posterior distribution of the same family as the prior. This property is particularly useful for computational and analytical reasons.

The Beta distribution is the conjugate prior for the Bernoulli distribution. To understand this intuitively:

\begin{itemize}
    \item Imagine we're trying to estimate the probability $p$ of a coin landing heads.
    \item Our prior belief about $p$ is represented by a Beta distribution, Beta$(a, b)$.
    \item We then observe a series of coin flips (Bernoulli trials).
    \item After observing these trials, our updated belief (the posterior) about $p$ is still a Beta distribution, just with updated parameters.
\end{itemize}

Mathematically, this relationship is expressed as:

\begin{equation}
\text{Prior: } p \sim \text{Beta}(\alpha, \beta)
\end{equation}
\begin{equation}
\text{Likelihood: } X | p \sim \text{Bernoulli}(p)
\end{equation}
\begin{equation}
\text{Posterior: } p | X \sim \text{Beta}(\alpha + \sum x_i, \beta + n - \sum x_i)
\end{equation}

where $n$ is the number of observations and $\sum x_i$ is the number of successes (heads).

\section{Comparative Analysis}
\label{app:sec:comparative-analysis}

In this section, we present the comprehensive results from the comparative study detailed in Sec.~\ref{sec:exp-comp}. 
Parameters used in a given method are listed next to this method's name; for \robx these are $\tau$ and variance, while for \betaRCE -- $\delta$ ($\alpha=0.9$). The values in each cell represent the mean $\pm$ standard error. 
The column \textbf{Type} sorts the methods by categories. 
The abbreviations Btsr and Arch used next to \betaRCE in the \textbf{Type} column stand for Bootstrap and Architecture, respectively.
Below present results for three base models: neural network (NN), LightGBM, logistic regression (LR).

\begin{itemize}
    \item \textbf{NN}: Tab.~\ref{app:tab:nn-gs} provides the extended version of Tab.~\ref{tab:comparative-analysis}, including results across all datasets, with \growingSpheres used as the base CFE method. Additionally, in Tab.~\ref{app:tab:nn-dice}, we present the results for when \dice generates base CFEs.

    \item \textbf{LightGBM}: Tab.~\ref{app:tab:lgbm-gs} contains the results for the scenario where LightGBM is the underlying black-box model, and \growingSpheres is employed as the base CFE generation method.

    \item \textbf{LR}: Tab.~\ref{app:tab:lr-gs} similarly provides results for \growingSpheres as a base CFE generation method, but with logistic regressor employed as a black-box model. 
    
\end{itemize}

\begin{table*}[ht]
    \centering
    \scriptsize
    \setlength{\tabcolsep}{2pt}
    \caption{Comparative study results when the underlying model is a neural network and both \robx~ and \betaRCE are using \growingSpheres as the base counterfactual explainer}
    \label{app:tab:nn-gs}
    \begin{tabular}{ccl|cccc|ccc}
    \toprule
        \multirow{2}{*}{\textbf{Dataset}} & \multirow{2}{*}{\textbf{Type}} & \multirow{2}{*}{\textbf{Method}} & \multicolumn{4}{c}{\textbf{Metrics}} & \multicolumn{3}{c}{\textbf{Empirical Robustness}} \\  \cmidrule(r){4-7} \cmidrule(r){8-10} 
         &  &  & Dist. to Base $\downarrow$ & Proximity L1 $\downarrow$ & Proximity L2 $\downarrow$ & Plausibility $\downarrow$ & Architecture $\uparrow$ & Bootstrap $\uparrow$ & Seed $\uparrow$ \\ \midrule
        \multirow{13}{*}{\rotatebox[origin=c]{90}{Diabetes}} &  \multirow{3}{*}{Standard CFEs} & \dice & - & 1.002 $\pm$ 0.001 & $0.645 \pm 0.001$ & 0.499 $\pm$ 0.001 & 0.916 $\pm$ 0.002 & 0.889 $\pm$ 0.003 & 0.916 $\pm$ 0.003 \\ 
         & & \growingSpheres & - & 0.800 $\pm$ 0.001 & $0.345 \pm 0.001$ & 0.358 $\pm$ 0.001 & 0.939 $\pm$ 0.003 & 0.852 $\pm$ 0.003 & 0.853 $\pm$ 0.003 \\ 
         & & \face & - & $0.880 \pm 0.001$ & $0.401 \pm 0.001$ & $0.248 \pm 0.001$ &  $0.869 \pm 0.005$  & $0.694 \pm 0.006$ & $0.721 \pm 0.006$ \\ \cline{2-10}
         & \multirow{2}{*}{Robust end-to-end}  & \rbr & - & $0.714 \pm 0.001$ & $0.339 \pm 0.001$ & $0.319 \pm 0.001$ & $0.618 \pm 0.006$ & $0.617 \pm 0.005$ & $0.576 \pm 0.005$ \\ 
         & & \roar & - & $10.887 \pm 0.001$ & $4.703 \pm 0.001$ & $4.424 \pm 0.001$ & $0.415 \pm 0.005$  & $0.417 \pm 0.005$  & $0.408 \pm 0.005$ \\ \cline{2-10}
         & \multirow{2}{*}{Robust post-hoc} & \robx(0.5,0.1) & 1.224 $\pm$ 0.001 & 1.432 $\pm$ 0.001 & $0.651 \pm 0.001$ & 0.324 $\pm$ 0.001 & 0.998 $\pm$ 0.001 & 0.947 $\pm$ 0.004 & 0.969 $\pm$ 0.004 \\ 
         & & \robx(0.5,0.01) & 0.429 $\pm$ 0.001 & 0.748 $\pm$ 0.001 & $0.339 \pm 0.001$& 0.289 $\pm$ 0.001 & 0.970 $\pm$ 0.003 & 0.872 $\pm$ 0.006 & 0.922 $\pm$ 0.005 \\ 
         & \multirow{2}{*}{\betaRCE Arch}  & \betaRCE(0.8) & 0.488 $\pm$ 0.013 & 0.870 $\pm$ 0.013 & $0.382 \pm 0.005$ & 0.372 $\pm$ 0.004 & 0.966 $\pm$ 0.004 & - & - \\ 
         &  & \betaRCE(0.9) & 0.607 $\pm$ 0.014 & 0.953 $\pm$ 0.013 & $0.420 \pm 0.006$ & 0.378 $\pm$ 0.004 & 0.975 $\pm$ 0.004 & - & - \\ 
         & \multirow{2}{*}{\betaRCE Btsr}  & \betaRCE(0.8) & 0.445 $\pm$ 0.006 & 0.840 $\pm$ 0.008 & $0.359 \pm 0.003$ & 0.359 $\pm$ 0.002 & - & 0.903 $\pm$ 0.006 & - \\ 
         &  & \betaRCE(0.9) & 0.583 $\pm$ 0.007 & 0.949 $\pm$ 0.008 & $0.407 \pm 0.003$ & 0.369 $\pm$ 0.002 & - & 0.937 $\pm$ 0.005 & - \\ 
         & \multirow{2}{*}{\betaRCE Seed}  & \betaRCE(0.8) & 0.247 $\pm$ 0.006 & 0.813 $\pm$ 0.008 & $0.346 \pm 0.003$ & 0.350 $\pm$ 0.002 & - & - & 0.871 $\pm$ 0.006 \\ 
         &  & \betaRCE(0.9) & 0.315 $\pm$ 0.006 & 0.862 $\pm$ 0.008 & $0.367 \pm 0.003$ & 0.353 $\pm$ 0.002 & - & - & 0.902 $\pm$ 0.006 \\ \midrule

         \multirow{13}{*}{\rotatebox[origin=c]{90}{HELOC}} & \multirow{3}{*}{Standard CFEs} & \dice & - & 3.190 $\pm$ 0.004 & $1.163 \pm 0.001$ & 1.003 $\pm$ 0.001 & 0.912 $\pm$ 0.002 & 0.781 $\pm$ 0.004 & 0.815 $\pm$ 0.003 \\ 
         &  & \growingSpheres & - & 2.782 $\pm$ 0.003 & $0.717 \pm 0.001$ & 0.773 $\pm$ 0.001 & 0.862 $\pm$ 0.003 & 0.794 $\pm$ 0.003 & 0.752 $\pm$ 0.004 \\ \ 
         & & \face & - & $2.254 \pm 0.001$ & $0.659 \pm 0.001$ & $0.441 \pm 0.001$ & $0.829 \pm 0.005$ & $0.717 \pm 0.006$ &  $ 0.717 \pm 0.006$ \\  \cline{2-10}
         & \multirow{2}{*}{Robust end-to-end} & \rbr & - & $1.682 \pm 0.001$ & $0.505 \pm 0.001$ & $0.468 \pm 0.001$ & $0.754 \pm 0.005$ & $0.690 \pm 0.005$ & $0.706 \pm 0.005$ \\  
         & & \roar & - & $19.803 \pm 0.001$ & $5.427 \pm 0.001$ & $4.786 \pm 0.001$ & $0.591 \pm 0.005$ & $0.51 \pm 0.005$ &  $0.588 \pm 0.005$  \\ \cline{2-10}
         & \multirow{2}{*}{Robust post-hoc} & \robx(0.5,0.01) & 1.145 $\pm$ 0.002 & 2.341 $\pm$ 0.001 & $0.636 \pm 0.001$ & 0.598 $\pm$ 0.001 & 0.939 $\pm$ 0.005 & 0.814 $\pm$ 0.007 & 0.890 $\pm$ 0.006 \\ 
         &  & \robx(0.5,0.1) & 3.548 $\pm$ 0.005 & 3.938 $\pm$ 0.004 & $1.144 \pm 0.001$ & 0.575 $\pm$ 0.001 & 0.991 $\pm$ 0.002 & 0.957 $\pm$ 0.004 & 0.955 $\pm$ 0.005 \\ \ 
         & \multirow{2}{*}{\betaRCE Arch}  & \betaRCE(0.8) & 1.538 $\pm$ 0.049 & 2.912 $\pm$ 0.053 & $0.749 \pm 0.014$& 0.802 $\pm$ 0.011 & 0.904 $\pm$ 0.006 & - & - \\ 
         &  & \betaRCE(0.9) & 1.697 $\pm$ 0.031 & 2.927 $\pm$ 0.036 & $0.753 \pm 0.009$ & 0.783 $\pm$ 0.007 & 0.935 $\pm$ 0.005 & - & - \\ 
         & \multirow{2}{*}{\betaRCE Btsr} & \betaRCE(0.8) & 2.288 $\pm$ 0.041 & 3.451 $\pm$ 0.044 & $0.889 \pm 0.011$ & 0.859 $\pm$ 0.008 & - & 0.833 $\pm$ 0.007 & - \\ 
         &  & \betaRCE(0.9) & 3.547 $\pm$ 0.071 & 4.501 $\pm$ 0.073 & $1.156 \pm 0.019$ & 1.044 $\pm$ 0.015 & - & 0.880 $\pm$ 0.006 & - \\ 
         & \multirow{2}{*}{\betaRCE Seed}  & \betaRCE(0.8) & 1.420 $\pm$ 0.021 & 2.526 $\pm$ 0.028 & $0.653 \pm 0.007$ & 0.726 $\pm$ 0.004 & - & - & 0.826 $\pm$ 0.007 \\ 
         &  & \betaRCE(0.9) & 1.927 $\pm$ 0.030 & 2.906 $\pm$ 0.035 & $0.750 \pm 0.009$ & 0.776 $\pm$ 0.006 & - & - & 0.902 $\pm$ 0.006 \\ \midrule

         \multirow{13}{*}{\rotatebox[origin=c]{90}{Wine}} & \multirow{3}{*}{Standard CFEs} & \dice & - & 0.888 $\pm$ 0.001          & 0.549 $\pm$ 0.001 & 0.413 $\pm$ 0.001 & 0.934 $\pm$ 0.002 & 0.839 $\pm$ 0.003 & 0.873 $\pm$ 0.003 \\  
         &  & \growingSpheres & - & 0.474 $\pm$ 0.001                                                                               & 0.174 $\pm$ 0.001 & 0.211 $\pm$ 0.001 & 0.877 $\pm$ 0.003 & 0.837 $\pm$ 0.003 & 0.877 $\pm$ 0.003 \\  
         & & \face & - &0.54 $\pm$ 0.001 & 0.216 $\pm$ 0.001	& 0.131 $\pm$ 0.001 & 0.78 $\pm$ 0.003	& 0.747 $\pm$ 0.003 & 0.783 $\pm$ 0.003  \\  \cline{2-10}
         & \multirow{2}{*}{Robust end-to-end} & \rbr & - & 0.508 $\pm$ 0.001 & 0.199 $\pm$ 0.001,	& 0.176 $\pm$ 0.001 & 0.749 $\pm$ 0.002	& 0.73 $\pm$ 0.002	& 0.764 $\pm$ 0.002\\  
         & & \roar & - & 15.768 $\pm$ 0.001 & 5.607 $\pm$ 0.001& 5.26 $\pm$ 0.001& 0.734 $\pm$ 0.002	& 0.755 $\pm$ 0.002	& 0.727 $\pm$ 0.002  \\ \cline{2-10}
         & \multirow{2}{*}{Robust post-hoc}  & \robx(0.5,0.01) & 0.564 $\pm$ 0.001 & 0.674 $\pm$ 0.001              & 0.293 $\pm$ 0.001 & 0.159 $\pm$ 0.001 & 1.000 $\pm$ 0.001 & 0.979 $\pm$ 0.003 & 0.997 $\pm$ 0.001 \\  
         &  & \robx(0.5,0.1) & 1.318 $\pm$ 0.002 & 1.378 $\pm$ 0.002                                                & 0.593 $\pm$ 0.001 & 0.236 $\pm$ 0.001 & 1.000 $\pm$ 0.001 & 0.970 $\pm$ 0.004 & 1.000 $\pm$ 0.001 \\  
         & \multirow{2}{*}{Arch}  & BetaRCE(0.8) & 0.330 $\pm$ 0.005 & 0.525 $\pm$ 0.006                       & 0.192 $\pm$ 0.002 & 0.219 $\pm$ 0.002   & 0.918 $\pm$ 0.005 & - & - \\  
         &  & BetaRCE(0.9) & 0.471 $\pm$ 0.006 & 0.647 $\pm$ 0.007                                             & 0.237 $\pm$ 0.003 & 0.242 $\pm$ 0.002  & 0.959 $\pm$ 0.004 & - & - \\  
         & \multirow{2}{*}{Btsp} & BetaRCE(0.8) & 0.335 $\pm$ 0.005  & 0.513 $\pm$ 0.006                       & 0.188 $\pm$ 0.002 & 0.215 $\pm$ 0.002 & - & 0.882 $\pm$ 0.006 & - \\  
         &  & BetaRCE(0.9) & 0.455 $\pm$ 0.006 & 0.610 $\pm$ 0.006                                             & 0.224 $\pm$ 0.002 & 0.229 $\pm$ 0.002 & - & 0.926 $\pm$ 0.005 & - \\ 
         & \multirow{2}{*}{Seed} & BetaRCE(0.8) & 0.308 $\pm$ 0.005 & 0.480 $\pm$ 0.006                        & 0.177 $\pm$ 0.002 & 0.210 $\pm$ 0.002 & - & - & 0.914 $\pm$ 0.005 \\  
         &  & BetaRCE(0.9) & 0.399 $\pm$ 0.006 & 0.555 $\pm$ 0.006                                             & 0.205 $\pm$ 0.002 & 0.217 $\pm$ 0.002 &  - & - & 0.954 $\pm$ 0.004 \\  \midrule

        \multirow{13}{*}{\rotatebox[origin=c]{90}{Breast Cancer}} & \multirow{3}{*}{Standard CFEs} & \dice & - & 3.004 $\pm$ 0.003    & 1.157 $\pm$ 0.001 & 1.113 $\pm$ 0.001 & 0.894 $\pm$ 0.003 & 0.816 $\pm$ 0.003 & 0.838 $\pm$ 0.003 \\
        & & \growingSpheres & - & 3.811 $\pm$ 0.001                                                      & 0.864 $\pm$ 0.001 & 0.949 $\pm$ 0.001 & 0.898 $\pm$ 0.003 & 0.886 $\pm$ 0.003 & 0.886 $\pm$ 0.003 \\
        & & \face & - &3.427 $\pm$ 0.001	& 0.785 $\pm$ 0.001	& 0.416 $\pm$ 0.001 & 0.93 $\pm$ 0.002	& 0.868 $\pm$ 0.003	& 0.905 $\pm$ 0.003 \\ \cline{2-10}
        & \multirow{2}{*}{Robust end-to-end} & \rbr & - & 2.653 $\pm$ 0.001	& 0.617 $\pm$ 0.001	& 0.547 $\pm$ 0.001& 0.377 $\pm$ 0.002	& 0.343 $\pm$ 0.002	& 0.352 $\pm$ 0.002 \\  
        & & \roar & - &9.271 $\pm$ 0.001	& 2.057 $\pm$ 0.001	& 1.517 $\pm$ 0.001	 & 0.386 $\pm$ 0.002	& 0.384 $\pm$ 0.002	& 0.378 $\pm$ 0.002 \\ \cline{2-10}
        & \multirow{2}{*}{Robust post-hoc} & \robx(0.5,0.01) & 1.251 $\pm$ 0.001 & 3.135 $\pm$ 0.001                 & 0.706 $\pm$ 0.001 & 0.728 $\pm$ 0.001  & 0.848 $\pm$ 0.007 & 0.957 $\pm$ 0.004 & 0.846 $\pm$ 0.007 \\
        & & \robx(0.5,0.1) & 3.163 $\pm$ 0.001 & 3.956 $\pm$ 0.001                                                   & 0.876 $\pm$ 0.001 & 0.563 $\pm$ 0.001 & 0.997 $\pm$ 0.001 & 1.000 $\pm$ 0.001 & 0.996 $\pm$ 0.001 \\
        & \multirow{2}{*}{Arch} & BetaRCE(0.8) & 1.684 $\pm$ 0.026 & 4.108 $\pm$ 0.04                          & 0.916 $\pm$ 0.009 & 0.969 $\pm$ 0.007 & 0.925 $\pm$ 0.005 & - & - \\
        & & BetaRCE(0.9) & 2.058 $\pm$ 0.028 & 4.324 $\pm$ 0.042                                               & 0.964 $\pm$ 0.009 & 0.996 $\pm$ 0.007 & 0.961 $\pm$ 0.004 & - & - \\ 
        & \multirow{2}{*}{Btsp} & BetaRCE(0.8) & 1.528 $\pm$ 0.024 & 3.937 $\pm$ 0.041                         & 0.901 $\pm$ 0.009 & 0.996 $\pm$ 0.008 & - & 0.926 $\pm$ 0.005 & - \\
        & & BetaRCE(0.9) & 1.965 $\pm$ 0.026 & 4.176 $\pm$ 0.042                                               & 0.954 $\pm$ 0.01 & 1.024 $\pm$ 0.008 & - & 0.955 $\pm$ 0.004 & - \\ 
        & \multirow{2}{*}{Seed} & BetaRCE(0.8) & 1.749 $\pm$ 0.028 & 3.788 $\pm$ 0.039                         & 0.863 $\pm$ 0.009 & 0.940 $\pm$ 0.007 & - & - & 0.920 $\pm$ 0.005 \\
        & & BetaRCE(0.9) & 2.115 $\pm$ 0.032 & 4.000 $\pm$ 0.042                                               & 0.91 $\pm$ 0.009 & 0.967 $\pm$ 0.007 & - & - & 0.945 $\pm$ 0.004 \\

        \midrule
        \multirow{13}{*}{\rotatebox[origin=c]{90}{Car eval}} &  \multirow{3}{*}{Standard CFEs} & \dice & - & 1.189 $\pm$ 0.001 & $0.899 \pm 0.001$ & 0.469 $\pm$ 0.001 & 0.866 $\pm$ 0.003 & 0.800 $\pm$ 0.004 & 0.825 $\pm$ 0.004 \\ 
         & & \growingSpheres & - & $0.965 \pm 0.001$ & $0.469 \pm 0.001$ & $0.487 \pm 0.001$ &  $0.592 \pm 0.006$  & $0.566 \pm 0.005$ & $0.608 \pm 0.005$ \\ 
         & & \face & - & 0.977 $\pm$ 0.001 & $0.633 \pm 0.001$ & 0.441 $\pm$ 0.001 & 0.761 $\pm$ 0.004 & 0.826 $\pm$ 0.004 & 0.766 $\pm$ 0.004 \\ 
         \cline{2-10}
         & \multirow{2}{*}{Robust end-to-end}  & \rbr & - & $0.822 \pm 0.001$ & $0.505 \pm 0.001$ & $0.460 \pm 0.001$ & $0.603 \pm 0.001$ & $0.649 \pm 0.001$ & $0.661 \pm 0.001$ \\ 
         & & \roar & - & $1.637 \pm 0.001$ & $0.739 \pm 0.001$ & $0.746 \pm 0.001$ & $0.277 \pm 0.001$  & $0.170 \pm 0.001$  & $0.329 \pm 0.002$ \\ \cline{2-10}
         & \multirow{2}{*}{Robust post-hoc} & RobX(0.5,0.01) & 0.254 $\pm$ 0.002 & 1.110 $\pm$ 0.013 & $0.573 \pm 0.006$ & 0.474 $\pm$ 0.001 & 0.844 $\pm$ 0.007 & 0.842 $\pm$ 0.007 & 0.903 $\pm$ 0.006 \\ 
         & & RobX(0.5,0.1) & 0.895 $\pm$ 0.006 & 1.604 $\pm$ 0.015 & $0.876 \pm 0.007$ & 0.456 $\pm$ 0.001 & 0.980 $\pm$ 0.003 & 0.995 $\pm$ 0.001 & 0.994 $\pm$ 0.001 \\ 
         & \multirow{2}{*}{\betaRCE Arch}  & \betaRCE(0.8) & 0.277 $\pm$ 0.014 & 0.948 $\pm$ 0.031 & $0.456 \pm 0.014$ & 0.474 $\pm 0.002$ & 0.817 $\pm$ 0.021 & - & - \\ 
         &  & \betaRCE(0.9) & 0.353 $\pm$ 0.018 & 0.966 $\pm$ 0.027 & $0.481 \pm 0.013$ & 0.474 $\pm 0.002$ & 0.850 $\pm$ 0.020 & - & - \\ 
         & \multirow{2}{*}{\betaRCE Btsr}  & \betaRCE(0.8) & 0.279 $\pm$ 0.005 & 1.148 $\pm$ 0.013 & $0.561 \pm 0.006$ & 0.487 $\pm$ 0.001 & - & 0.925 $\pm$ 0.005 & - \\ 
         &  & \betaRCE(0.9) & 0.324 $\pm$ 0.006 & 1.172 $\pm$ 0.013 & $0.573 \pm 0.006$ & 0.485 $\pm$ 0.001 & - & 0.948 $\pm$ 0.004 & - \\ 
         & \multirow{2}{*}{\betaRCE Seed}  & \betaRCE(0.8) & 0.27 $\pm$ 0.005 & 1.154 $\pm$ 0.013 & $0.561 \pm 0.006$ & 0.487 $\pm$ 0.001 & - & - & 0.939 $\pm$ 0.005 \\ 
         &  & \betaRCE(0.9) & 0.331 $\pm$ 0.005 & 1.193 $\pm$ 0.013 & $0.581 \pm 0.006$ & 0.484 $\pm$ 0.001 & - & - & 0.972 $\pm$ 0.003 \\ \midrule

        \multirow{13}{*}{\rotatebox[origin=c]{90}{Rice}} &  \multirow{3}{*}{Standard CFEs} & \dice & - & 0.905 $\pm$ 0.001 & $0.681 \pm 0.001$ & 0.493 $\pm$ 0.001 & 0.791 $\pm$ 0.004 & 0.804 $\pm$ 0.004 & 0.751 $\pm$ 0.004 \\ 
         & & \growingSpheres & - & $0.863 \pm 0.001$ & $0.391 \pm 0.001$ & $0.250 \pm 0.001$ &  $0.615 \pm 0.005$  & $0.669 \pm 0.005$ & $0.530 \pm 0.005$ \\ 
         & & \face & - & 0.805 $\pm$ 0.001 & $0.341 \pm 0.001$ & 0.076 $\pm$ 0.001 & 0.619 $\pm$ 0.005 & 0.611 $\pm$ 0.005 & 0.566 $\pm$ 0.005 \\ 
         \cline{2-10}
         & \multirow{2}{*}{Robust end-to-end}  & \rbr & - & $0.890 \pm 0.001$ & $0.396 \pm 0.001$ & $0.213 \pm 0.001$ & $0.413 \pm 0.001$ & $0.421 \pm 0.001$ & $0.425 \pm 0.001$ \\ 
         & & \roar & - & $1.974 \pm 0.001$ & $0.813 \pm 0.001$ & $0.606 \pm 0.001$ & $0.231 \pm 0.001$  & $0.340 \pm 0.002$  & $0.284 \pm 0.001$ \\ \cline{2-10}
         & \multirow{2}{*}{Robust post-hoc} & RobX(0.5,0.01) & 0.413 $\pm$ 0.004 & 1.036 $\pm$ 0.008 & $0.445 \pm 0.004$ & 0.141 $\pm$ 0.002 & 0.940 $\pm$ 0.005 & 0.979 $\pm$ 0.003 & 0.953 $\pm$ 0.004 \\ 
         & & RobX(0.5,0.1) & 1.021 $\pm$ 0.005 & 1.627 $\pm$ 0.009 & $0.694 \pm 0.004$ & 0.101 $\pm$ 0.001 & 1.000 $\pm$ 0.001 & 1.000 $\pm$ 0.001 & 1.000 $\pm$ 0.001 \\ 
         & \multirow{2}{*}{\betaRCE Arch}  & \betaRCE(0.8) & 0.143 $\pm$ 0.003 & 1.063 $\pm$ 0.013 & $0.487 \pm 0.006$ & 0.245 $\pm 0.001$ & 0.821 $\pm$ 0.021 & - & - \\ 
         &  & \betaRCE(0.9) & 0.205 $\pm$ 0.003 & 1.070 $\pm$ 0.013 & $0.491 \pm 0.006$ & 0.250 $\pm 0.001$ & 0.850 $\pm$ 0.020 & - & - \\ 
         & \multirow{2}{*}{\betaRCE Btsr}  & \betaRCE(0.8) & 0.138 $\pm$ 0.003 & 1.167 $\pm$ 0.009 & $0.591 \pm 0.004$ & 0.265 $\pm$ 0.001 & - & 0.895 $\pm$ 0.004 & - \\ 
         &  & \betaRCE(0.9) & 0.187 $\pm$ 0.004 & 1.172 $\pm$ 0.009 & $0.593 \pm 0.004$ & 0.266 $\pm$ 0.001 & - & 0.923 $\pm$ 0.003 & - \\ 
         & \multirow{2}{*}{\betaRCE Seed}  & \betaRCE(0.8) & 0.286 $\pm$ 0.005 & 1.156 $\pm$ 0.009 & $0.587 \pm 0.004$ & 0.261 $\pm$ 0.001 & - & - & 0.919 $\pm$ 0.004 \\ 
         &  & \betaRCE(0.9) & 0.345 $\pm$ 0.005 & 1.162 $\pm$ 0.009 & $0.589 \pm 0.004$ & 0.263 $\pm$ 0.001 & - & - & 0.953 $\pm$ 0.003 \\ 
    \bottomrule 
         
    \end{tabular}
\end{table*}

\begin{table*}[ht]
    \centering
    \scriptsize
    \setlength{\tabcolsep}{2pt}
    \caption{Comparative study results when the underlying model is a neural network and both \robx~ and \betaRCE are using \dice as the base counterfactual explainer.}
    \label{app:tab:nn-dice}
    \begin{tabular}{ccl|cccc|ccc}
    \toprule
        \multirow{2}{*}{\textbf{Dataset}} & \multirow{2}{*}{\textbf{Type}} & \multirow{2}{*}{\textbf{Method}} & \multicolumn{4}{c}{\textbf{Metrics}} & \multicolumn{3}{c}{\textbf{Empirical Robustness}} \\  \cmidrule(r){4-7} \cmidrule(r){8-10} 
         &  &  & Dist. to Base $\downarrow$ & Proximity L1 $\downarrow$ & Proximity L2 $\downarrow$ & Plausibility $\downarrow$ & Architecture $\uparrow$ & Bootstrap $\uparrow$ & Seed $\uparrow$ \\ \midrule

\multirow{13}{*}{\rotatebox[origin=c]{90}{Diabetes}} &  \multirow{3}{*}{Standard CFEs} & \dice & -& $0.872 \pm 0.001$ & $0.685 \pm 0.001$ & $0.49 \pm 0.001$ & $0.866 \pm 0.001$ & $0.7 \pm 0.002$ & $0.745 \pm 0.002$ \\ 
         & & \growingSpheres & -& $0.596 \pm 0.001$ & $0.257 \pm 0.001$ & $0.335 \pm 0.001$ & $0.726 \pm 0.009$ & $0.639 \pm 0.011$ & $0.552 \pm 0.01$ \\ 
         & & \face & -& $0.846 \pm 0.001$ & $0.39 \pm 0.001$ & $0.248 \pm 0.001$ & $0.864 \pm 0.003$ & $0.692 \pm 0.004$ & $0.726 \pm 0.004$ \\ \cline{2-10}
         & \multirow{2}{*}{Robust end-to-end}  & \rbr & -& $0.718 \pm 0.001$ & $0.339 \pm 0.001$ & $0.318 \pm 0.001$ & $0.606 \pm 0.002$ & $0.594 \pm 0.002$ & $0.569 \pm 0.002$ \\ 
         & & \roar & -& $5.533 \pm 0.001$ & $2.58 \pm 0.001$ & $2.389 \pm 0.001$ & $0.346 \pm 0.002$ & $0.36 \pm 0.002$ & $0.346 \pm 0.002$ \\ \cline{2-10}
         & \multirow{2}{*}{Robust post-hoc} & \robx(0.5,0.01) & $0.6 \pm 0.001$ & $0.796 \pm 0.001$ & $0.377 \pm 0.001$ & $0.274 \pm 0.001$ & $0.982 \pm 0.002$ & $0.765 \pm 0.006$ & $0.816 \pm 0.006$ \\ 
         & & \robx(0.6,0.01) & $0.814 \pm 0.001$ & $0.989 \pm 0.001$ & $0.47 \pm 0.001$ & $0.29 \pm 0.001$ & $0.996 \pm 0.001$ & $0.823 \pm 0.006$ & $0.873 \pm 0.005$ \\ 
         & \multirow{2}{*}{\betaRCE Arch}  & \betaRCE(0.8) & $0.338 \pm 0.004$ & $0.655 \pm 0.006$ & $0.286 \pm 0.003$ & $0.329 \pm 0.002$ & $0.928 \pm 0.006$ & -& - \\ 
         &  & \betaRCE(0.9) & $0.432 \pm 0.005$ & $0.73 \pm 0.006$ & $0.318 \pm 0.003$ & $0.339 \pm 0.002$ & $0.953 \pm 0.005$ & - & - \\ 
         & \multirow{2}{*}{\betaRCE Btsr}  & \betaRCE(0.8) & $0.523 \pm 0.005$ & $0.792 \pm 0.006$ & $0.347 \pm 0.003$ & $0.343 \pm 0.002$ & - & $0.878 \pm 0.006$ & - \\ 
         &  & \betaRCE(0.9) & $0.705 \pm 0.005$ & $0.949 \pm 0.007$ & $0.414 \pm 0.003$ & $0.368 \pm 0.002$ & -& $0.886 \pm 0.006$ & - \\ 
         & \multirow{2}{*}{\betaRCE Seed}  & \betaRCE(0.8) & $0.307 \pm 0.004$ & $0.624 \pm 0.006$ & $0.276 \pm 0.003$ & $0.33 \pm 0.002$ & -& -& $0.87 \pm 0.008$ \\ 
         &  & \betaRCE(0.9) & $0.406 \pm 0.005$ & $0.703 \pm 0.007$ & $0.31 \pm 0.003$ & $0.34 \pm 0.002$ & - & - & $0.884 \pm 0.007$ \\ \midrule

\multirow{13}{*}{\rotatebox[origin=c]{90}{HELOC}} & \multirow{3}{*}{Standard CFEs} & \dice & -& $1.241 \pm 0.001$ & $0.9 \pm 0.001$ & $0.855 \pm 0.001$ & $0.602 \pm 0.002$ & $0.56 \pm 0.002$ & $0.589 \pm 0.002$ \\ 
         & & \growingSpheres & -& $1.946 \pm 0.001$ & $0.504 \pm 0.001$ & $0.674 \pm 0.001$ & $0.543 \pm 0.01$ & $0.556 \pm 0.01$ & $0.467 \pm 0.01$ \\ 
         & & \face & -& $2.235 \pm 0.001$ & $0.653 \pm 0.001$ & $0.439 \pm 0.001$ & $0.826 \pm 0.003$ & $0.712 \pm 0.004$ & $0.707 \pm 0.004$ \\ \cline{2-10}
         & \multirow{2}{*}{Robust end-to-end} & \rbr & -& $1.658 \pm 0.001$ & $0.496 \pm 0.001$ & $0.466 \pm 0.001$ & $0.759 \pm 0.002$ & $0.664 \pm 0.002$ & $0.633 \pm 0.002$ \\  
         & & \roar & -& $9.129 \pm 0.001$ & $2.515 \pm 0.001$ & $2.015 \pm 0.001$ & $0.35 \pm 0.002$ & $0.369 \pm 0.002$ & $0.365 \pm 0.002$ \\ \cline{2-10}
         & \multirow{2}{*}{Robust post-hoc} & \robx(0.5,0.01) & $1.761 \pm 0.001$ & $1.943 \pm 0.001$ & $0.572 \pm 0.001$ & $0.473 \pm 0.001$ & $0.919 \pm 0.004$ & $0.762 \pm 0.006$ & $0.859 \pm 0.005$ \\ 
         & & \robx(0.6,0.01) & $2.52 \pm 0.001$ & $2.59 \pm 0.001$ & $0.763 \pm 0.001$ & $0.461 \pm 0.001$ & $0.987 \pm 0.002$ & $0.886 \pm 0.005$ & $0.966 \pm 0.003$ \\ 
         & \multirow{2}{*}{\betaRCE Arch}  & \betaRCE(0.8) & $1.989 \pm 0.054$ & $2.486 \pm 0.063$ & $0.638 \pm 0.016$ & $0.78 \pm 0.011$ & $0.874 \pm 0.006$ & - & - \\ 
         &  & \betaRCE(0.9) & $2.797 \pm 0.076$ & $3.193 \pm 0.084$ & $0.819 \pm 0.021$ & $0.895 \pm 0.016$ & $0.929 \pm 0.005$ & - & - \\ 
         & \multirow{2}{*}{\betaRCE Btsr} & \betaRCE(0.8) & $2.511 \pm 0.038$ & $3.046 \pm 0.044$ & $0.78 \pm 0.011$ & $0.831 \pm 0.008$ & -& $0.77 \pm 0.008$ &- \\ 
         &  & \betaRCE(0.9) & $3.793 \pm 0.054$ & $4.228 \pm 0.059$ & $1.08 \pm 0.015$ & $1.028 \pm 0.012$ & -& $0.807 \pm 0.008$ & - \\ 
         & \multirow{2}{*}{\betaRCE Seed}  & \betaRCE(0.8) & $1.978 \pm 0.034$ & $2.438 \pm 0.036$ & $0.629 \pm 0.009$ & $0.76 \pm 0.006$ & - & -& $0.927 \pm 0.005$ \\ 
         &  & \betaRCE(0.9) & $2.813 \pm 0.049$ & $3.192 \pm 0.05$ & $0.821 \pm 0.013$ & $0.885 \pm 0.009$ & - &-& $0.95 \pm 0.004$ \\ \midrule

\multirow{13}{*}{\rotatebox[origin=c]{90}{Wine}} & \multirow{3}{*}{Standard CFEs} & \dice & - & 0.674 $\pm$ 0.001 & 0.556 $\pm$ 0.001 & 0.433 $\pm$ 0.001 & $0.781 \pm 0.002$ & $0.719 \pm 0.002$ & $0.749 \pm 0.002$ \\  
         & & \growingSpheres & -& $0.294 \pm 0.001$ & $0.108 \pm 0.001$ & $0.187 \pm 0.001$ & $0.539 \pm 0.01$ & $0.526 \pm 0.01$ & $0.525 \pm 0.01$ \\  
         & & \face & -& $0.528 \pm 0.001$ & $0.21 \pm 0.001$ & $0.132 \pm 0.001$ & $0.78 \pm 0.003$ & $0.747 \pm 0.003$ & $0.783 \pm 0.003$ \\  \cline{2-10}
         & \multirow{2}{*}{Robust end-to-end} & \rbr & -& $0.506 \pm 0.001$ & $0.198 \pm 0.001$ & $0.175 \pm 0.001$ & $0.749 \pm 0.002$ & $0.73 \pm 0.002$ & $0.764 \pm 0.002$ \\  
         & & \roar & -& $8.395 \pm 0.001$ & $3.19 \pm 0.001$ & $2.859 \pm 0.001$ & $0.734 \pm 0.002$ & $0.755 \pm 0.002$ & $0.727 \pm 0.002$ \\ \cline{2-10}
         & \multirow{2}{*}{Robust post-hoc} & \robx(0.5,0.01) & $0.546 \pm 0.001$ & $0.641 \pm 0.001$ & $0.284 \pm 0.001$ & $0.16 \pm 0.001$ & $0.935 \pm 0.004$ & $0.902 \pm 0.004$ & $0.924 \pm 0.004$ \\  
         & & \robx(0.6,0.01) & $0.733 \pm 0.001$ & $0.815 \pm 0.001$ & $0.374 \pm 0.001$ & $0.156 \pm 0.001$ & $0.95 \pm 0.003$ & $0.931 \pm 0.004$ & $0.968 \pm 0.003$ \\  
         & \multirow{2}{*}{Arch} & \betaRCE(0.8) & $0.342 \pm 0.005$ & $0.55 \pm 0.005$ & $0.206 \pm 0.002$ & $0.238 \pm 0.002$ & $0.884 \pm 0.006$ &  - & - \\  
         & & \betaRCE(0.9) & $0.435 \pm 0.006$ & $0.622 \pm 0.006$ & $0.233 \pm 0.002$ & $0.249 \pm 0.002$ & $0.909 \pm 0.006$ & - & -\\  
         & \multirow{2}{*}{Btsp} & \betaRCE(0.8) & $0.528 \pm 0.005$ & $0.701 \pm 0.005$ & $0.265 \pm 0.002$ & $0.257 \pm 0.002$ & - & $0.829 \pm 0.007$ & - \\  
         & & \betaRCE(0.9) & $0.678 \pm 0.005$ & $0.831 \pm 0.006$ & $0.315 \pm 0.002$ & $0.277 \pm 0.002$ & - & $0.847 \pm 0.007$  & - \\ 
         & \multirow{2}{*}{Seed} & \betaRCE(0.8) & $0.281 \pm 0.004$ & $0.476 \pm 0.004$ & $0.179 \pm 0.002$ & $0.222 \pm 0.002$ & - & - &  $0.875 \pm 0.006$\\  
         & & \betaRCE(0.9) & $0.418 \pm 0.005$ & $0.585 \pm 0.005$ & $0.219 \pm 0.002$ & $0.238 \pm 0.002$ & - & - & $0.906 \pm 0.006$  \\   \midrule

\multirow{13}{*}{\rotatebox[origin=c]{90}{Breast Cancer}} & \multirow{3}{*}{Standard CFEs} & \dice & - & 1.623 $\pm$ 0.001 & 1.016 $\pm$ 0.001 & 1.056 $\pm$ 0.001 & $0.559 \pm 0.002$ & $0.596 \pm 0.002$ & $0.505 \pm 0.002$ \\  
         & & \growingSpheres & -& $3.086 \pm 0.003$ & $0.701 \pm 0.001$ & $0.853 \pm 0.001$ & $0.543 \pm 0.01$ & $0.537 \pm 0.01$ & $0.472 \pm 0.01$ \\  
         & & \face & -& $3.427 \pm 0.001$ & $0.785 \pm 0.001$ & $0.416 \pm 0.001$ & $0.93 \pm 0.002$ & $0.868 \pm 0.003$ & $0.905 \pm 0.003$ \\  \cline{2-10}
         & \multirow{2}{*}{Robust end-to-end} & \rbr & -& $2.653 \pm 0.001$ & $0.617 \pm 0.001$ & $0.547 \pm 0.001$ & $0.377 \pm 0.002$ & $0.343 \pm 0.002$ & $0.352 \pm 0.002$ \\  
         & & \roar & -& $9.271 \pm 0.001$ & $2.057 \pm 0.001$ & $1.517 \pm 0.001$ & $0.386 \pm 0.002$ & $0.384 \pm 0.002$ & $0.378 \pm 0.002$ \\ \cline{2-10}
         & \multirow{2}{*}{Robust post-hoc} & \robx(0.5,0.01) & $2.849 \pm 0.001$ & $3.116 \pm 0.001$ & $0.71 \pm 0.001$ & $0.471 \pm 0.001$ & $0.904 \pm 0.004$ & $0.891 \pm 0.005$ & $0.873 \pm 0.005$ \\  
         & & \robx(0.6,0.01) & $3.321 \pm 0.001$ & $3.474 \pm 0.001$ & $0.792 \pm 0.001$ & $0.443 \pm 0.001$ & $0.955 \pm 0.003$ & $0.952 \pm 0.003$ & $0.919 \pm 0.004$ \\  
         & \multirow{2}{*}{Arch} & \betaRCE(0.8) & $1.868 \pm 0.05$ & $3.336 \pm 0.059$ & $0.752 \pm 0.013$ & $0.868 \pm 0.011$ & $0.949 \pm 0.004$ & - & - \\  
         & & \betaRCE(0.9) & $2.547 \pm 0.065$ & $3.822 \pm 0.072$ & $0.858 \pm 0.016$ & $0.94 \pm 0.013$ & $0.961 \pm 0.004$ & -  & - \\  
         & \multirow{2}{*}{Btsp} & \betaRCE(0.8) & $4.707 \pm 0.096$ & $5.454 \pm 0.105$ & $1.213 \pm 0.023$ & $1.211 \pm 0.02$ & - & $0.845 \pm 0.019$ & - \\  
         & & \betaRCE(0.9) & $6.831 \pm 0.137$ & $7.412 \pm 0.144$ & $1.642 \pm 0.032$ & $1.552 \pm 0.027$ & - & $0.85 \pm 0.019$ & - \\ 
         & \multirow{2}{*}{Seed} & \betaRCE(0.8) & $2.813 \pm 0.059$ & $3.66 \pm 0.056$ & $0.824 \pm 0.013$ & $0.927 \pm 0.01$ & - & - & $0.894 \pm 0.006$ \\  
         & & \betaRCE(0.9) & $3.555 \pm 0.067$ & $4.269 \pm 0.065$ & $0.957 \pm 0.014$ & $1.023 \pm 0.012$ & - & - & $0.923 \pm 0.005$ \\  

\midrule

\multirow{13}{*}{\rotatebox[origin=c]{90}{Car eval}} 
         & \multirow{3}{*}{Standard CFEs} 
         & \dice & - & $1.189 \pm 0.0$ & $0.899 \pm 0.0$ & $0.469 \pm 0.0$ & $0.866 \pm 0.003$ & $0.8 \pm 0.003$ & $0.825 \pm 0.003$ \\ 
         & & \growingSpheres & - & $0.965 \pm 0.001$ & $0.469 \pm 0.0$ & $0.487 \pm 0.0$ & $0.592 \pm 0.005$ & $0.566 \pm 0.004$ & $0.608 \pm 0.004$ \\ 
         & & \face & - & $0.977 \pm 0.0$ & $0.633 \pm 0.0$ & $0.441 \pm 0.0$ & $0.761 \pm 0.003$ & $0.826 \pm 0.003$ & $0.766 \pm 0.003$ \\ \cline{2-10}
         & \multirow{2}{*}{Robust end-to-end}  
         & \rbr & - & $0.822 \pm 0.0$ & $0.505 \pm 0.0$ & $0.46 \pm 0.0$ & $0.603 \pm 0.001$ & $0.649 \pm 0.001$ & $0.661 \pm 0.001$ \\ 
         & & \roar & - & $1.637 \pm 0.0$ & $0.739 \pm 0.0$ & $0.746 \pm 0.0$ & $0.277 \pm 0.001$ & $0.17 \pm 0.001$ & $0.329 \pm 0.001$ \\ \cline{2-10}
         & \multirow{2}{*}{Robust post-hoc} 
         & \robx(0.5,0.01) & $0.062 \pm 0.002$ & $1.194 \pm 0.007$ & $0.88 \pm 0.004$ & $0.468 \pm 0.0$ & $0.903 \pm 0.004$ & $0.874 \pm 0.005$ & $0.936 \pm 0.003$ \\ 
         & & \robx(0.5,0.1) & $0.528 \pm 0.007$ & $1.59 \pm 0.009$ & $1.011 \pm 0.004$ & $0.457 \pm 0.0$ & $0.987 \pm 0.002$ & $0.991 \pm 0.001$ & $0.994 \pm 0.001$ \\ 
         & \multirow{2}{*}{\betaRCE Arch}  
         & \betaRCE(0.8) & $0.062 \pm 0.003$ & $1.197 \pm 0.008$ & $0.872 \pm 0.005$ & $0.467 \pm 0.0$ & $0.966 \pm 0.002$ & - & - \\ 
         & & \betaRCE(0.9) & $0.083 \pm 0.003$ & $1.217 \pm 0.008$ & $0.88 \pm 0.005$ & $0.468 \pm 0.0$ & $0.98 \pm 0.002$ & - & - \\ 
         & \multirow{2}{*}{\betaRCE Btsr}  
         & \betaRCE(0.8) & $0.133 \pm 0.004$ & $1.331 \pm 0.008$ & $0.942 \pm 0.005$ & $0.472 \pm 0.001$ & - & $0.977 \pm 0.002$ & - \\ 
         & & \betaRCE(0.9) & $0.156 \pm 0.004$ & $1.34 \pm 0.008$ & $0.94 \pm 0.005$ & $0.47 \pm 0.001$ & - & $0.988 \pm 0.001$ & - \\ 
         & \multirow{2}{*}{\betaRCE Seed}  
         & \betaRCE(0.8) & $0.103 \pm 0.004$ & $1.34 \pm 0.008$ & $0.944 \pm 0.005$ & $0.469 \pm 0.0$ & - & - & $0.944 \pm 0.003$ \\ 
         & & \betaRCE(0.9) & $0.143 \pm 0.004$ & $1.37 \pm 0.008$ & $0.948 \pm 0.005$ & $0.469 \pm 0.0$ & - & - & $0.964 \pm 0.003$ \\ 

\midrule

\multirow{13}{*}{\rotatebox[origin=c]{90}{Rice}} 
         & \multirow{3}{*}{Standard CFEs} 
         & \dice & - & $0.905 \pm 0.0$ & $0.681 \pm 0.0$ & $0.493 \pm 0.0$ & $0.791 \pm 0.003$ & $0.804 \pm 0.003$ & $0.751 \pm 0.003$ \\ 
         & & \growingSpheres & - & $0.863 \pm 0.0$ & $0.391 \pm 0.0$ & $0.25 \pm 0.0$ & $0.615 \pm 0.004$ & $0.669 \pm 0.004$ & $0.53 \pm 0.004$ \\ 
         & & \face & - & $0.805 \pm 0.0$ & $0.341 \pm 0.0$ & $0.076 \pm 0.0$ & $0.619 \pm 0.004$ & $0.611 \pm 0.004$ & $0.566 \pm 0.004$ \\ \cline{2-10}
         & \multirow{2}{*}{Robust end-to-end}  
         & \rbr & - & $0.89 \pm 0.0$ & $0.396 \pm 0.0$ & $0.213 \pm 0.0$ & $0.413 \pm 0.001$ & $0.421 \pm 0.001$ & $0.425 \pm 0.001$ \\ 
         & & \roar & - & $1.974 \pm 0.0$ & $0.813 \pm 0.0$ & $0.606 \pm 0.0$ & $0.231 \pm 0.001$ & $0.34 \pm 0.001$ & $0.284 \pm 0.001$ \\ \cline{2-10}
         & \multirow{2}{*}{Robust post-hoc} 
         & \robx(0.5,0.01) & $0.286 \pm 0.006$ & $1.031 \pm 0.005$ & $0.647 \pm 0.003$ & $0.393 \pm 0.003$ & $0.907 \pm 0.004$ & $0.959 \pm 0.003$ & $0.889 \pm 0.005$ \\ 
         & & \robx(0.5,0.1) & $0.503 \pm 0.007$ & $1.423 \pm 0.006$ & $0.865 \pm 0.003$ & $0.415 \pm 0.001$ & $0.979 \pm 0.002$ & $0.989 \pm 0.002$ & $0.993 \pm 0.001$ \\ 
         & \multirow{2}{*}{\betaRCE Arch}  
         & \betaRCE(0.8) & $0.256 \pm 0.006$ & $1.022 \pm 0.005$ & $0.642 \pm 0.004$ & $0.392 \pm 0.003$ & $0.962 \pm 0.002$ & - & - \\ 
         & & \betaRCE(0.9) & $0.276 \pm 0.006$ & $1.035 \pm 0.005$ & $0.648 \pm 0.004$ & $0.393 \pm 0.003$ & $0.978 \pm 0.002$ & - & - \\ 
         & \multirow{2}{*}{\betaRCE Btsr}  
         & \betaRCE(0.8) & $0.306 \pm 0.006$ & $1.044 \pm 0.005$ & $0.65 \pm 0.004$ & $0.392 \pm 0.003$ & - & $0.984 \pm 0.002$ & - \\ 
         & & \betaRCE(0.9) & $0.326 \pm 0.006$ & $1.051 \pm 0.005$ & $0.654 \pm 0.004$ & $0.392 \pm 0.003$ & - & $0.992 \pm 0.001$ & - \\ 
         & \multirow{2}{*}{\betaRCE Seed}  
         & \betaRCE(0.8) & $0.276 \pm 0.006$ & $1.045 \pm 0.005$ & $0.649 \pm 0.004$ & $0.392 \pm 0.003$ & - & - & $0.899 \pm 0.005$ \\ 
         & & \betaRCE(0.9) & $0.296 \pm 0.006$ & $1.053 \pm 0.005$ & $0.653 \pm 0.004$ & $0.392 \pm 0.003$ & - & - & $0.909 \pm 0.005$ \\

    \bottomrule
    \end{tabular}
\end{table*}

\begin{table*}[ht]
    \centering
    \scriptsize 
    \setlength{\tabcolsep}{2pt}
    \caption{Comparative study results when the underlying model is LightGBM and both \robx~ and \betaRCE are using \growingSpheres as the base counterfactual explainer.  Note, there is no seed experiment presented for LightGBM as different seeds were yielding the same models.}
    \label{app:tab:lgbm-gs}
    
    \begin{tabular}{ccl|cccc|cc}
    \toprule
        \multirow{2}{*}{\textbf{Dataset}} & \multirow{2}{*}{\textbf{Type}} & \multirow{2}{*}{\textbf{Method}} & \multicolumn{4}{c}{\textbf{Metrics}} & \multicolumn{2}{c}{\textbf{Empirical Robustness}} \\  \cmidrule(r){4-7} \cmidrule(r){8-9} 
         &  &  & Dist. to Base $\downarrow$ & Proximity L1 $\downarrow$ & Proximity L2 $\downarrow$ & Plausibility $\downarrow$ & Architecture $\uparrow$ & Bootstrap $\uparrow$ \\ \midrule

\multirow{13}{*}{\rotatebox[origin=c]{90}{Diabetes}} &  \multirow{3}{*}{Standard CFEs} & \dice & - & $0.872 \pm 0.001$ & $0.685 \pm 0.001$ & $0.49 \pm 0.001$ & $0.866 \pm 0.001$ & $0.7 \pm 0.002$  \\ 
         & & \growingSpheres & - & $0.596 \pm 0.001$ & $0.257 \pm 0.001$ & $0.335 \pm 0.001$ & $0.726 \pm 0.009$ & $0.639 \pm 0.011$  \\ 
         & & \face & - & $0.846 \pm 0.001$ & $0.39 \pm 0.001$ & $0.248 \pm 0.001$ & $0.864 \pm 0.003$ & $0.692 \pm 0.004$  \\ \cline{2-9}
         & \multirow{2}{*}{Robust end-to-end}  & \rbr & - & $0.718 \pm 0.001$ & $0.339 \pm 0.001$ & $0.318 \pm 0.001$ & $0.606 \pm 0.002$ & $0.594 \pm 0.002$ \\ 
         & & \roar & - & $5.533 \pm 0.001$ & $2.58 \pm 0.001$ & $2.389 \pm 0.001$ & $0.346 \pm 0.002$ & $0.36 \pm 0.002$ \\ \cline{2-9}
        & \multirow{2}{*}{Robust post-hoc} & \robx(0.5,0.01) & $0.6 \pm 0.001$ & $0.796 \pm 0.001$ & $0.377 \pm 0.001$ & $0.274 \pm 0.001$ & $0.934 \pm 0.004$ & $0.878 \pm 0.005$  \\ 
        & & \robx(0.6,0.01) & $0.814 \pm 0.001$ & $0.989 \pm 0.001$ & $0.47 \pm 0.001$ & $0.29 \pm 0.001$ & $0.998 \pm 0.001$ & $0.961 \pm 0.003$  \\ 
        & \multirow{2}{*}{\betaRCE Arch}  & \betaRCE(0.8) & $0.338 \pm 0.004$ & $0.655 \pm 0.006$ & $0.286 \pm 0.003$ & $0.329 \pm 0.002$ & $0.848 \pm 0.005$ & - \\ 
        &  & \betaRCE(0.9) & $0.432 \pm 0.005$ & $0.73 \pm 0.006$ & $0.318 \pm 0.003$ & $0.339 \pm 0.002$ & $0.888 \pm 0.005$ & - \\ 
        & \multirow{2}{*}{\betaRCE Btsr}  & \betaRCE(0.8) & $0.523 \pm 0.005$ & $0.792 \pm 0.006$ & $0.347 \pm 0.003$ & $0.343 \pm 0.002$ & - & $0.902 \pm 0.005$   \\ 
        &  & \betaRCE(0.9) & $0.705 \pm 0.005$ & $0.949 \pm 0.007$ & $0.414 \pm 0.003$ & $0.368 \pm 0.002$ & - & $0.955 \pm 0.003$  \\ \midrule

        \multirow{13}{*}{\rotatebox[origin=c]{90}{HELOC}} & \multirow{3}{*}{Standard CFEs} & \dice & - & $1.241 \pm 0.001$ & $0.9 \pm 0.001$ & $0.855 \pm 0.001$ & $0.602 \pm 0.002$ & $0.56 \pm 0.002$  \\ 
         &  & \growingSpheres & - & $1.946 \pm 0.001$ & $0.504 \pm 0.001$ & $0.674 \pm 0.001$ & $0.543 \pm 0.01$ & $0.556 \pm 0.01$  \\ 
         & & \face & - & $2.235 \pm 0.001$ & $0.653 \pm 0.001$ & $0.439 \pm 0.001$ & $0.826 \pm 0.003$ & $0.712 \pm 0.004$ \\  \cline{2-9}
         & \multirow{2}{*}{Robust end-to-end} & \rbr & - & $1.658 \pm 0.001$ & $0.496 \pm 0.001$ & $0.466 \pm 0.001$ & $0.759 \pm 0.002$ & $0.664 \pm 0.002$  \\  
         & & \roar & - & $9.129 \pm 0.001$ & $2.515 \pm 0.001$ & $2.015 \pm 0.001$ & $0.35 \pm 0.002$ & $0.369 \pm 0.002$  \\ \cline{2-9}
        & \multirow{2}{*}{Robust post-hoc} & \robx(0.5,0.01) & $1.761 \pm 0.001$ & $1.943 \pm 0.001$ & $0.572 \pm 0.001$ & $0.473 \pm 0.001$ & $0.914 \pm 0.004$ & $0.82 \pm 0.006$  \\ 
        & & \robx(0.6,0.01) & $2.52 \pm 0.001$ & $2.59 \pm 0.001$ & $0.763 \pm 0.001$ & $0.461 \pm 0.001$ & $1.0 \pm 0.001$ & $0.962 \pm 0.003$  \\ 
        & \multirow{2}{*}{\betaRCE Arch}  & \betaRCE(0.8) & $1.989 \pm 0.054$ & $2.486 \pm 0.063$ & $0.638 \pm 0.016$ & $0.78 \pm 0.011$ & 0.867 $\pm$ 0.009 & - \\ 
        &  & \betaRCE(0.9) & $2.797 \pm 0.076$ & $3.193 \pm 0.084$ & $0.819 \pm 0.021$ & $0.895 \pm 0.016$ & 0.889 $\pm$ 0.008 & - \\ 
        & \multirow{2}{*}{\betaRCE Btsr}  & \betaRCE(0.8) & $2.511 \pm 0.038$ & $3.046 \pm 0.044$ & $0.78 \pm 0.011$ & $0.831 \pm 0.008$ & - & 0.875 $\pm$ 0.009  \\ 
        &  & \betaRCE(0.9) & $3.793 \pm 0.054$ & $4.228 \pm 0.059$ & $1.08 \pm 0.015$ & $1.028 \pm 0.012$ & - & 0.94 $\pm$ 0.006 \\ \midrule

    \multirow{13}{*}{\rotatebox[origin=c]{90}{Wine}} & \multirow{3}{*}{Standard CFEs} & \dice & - & $0.674 \pm 0.001$ & $0.556 \pm 0.001$ & $0.433 \pm 0.001$ & $0.781 \pm 0.002$ & $0.719 \pm 0.002$ \\  
         &  & \growingSpheres & - & $0.294 \pm 0.001$ & $0.108 \pm 0.001$ & $0.187 \pm 0.001$ & $0.539 \pm 0.01$ & $0.526 \pm 0.01$  \\  
         & & \face & - & $0.528 \pm 0.001$ & $0.21 \pm 0.001$ & $0.132 \pm 0.001$ & $0.78 \pm 0.003$ & $0.747 \pm 0.003$  \\  \cline{2-9}
         & \multirow{2}{*}{Robust end-to-end} & \rbr & - & $0.506 \pm 0.001$ & $0.198 \pm 0.001$ & $0.175 \pm 0.001$ & $0.749 \pm 0.002$ & $0.73 \pm 0.002$  \\  
         & & \roar & - & $8.395 \pm 0.001$ & $3.19 \pm 0.001$ & $2.859 \pm 0.001$ & $0.734 \pm 0.002$ & $0.755 \pm 0.002$  \\ \cline{2-9}
        & \multirow{2}{*}{Robust post-hoc} & \robx(0.5,0.1) & $0.546 \pm 0.001$ & $0.641 \pm 0.001$ & $0.284 \pm 0.001$ & $0.16 \pm 0.001$ & $0.984 \pm 0.002$ & $0.948 \pm 0.003$  \\ 
        & & \robx(0.6,0.1) & $0.733 \pm 0.001$ & $0.815 \pm 0.001$ & $0.374 \pm 0.001$ & $0.156 \pm 0.001$ & $1.0 \pm 0.001$ & $0.994 \pm 0.001$ \\ 
        & \multirow{2}{*}{\betaRCE Arch}  & \betaRCE(0.8) & $0.342 \pm 0.005$ & $0.55 \pm 0.005$ & $0.206 \pm 0.002$ & $0.238 \pm 0.002$ & $0.893 \pm 0.005$ & - \\ 
        &  & \betaRCE(0.9) & $0.435 \pm 0.006$ & $0.622 \pm 0.006$ & $0.233 \pm 0.002$ & $0.249 \pm 0.002$ & $0.928 \pm 0.004$ & - \\ 
        & \multirow{2}{*}{\betaRCE Btsr}  & \betaRCE(0.8) & $0.528 \pm 0.005$ & $0.701 \pm 0.005$ & $0.265 \pm 0.002$ & $0.257 \pm 0.002$ & - & $0.936 \pm 0.004$  \\ 
        &  & \betaRCE(0.9) & $0.678 \pm 0.005$ & $0.831 \pm 0.006$ & $0.315 \pm 0.002$ & $0.277 \pm 0.002$ & - &  $0.97 \pm 0.003$ \\ \midrule

\multirow{13}{*}{\rotatebox[origin=c]{90}{Breast Cancer}} & \multirow{3}{*}{Standard CFEs} & \dice & - & $1.623 \pm 0.001$ & $1.016 \pm 0.001$ & $1.056 \pm 0.001$ & $0.559 \pm 0.002$ & $0.596 \pm 0.002$  \\ 
         &  & \growingSpheres & - & $3.086 \pm 0.003$ & $0.701 \pm 0.001$ & $0.853 \pm 0.001$ & $0.543 \pm 0.01$ & $0.537 \pm 0.01$  \\ 
         & & \face & - & $3.427 \pm 0.001$ & $0.785 \pm 0.001$ & $0.416 \pm 0.001$ & $0.93 \pm 0.002$ & $0.868 \pm 0.003$  \\  \cline{2-9}
         & \multirow{2}{*}{Robust end-to-end} & \rbr & - & $2.653 \pm 0.001$ & $0.617 \pm 0.001$ & $0.547 \pm 0.001$ & $0.377 \pm 0.002$ & $0.343 \pm 0.002$  \\  
         & & \roar & - & $9.271 \pm 0.001$ & $2.057 \pm 0.001$ & $1.517 \pm 0.001$ & $0.386 \pm 0.002$ & $0.384 \pm 0.002$ \\ \cline{2-9}
        & \multirow{2}{*}{Robust post-hoc} & \robx(0.5,0.1) & $2.849 \pm 0.001$ & $3.116 \pm 0.001$ & $0.71 \pm 0.001$ & $0.471 \pm 0.001$ & $0.959 \pm 0.003$ & $0.904 \pm 0.004$  \\ 
        & & \robx(0.6,0.1) & $3.321 \pm 0.001$ & $3.474 \pm 0.001$ & $0.792 \pm 0.001$ & $0.443 \pm 0.001$ & $0.997 \pm 0.001$ & $0.971 \pm 0.002$ \\ 
        & \multirow{2}{*}{\betaRCE Arch}  & \betaRCE(0.8) & $1.868 \pm 0.05$ & $3.336 \pm 0.059$ & $0.752 \pm 0.013$ & $0.868 \pm 0.011$ & $0.902 \pm 0.008$ & -  \\ 
        &  & \betaRCE(0.9) & $2.547 \pm 0.065$ & $3.822 \pm 0.072$ & $0.858 \pm 0.016$ & $0.94 \pm 0.013$ & $0.936 \pm 0.007$ & -  \\ 
        & \multirow{2}{*}{\betaRCE Btsr}  & \betaRCE(0.8) & $4.707 \pm 0.096$ & $5.454 \pm 0.105$ & $1.213 \pm 0.023$ & $1.211 \pm 0.02$ & - &$0.931 \pm 0.009$  \\ 
        &  & \betaRCE(0.9) & $6.831 \pm 0.137$ & $7.412 \pm 0.144$ & $1.642 \pm 0.032$ & $1.552 \pm 0.027$ & - & $0.964 \pm 0.006$  \\ 
    \bottomrule
    \end{tabular}
\end{table*}

\begin{table*}[ht]
    \centering
    \scriptsize 
    \setlength{\tabcolsep}{2pt}
    \caption{Comparative study results when the underlying model is logistic regression and both \robx~ and \betaRCE are using \dice as the base counterfactual explainer.}
    \label{app:tab:lr-gs}
    \begin{tabular}{ccl|cccc|ccc}
    \toprule
        \multirow{2}{*}{\textbf{Dataset}} & \multirow{2}{*}{\textbf{Type}} & \multirow{2}{*}{\textbf{Method}} & \multicolumn{4}{c}{\textbf{Metrics}} & \multicolumn{3}{c}{\textbf{Empirical Robustness}} \\  \cmidrule(r){4-7} \cmidrule(r){8-10} 
         &  &  & Dist. to Base $\downarrow$ & Proximity L1 $\downarrow$ & Proximity L2 $\downarrow$ & Plausibility $\downarrow$ & Architecture $\uparrow$ & Bootstrap $\uparrow$ & Seed $\uparrow$ \\ \midrule

\multirow{11}{*}{\rotatebox[origin=c]{90}{Breast Cancer}} 
         & Standard CFEs & \growingSpheres & - & $4.53 \pm 0.002$ & $1.029 \pm 0.000$ & $1.044 \pm 0.000$ & $0.595 \pm 0.009$ & $0.576 \pm 0.010$ & $0.704 \pm 0.009$ \\ \cline{2-10}
         & \multirow{2}{*}{Robust end-to-end} & \rbr & - & $2.906 \pm 0.000$ & $0.68 \pm 0.000$ & $0.503 \pm 0.000$ & $0.457 \pm 0.002$ & $0.520 \pm 0.002$ & $0.528 \pm 0.002$ \\ 
         &  & \roar & - & $0.295 \pm 0.002$ & $0.062 \pm 0.000$ & $0.498 \pm 0.000$ & $0.079 \pm 0.001$ & $0.047 \pm 0.001$ & $0.020 \pm 0.001$ \\ \cline{2-10}
         & \multirow{2}{*}{Robust post-hoc} & \robx(0.5,0.01) & $1.141 \pm 0.016$ & $3.986 \pm 0.041$ & $0.905 \pm 0.009$ & $0.832 \pm 0.008$ & $0.782 \pm 0.008$ & $0.769 \pm 0.008$ & $0.944 \pm 0.004$ \\ 
         &  & \robx(0.5,0.1) & $3.170 \pm 0.022$ & $4.293 \pm 0.031$ & $0.963 \pm 0.007$ & $0.582 \pm 0.005$ & $0.987 \pm 0.002$ & $0.984 \pm 0.002$ & $0.999 \pm 0.001$ \\ 
         & \multirow{2}{*}{\betaRCE Arch}  & \betaRCE(0.8,0.9) & $1.649 \pm 0.025$ & $5.083 \pm 0.046$ & $1.152 \pm 0.010$ & $1.104 \pm 0.008$ & $0.937 \pm 0.005$ & - & - \\ 
         &  & \betaRCE(0.9,0.9) & $2.092 \pm 0.027$ & $5.302 \pm 0.047$ & $1.202 \pm 0.011$ & $1.127 \pm 0.008$ & $0.986 \pm 0.002$ & - & - \\ 
         & \multirow{2}{*}{\betaRCE Btsr}  & \betaRCE(0.8,0.9) & $1.529 \pm 0.026$ & $5.166 \pm 0.047$ & $1.173 \pm 0.010$ & $1.134 \pm 0.008$ & - & $0.957 \pm 0.004$ & - \\ 
         &  & \betaRCE(0.9,0.9) & $1.751 \pm 0.028$ & $5.275 \pm 0.047$ & $1.197 \pm 0.011$ & $1.142 \pm 0.008$ & - & $0.988 \pm 0.002$ & - \\ 
         & \multirow{2}{*}{\betaRCE Seed}  & \betaRCE(0.8,0.9) & $0.557 \pm 0.014$ & $5.006 \pm 0.053$ & $1.135 \pm 0.012$ & $1.096 \pm 0.009$ & - & - & $0.956 \pm 0.004$ \\ 
         &  & \betaRCE(0.9,0.9) & $0.748 \pm 0.016$ & $5.075 \pm 0.053$ & $1.153 \pm 0.012$ & $1.103 \pm 0.009$ & - & - & $0.991 \pm 0.002$ \\ 
 \midrule 

 \multirow{11}{*}{\rotatebox[origin=c]{90}{Car Evaluation}} 
     & Standard CFEs & \growingSpheres & - & $0.514 \pm 0.000$ & $0.506 \pm 0.000$ & $1.057 \pm 0.001$ & $0.635 \pm 0.009$ & $0.626 \pm 0.009$ & $0.662 \pm 0.009$ \\ \cline{2-10}
     & \multirow{2}{*}{Robust end-to-end} & \rbr & - & $0.881 \pm 0.000$ & $0.573 \pm 0.000$ & $0.475 \pm 0.000$ & $0.643 \pm 0.002$ & $0.658 \pm 0.002$ & $0.600 \pm 0.002$ \\ 
     &  & \roar & - & $2.323 \pm 0.000$ & $1.018 \pm 0.000$ & $0.912 \pm 0.000$ & $0.391 \pm 0.002$ & $0.424 \pm 0.002$ & $0.422 \pm 0.002$ \\ \cline{2-10}
     & \multirow{2}{*}{Robust post-hoc} & \robx(0.5,0.01) & $0.264 \pm 0.003$ & $1.129 \pm 0.013$ & $0.586 \pm 0.006$ & $0.485 \pm 0.001$ & $0.920 \pm 0.005$ & $0.879 \pm 0.006$ & $0.949 \pm 0.004$ \\ 
     &  & \robx(0.5,0.1) & $0.774 \pm 0.004$ & $1.455 \pm 0.012$ & $0.814 \pm 0.006$ & $0.461 \pm 0.001$ & $1.000 \pm 0.000$ & $0.991 \pm 0.002$ & $1.000 \pm 0.000$ \\ 
     & \multirow{2}{*}{\betaRCE Arch}  & \betaRCE(0.8,0.9) & $0.184 \pm 0.003$ & $1.164 \pm 0.014$ & $0.579 \pm 0.007$ & $0.513 \pm 0.001$ & $0.948 \pm 0.004$ & - & - \\ 
     &  & \betaRCE(0.9,0.9) & $0.210 \pm 0.003$ & $1.192 \pm 0.014$ & $0.593 \pm 0.007$ & $0.513 \pm 0.001$ & $0.984 \pm 0.002$ & - & - \\ 
     & \multirow{2}{*}{\betaRCE Btsr}  & \betaRCE(0.8,0.9) & $0.226 \pm 0.004$ & $1.313 \pm 0.016$ & $0.633 \pm 0.007$ & $0.520 \pm 0.002$ & - & $0.949 \pm 0.004$ & - \\ 
     &  & \betaRCE(0.9,0.9) & $0.281 \pm 0.005$ & $1.355 \pm 0.016$ & $0.654 \pm 0.007$ & $0.522 \pm 0.002$ & - & $0.980 \pm 0.003$ & - \\ 
     & \multirow{2}{*}{\betaRCE Seed}  & \betaRCE(0.8,0.9) & $0.158 \pm 0.003$ & $1.124 \pm 0.014$ & $0.543 \pm 0.007$ & $0.504 \pm 0.001$ & - & - & $0.953 \pm 0.004$ \\ 
     &  & \betaRCE(0.9,0.9) & $0.181 \pm 0.003$ & $1.147 \pm 0.014$ & $0.553 \pm 0.007$ & $0.505 \pm 0.001$ & - & - & $0.987 \pm 0.002$ \\ 

\midrule

\multirow{11}{*}{\rotatebox[origin=c]{90}{Diabetes}} 
     & Standard CFEs & \growingSpheres & - & $0.545 \pm 0.003$ & $0.234 \pm 0.001$ & $0.310 \pm 0.001$ & $0.543 \pm 0.010$ & $0.535 \pm 0.035$ & $0.410 \pm 0.009$ \\ \cline{2-10}
     & \multirow{2}{*}{Robust end-to-end} & \rbr & - & $0.772 \pm 0.000$ & $0.367 \pm 0.000$ & $0.311 \pm 0.000$ & $0.536 \pm 0.002$ & $0.751 \pm 0.003$ & $0.550 \pm 0.002$ \\ 
     &  & \roar & - & $0.766 \pm 0.002$ & $0.303 \pm 0.001$ & $0.443 \pm 0.000$ & $0.306 \pm 0.002$ & $0.112 \pm 0.002$ & $0.156 \pm 0.002$ \\ \cline{2-10}
     & \multirow{2}{*}{Robust post-hoc} & \robx(0.5,0.01) & $0.405 \pm 0.003$ & $0.838 \pm 0.007$ & $0.372 \pm 0.003$ & $0.310 \pm 0.002$ & $0.711 \pm 0.009$ & $0.990 \pm 0.007$ & $0.718 \pm 0.009$ \\ 
     &  & \robx(0.5,0.1) & $1.130 \pm 0.006$ & $1.422 \pm 0.007$ & $0.664 \pm 0.003$ & $0.390 \pm 0.002$ & $0.903 \pm 0.006$ & $1.000 \pm 0.000$ & $0.905 \pm 0.006$ \\ 
     & \multirow{2}{*}{\betaRCE Arch}  & \betaRCE(0.8,0.9) & $0.789 \pm 0.010$ & $1.225 \pm 0.012$ & $0.517 \pm 0.005$ & $0.443 \pm 0.003$ & $0.925 \pm 0.005$ & - & - \\ 
     &  & \betaRCE(0.9,0.9) & $1.068 \pm 0.012$ & $1.501 \pm 0.013$ & $0.626 \pm 0.005$ & $0.499 \pm 0.003$ & $0.977 \pm 0.003$ & - & - \\ 
     & \multirow{2}{*}{\betaRCE Btsr}  & \betaRCE(0.8,0.9) & $0.192 \pm 0.010$ & $0.535 \pm 0.011$ & $0.230 \pm 0.005$ & $0.283 \pm 0.004$ & - & $0.940 \pm 0.017$ & - \\ 
     &  & \betaRCE(0.9,0.9) & $0.200 \pm 0.007$ & $0.542 \pm 0.013$ & $0.239 \pm 0.006$ & $0.285 \pm 0.004$ & - & $0.980 \pm 0.010$ & - \\ 
     & \multirow{2}{*}{\betaRCE Seed}  & \betaRCE(0.8,0.9) & $0.746 \pm 0.014$ & $1.155 \pm 0.016$ & $0.483 \pm 0.006$ & $0.449 \pm 0.004$ & - & - & $0.957 \pm 0.004$ \\ 
     &  & \betaRCE(0.9,0.9) & $0.786 \pm 0.014$ & $1.184 \pm 0.015$ & $0.503 \pm 0.006$ & $0.460 \pm 0.004$ & - & - & $0.987 \pm 0.002$ \\ 

\midrule

 \multirow{11}{*}{\rotatebox[origin=c]{90}{Rice}} 
     & Standard CFEs & \growingSpheres & - & $0.992 \pm 0.000$ & $0.447 \pm 0.000$ & $0.246 \pm 0.000$ & $0.684 \pm 0.009$ & $0.807 \pm 0.008$ & $0.733 \pm 0.009$ \\ \cline{2-10}
     & \multirow{2}{*}{Robust end-to-end} & \rbr & - & $0.932 \pm 0.000$ & $0.420 \pm 0.000$ & $0.193 \pm 0.000$ & $0.414 \pm 0.002$ & $0.409 \pm 0.002$ & $0.424 \pm 0.002$ \\ 
     &  & \roar & - & $1.562 \pm 0.000$ & $0.665 \pm 0.000$ & $0.414 \pm 0.000$ & $0.230 \pm 0.002$ & $0.190 \pm 0.002$ & $0.253 \pm 0.002$ \\ \cline{2-10}
     & \multirow{2}{*}{Robust post-hoc} & \robx(0.5,0.01) & $0.285 \pm 0.003$ & $1.084 \pm 0.009$ & $0.469 \pm 0.004$ & $0.153 \pm 0.002$ & $0.891 \pm 0.006$ & $0.997 \pm 0.001$ & $0.984 \pm 0.002$ \\ 
     &  & \robx(0.5,0.1) & $0.735 \pm 0.003$ & $1.511 \pm 0.008$ & $0.644 \pm 0.003$ & $0.097 \pm 0.001$ & $0.989 \pm 0.002$ & $1.000 \pm 0.000$ & $1.000 \pm 0.000$ \\ 
     & \multirow{2}{*}{\betaRCE Arch}  & \betaRCE(0.8,0.9) & $0.247 \pm 0.005$ & $1.106 \pm 0.010$ & $0.490 \pm 0.004$ & $0.239 \pm 0.002$ & $0.937 \pm 0.005$ & - & - \\ 
     &  & \betaRCE(0.9,0.9) & $0.369 \pm 0.006$ & $1.158 \pm 0.011$ & $0.513 \pm 0.005$ & $0.242 \pm 0.002$ & $0.982 \pm 0.003$ & - & - \\ 
     & \multirow{2}{*}{\betaRCE Btsr}  & \betaRCE(0.8,0.9) & $0.056 \pm 0.002$ & $1.039 \pm 0.010$ & $0.469 \pm 0.004$ & $0.250 \pm 0.002$ & - & $0.969 \pm 0.003$ & - \\ 
     &  & \betaRCE(0.9,0.9) & $0.078 \pm 0.002$ & $1.056 \pm 0.009$ & $0.476 \pm 0.004$ & $0.251 \pm 0.002$ & - & $0.991 \pm 0.002$ & - \\ 
     & \multirow{2}{*}{\betaRCE Seed}  & \betaRCE(0.8,0.9) & $0.079 \pm 0.002$ & $1.029 \pm 0.010$ & $0.461 \pm 0.004$ & $0.246 \pm 0.002$ & - & - & $0.963 \pm 0.004$ \\ 
     &  & \betaRCE(0.9,0.9) & $0.110 \pm 0.002$ & $1.045 \pm 0.010$ & $0.468 \pm 0.004$ & $0.245 \pm 0.002$ & - & - & $0.988 \pm 0.002$ \\

    \bottomrule
    \end{tabular}
\end{table*}

\end{document}